\pdfoutput=1
\documentclass{article}
\usepackage{placeins}
\usepackage{iclr2027_conference,times}
\usepackage{amsmath}
\usepackage{booktabs}
\usepackage{xcolor}
\usepackage{comment}
\usepackage{adjustbox}

\usepackage{subcaption}

\usepackage{amsmath,amsfonts,bm}

\def\eqref#1{equation~\ref{#1}}

\def\1{\bm{1}}

\DeclareMathAlphabet{\mathsfit}{\encodingdefault}{\sfdefault}{m}{sl}
\SetMathAlphabet{\mathsfit}{bold}{\encodingdefault}{\sfdefault}{bx}{n}

\usepackage{hyperref}
\hypersetup{colorlinks=true, allcolors=blue!65!black}
\usepackage{url}

\usepackage{natbib}  
\usepackage{caption} 
\usepackage{subcaption} 
\usepackage{booktabs}
\usepackage{algorithm}
\usepackage{algorithmic}
\usepackage{xcolor}
\definecolor{teal}{rgb}{0,0.5,0.5}

\usepackage{tikz}
\usetikzlibrary{arrows.meta,positioning,fit,backgrounds}
\definecolor{SurveyFill}{RGB}{228,239,249}
\definecolor{SurveyEdge}{RGB}{46,90,138}
\definecolor{ModelFill}{RGB}{252,238,221}
\definecolor{ModelEdge}{RGB}{172,106,32}
\definecolor{CultureFill}{RGB}{227,243,229}
\definecolor{CultureEdge}{RGB}{45,110,60}
\definecolor{MetricFill}{RGB}{238,233,246}
\definecolor{MetricEdge}{RGB}{93,62,140}

\usepackage{newfloat}
\usepackage{listings}

\title{Population Fidelity: Evaluating Population Representativeness in LLMs}

\author {Neemias B. da Silva\textsuperscript{1,2}, Martin Lukk\textsuperscript{1}, Ali Sutani\textsuperscript{1}, Abhishek Moturu\textsuperscript{1}, Harris Yang\textsuperscript{1},\\ \textbf{Daniel Silver\textsuperscript{1}, Matt Ratto\textsuperscript{1}, Thiago H. Silva\textsuperscript{1,2}}  \\
\textsuperscript{1} University of Toronto, Canada\\
\textsuperscript{2} Federal University of Technology -- Paran\'a, Brazil\\
\tiny
\texttt{\{neemias.buceli,ali.sutani,abhishek.moturu,harris.yang\}@mail.utoronto.ca} \\  \tiny \texttt{\{martin.lukk,dan.silver,matt.ratto,th.silva\}@utoronto.ca}
}

\iclrfinalcopy 
\begin{document}

\maketitle

\begin{abstract}
Large language models (LLMs) show considerable potential in simulating human attitudes and preferences. Prior work finds that LLM-generated responses can compress the range of attitudes found within populations and misrepresent particular subgroups in ways that vary across models and topics. We introduce Population Fidelity, an evaluation framework that distinguishes key conditions required for a set of LLM-generated responses to represent a population. It incorporates three dimensions: group-level accuracy, the amount of between-group variation, and the structure of that variation. We demonstrate the framework's utility in two ways. First, we reproduce a prior study of ``machine bias'' in LLM survey responses and apply the framework to its models and more recent ones, showing that poor representation reflects not only insufficient between-group variation but also variation assigned to the wrong groups. Second, we evaluate one proposed approach to improving models' population representativeness: cultural fine-tuning. We find that cultural fine-tuning can improve alignment with the survey center without improving the representation of within-population differences, a distinction that measures of aggregate agreement do not capture. We argue that representing a population requires models to reproduce several features of human attitudinal variation simultaneously. Our framework organizes these features and provides reusable code, data, and trained models for evaluating population fidelity across substantive domains and assessing proposed alignment methods.
\end{abstract}

\section{Introduction}

Large language models (LLMs) show considerable potential in simulating human attitudes, behaviors, and preferences in wide-ranging applied and research contexts \citep{bailCanGenerativeAI2024, davidsonIntegratingGenerativeArtificial2025, karetnikovLargeLanguageModels2026}. There is particular interest in using LLMs to generate synthetic survey responses as proxies for human responses. With appropriate prompting and context, the resulting ``silicon samples'' should reproduce the response distributions of human populations \citep{argyle2023out, dillionCanAILanguage2023}. This could enable inferences about population-level attitudes on important social, cultural, and political issues, especially as traditional survey methods continue to face threats to their reliability \citep{keeterWhatLowResponse2017, zhangGenerativeAIMeets2025}.

In practice, researchers have raised serious concerns about LLMs' ability to reproduce human response patterns. Several studies, focusing on GPT models, have documented how LLMs reproduce only a limited range of attitudinal variation and misrepresent the positions of particular groups and populations as a whole \citep{bisbee2024synthetic, santurkar2023whose}. Recently, \citet{boelaert2025machine} documented a broader pattern of ``machine bias,'' whereby LLMs' responses show unrealistically limited variation across demographic groups and misrepresent human responses in inconsistent ways. Rather than LLMs as a whole being systematically biased toward particular demographic groups, this finding highlights how patterns of (mis-)representation vary across models and topics. This raises the question of how to systematically evaluate such differences across LLMs. This is particularly urgent given that researchers and application developers must choose from among a wide range of models, with distinct representational capabilities, when implementing different types of social simulations.

Model selection is further complicated by proposed methods for improving LLMs' population representativeness. One approach involves fine-tuning language models on empirical data to better reproduce observed response distributions, for tasks including missing data imputation in surveys and predicting responses to unasked questions \citep{kimAIAugmentedSurveysLeveraging2026, suh2025language}. CultureLLM, for example, uses cross-national World Values Survey (WVS) responses for fine-tuning culture-specific LLMs that address foundation models' knowledge deficiencies about cultures and languages underrepresented in training corpora \citep{li2024culturellm}. These approaches often target specific forms of alignment, such as matching a model's responses to a \emph{country-level} response distribution for a set of survey questions \citep{cao2025specializing}. While such fine-tuning can bring model responses closer to those of a target population, it does not necessarily reproduce variation \emph{within} that population and among its demographic subgroups \citep{wightman2026people}. It may even compress or misplace differences among social groups, exacerbating existing sources of misalignment \citep{adilazuarda2025surveys}. Recent work cautions against inferring alignment from marginal response distributions, which may obscure how models misrepresent populations' underlying attitudinal structures \citep{williams2026beyond}. We contribute to this line of research with a framework for assessing how well models recover the magnitude and structure of between-group differences within survey items, properties that measures of item-level agreement and cross-item correlation do not directly assess.

We introduce Population Fidelity, an evaluation framework that distinguishes key conditions required for a set of LLM-generated responses to represent a population. It decomposes population representativeness into three dimensions. \emph{Accuracy} measures model responses' correspondence with each population subgroup's observed response distribution. \emph{Adaptability} measures whether the model reproduces the observed amount of variation between subgroups. \emph{Structure} measures whether the same groups are relatively similar or different in generated and observed responses. The Population Fidelity Score (PFS) combines accuracy, adaptability, and structure, providing a summary measure of model representativeness in relation to a given set of survey responses. Considering the overall score alongside its components and \emph{center alignment} (a separate measure of agreement between model and survey centers) demonstrates how LLMs can achieve some aspects of representativeness despite failing to achieve overall fidelity and how better center alignment need not imply better representation of within-population variation.

\begin{figure}[!h]
\centering
\includegraphics[width=\columnwidth]{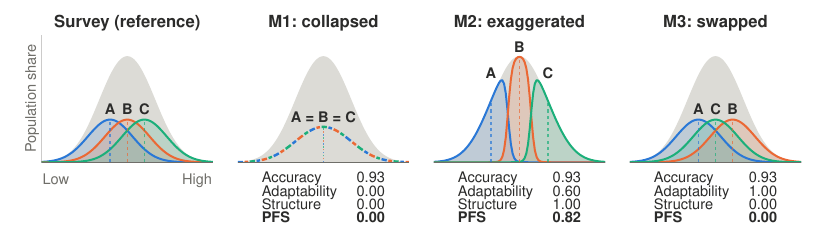}
\caption{Stylized example with three subpopulations (A--C). Colored curves show each group's share of respondents and sum to the pooled distribution (gray), which is identical in every panel ($S_{\mathrm{center}}=1$ for all models); dashed lines mark group means. Calculations are in Appendix~\ref{app:toy-example}.}
\label{fig:pfs-toy}
\end{figure}

Figure~\ref{fig:pfs-toy} illustrates why aggregate agreement alone is insufficient to establish Population Fidelity. Each model-generated distribution appears representative:  pooled across groups, each reproduces the survey's overall response distribution exactly (the gray area in every panel), and the three have nearly identical accuracy. Yet each misrepresents between-group differences: M1 eliminates them by giving every group the same distribution; M2 exaggerates them, making groups appear farther apart than observed; and M3 reproduces the correct amount of difference but assigns it to the wrong groups. Center alignment and accuracy therefore cannot distinguish these models, whereas adaptability and structure can. M1 and M3 receive $\mathrm{PFS}=0$ for different reasons, while M2, which preserves which groups are more alike, is penalized mainly for excess variation.

We demonstrate the framework's utility in two ways. First, we reproduce the results of \citet{boelaert2025machine} from their original replication materials and apply the framework to both their models and newer ones. Their diagnosis centers on too little variation across demographic groups; incorporating structure shows that this is only part of the problem. Across 164 evaluated combinations of model variant, survey question, and elicitation method, median accuracy is approximately $0.83$, whereas median structure is approximately $0.06$ and is the limiting PFS component in 159 cases, and even models that match or exceed the survey's between-group variation rarely reproduce its structure. Second, we evaluate one proposed approach to improving models' population representativeness: cultural fine-tuning. Specifically, we apply CultureLLM-style fine-tuning \citep{li2024culturellm} and compare German and Mexican adapted models with their corresponding as-released models. CultureLLM-style fine-tuning does not improve PFS on average: mean changes are
$-0.024$ for German and $-0.014$ for Mexican fine-tuning. Within Mexico, Mexican fine-tuning improves center alignment by $0.038$ while target-country PFS changes by $-0.010$. These results show that fine-tuning can improve alignment with a target survey center without improving the representation of differences within that population, a distinction that measures of aggregate agreement overlook.

Overall, we argue that representing a population requires models to reproduce several features of human attitudinal variation simultaneously. Our framework organizes these dimensions and we provide reusable code and data for evaluating population fidelity and alignment methods at \url{https://github.com/CriticalMaking/LLM-population-fidelity}.

\section{Related Work}

\subsection{Misalignment in LLM-simulated human attitudes}

Despite its considerable potential \citep{argyle2023out}, recent work has highlighted serious limitations in LLMs' abilities to generate synthetic survey responses that faithfully represent human populations. Although models can reproduce human attitudes at the coarsest levels of analysis, including average responses that are close to survey-based population averages, they fail to reproduce much of the underlying variation within populations, including the true range of positions expressed and their relationships to demographic characteristics \citep{bisbee2024synthetic}. Recent studies document several kinds of misalignment in LLM-generated responses, including bias towards dominant viewpoints and flattening the perspectives of smaller countries and population subgroups \citep{santurkar2023whose, wang2025large}. \citet{boelaert2025machine} identify a broader pattern of ``machine bias'' across a range of models, referring to their tendency to not only misrepresent population groups but to do so inconsistently, favoring different demographic groups in ways that vary arbitrarily across models and topics. These biases constitute potential sources of harm for misrepresented groups \citep{gallegosBiasFairnessLarge2024}, with concerning downstream consequences given the wide-ranging and increasingly consequential scenarios in which LLMs are deployed \citep{chengMarkedPersonasUsing2023, lukkFairFundBenchEvaluatingDistributive2026}. Because models' representational capabilities vary across models and questions, it is crucial to develop evaluations that assess LLMs' performance on several key aspects of population representativeness.

\subsection{Cultural Alignment and Steering Methods}

Researchers have proposed several ways to steer models to improve alignment with cultural and population targets. A variety of studies investigate the effects of techniques applied at inference time, most notably demographic persona-based prompting \citep{argyle2023out, liuEvaluatingLargeLanguage2024}, with evidence that these do not significantly improve model representativeness \citep{santurkar2023whose, bisbee2024synthetic} and may even worsen certain kinds of social biases \citep{deshpandeToxicityChatgptAnalyzing2023, wanArePersonalizedStochastic2023}. An alternative approach involves fine-tuning language models on relevant empirical data to better reproduce real-world responses or predict responses on unasked survey questions \citep{kimAIAugmentedSurveysLeveraging2026}. CultureLLM uses selected World Values Survey questions as seed data to generate semantically equivalent training examples and fine-tune models aligned with particular cultural perspectives, especially those likely to be underrepresented in model training data \citep{li2024culturellm}. \citet{cao2025specializing} instead fine-tune models to reproduce country-level response distributions, while \citet{suh2025language} train on distributions for demographic and ideological subpopulations. These promising approaches encode different notions of what successful population adaptation entails, further motivating a framework that distinguishes the aspects of population representation recovered by different alignment methods.

\subsection{Evaluating Population Representation}

Previous works primarily use single-metric evaluations to compare model responses to those of human survey respondents, including correlation and regression coefficients for individual quantitative outcomes \citep{argyle2023out, bisbee2024synthetic} and distance measures of pairs of opinion distributions \citep{santurkar2023whose, boelaert2025machine}. An exception to this is work by \citet{williams2026beyond}, who, in addition to comparing simulated and observed response distributions, measure whether models accurately reproduce how human attitudes are structured \emph{across} survey questions. In contrast to these approaches, we enumerate multiple conditions required for a set of LLM-generated responses to represent a  population, with an emphasis on faithful variation among social groups \emph{within} a given question. We develop metrics to assess those conditions and propose a summary Population Fidelity Score to enable comparisons across models and survey items.

\section{Machine Bias Design and Reproduction}
\label{sec:machine-bias}
\label{sec:original-machine-bias}
\label{sec:reproduction-protocol}

Our evaluation uses the survey data, demographic cells, prompts, and distance measure of \citet{boelaert2025machine}. Their design compares LLM responses with three waves of the WVS across happiness, political ideology, religious attendance, and social trust. Respondents are grouped by country, survey wave, sex, age, education, employment, and marital status, yielding 687 demographic cells for three questions and 639 for political ideology. Each cell is represented by an interview-style demographic prompt. Responses are elicited either from renormalized next-token probabilities (NTP) or from sampled full answers (FA). Correspondence between survey and model distributions is measured using normalized 
Earth Mover's Distance:
\begin{equation}
\operatorname{nEMD}(p,q) = 
\frac{1}{K-1}
\sum_{k=1}^{K-1}
\left|
\sum_{j=1}^{k}(p_j-q_j)
\right|,
\label{eq:nemd}
\end{equation}
where $p$ and $q$ are distributions over $K$ ordered response options, so that
$\operatorname{nEMD}\in[0,1]$. This metric supplies the distance used throughout
Section~\ref{sec:population-fidelity}.

Before applying our framework, we check that our independent implementation reproduces the original results. Passing the archived generations released by \citet{boelaert2025machine} through our pipeline reproduces every full-precision checkpoint reported for Mixtral-8x7B, along with the published nEMD distributions, regression analyses, and multidimensional-scaling results. We also regenerate Mixtral-8x7B responses from the public weights using the original prompts. This fresh run is equivalent to the archived one under NTP but not fully under FA; the two runs differ in quantization and inference backend as well as sampling. We retain both as reference series, since the archived run preserves comparability with the published results, and the fresh run provides a reference under our current inference setup. Appendix~\ref{app:reproduction} reports the reproduction and equivalence tests in full.

\section{The Population Fidelity Framework}
\label{sec:population-fidelity}

The analyses of \citet{boelaert2025machine} diagnose machine bias at the level of an entire model run. We decompose population fidelity into quantities that can be computed for a complete population or any sufficiently large subset of its subpopulations. The framework distinguishes whether a model approximates particular groups, reproduces the observed amount of variation between groups, and preserves the structure of those differences.

\subsection{Unit of Analysis and Notation}
\label{sec:setting}

Consider a survey question with $K$ ordered response options and a model evaluated on $n$ retained subpopulations, which we call \emph{cells}. Let $i=1,\ldots,n$ index the cells, with $s_i$ denoting the empirical survey-response distribution and $m_i$ the corresponding model-response distribution. Both are probability vectors over the same response categories. Distances are calculated using Equation~\ref{eq:nemd}.

\subsection{Population Fidelity Dimensions}
\label{sec:four}

\par\noindent\textbf{Accuracy.}\ 
Average cell-level error is
$
E
=
\frac{1}{n}
\sum_{i=1}^{n}
\operatorname{nEMD}(s_i,m_i).
\label{eq:accuracy}
$
We define the accuracy score as $S_{\mathrm{acc}}=1-E$. It measures how closely the model approximates each group's observed response distribution and corresponds to the per-cell distances used in the original Machine Bias study.

\par\noindent\textbf{Adaptability.}\ 
Accuracy does not indicate whether the model reproduces variation between social groups. To measure this, we first calculate the median pairwise distances among model and survey cells,
$D_{\mathrm{LLM}}=\operatorname{median}_{i<j}\{\operatorname{nEMD}(m_i,m_j)\}$ and
$D_{\mathrm{WVS}}=\operatorname{median}_{i<j}\{\operatorname{nEMD}(s_i,s_j)\}$.
We define the adaptability ratio as
$
A
=
\frac{D_{\mathrm{LLM}}}{D_{\mathrm{WVS}}}.
\label{eq:adaptability}
$
This measures whether the model produces as much between-group variation as the survey. A value of $A=1$ indicates equal dispersion, while values below and above one indicate compression and amplification, respectively. Because both compression and amplification depart from the survey, we define the symmetric adaptability score as
$S_{\mathrm{adapt}}=\min(A,A^{-1})$ for $A>0$, and set
$S_{\mathrm{adapt}}=0$ when $A=0$. Halving and doubling the survey's
dispersion therefore receive the same score.

\par\noindent\textbf{Structure.}\ 
A model can reproduce the overall amount of variation while assigning it to the wrong groups. We compare the complete vectors of pairwise distances using
$
\rho
=
\operatorname{corr}
\left(
\left\{
\operatorname{nEMD}(s_i,s_j)
\right\}_{i<j},
\left\{
\operatorname{nEMD}(m_i,m_j)
\right\}_{i<j}
\right),
\label{eq:structure}
$
where $\operatorname{corr}$ is Spearman's rank correlation. A value near one indicates that the same pairs of groups are relatively similar or different in the model and survey responses. A value near zero indicates little correspondence between the two structures. The combination $A>1$ and $\rho\approx 0$, which we call \emph{over-steerability without social fidelity}, occurs when demographic conditioning produces strong differences that do not correspond to those observed among human groups. The pairwise-distance vectors contain $\binom{n}{2}$ entries that share cells and are not independent. We therefore treat $\rho$ as a descriptive effect size rather than interpreting its naive significance test \citep{mantel1967detection}. The structure score is $S_{\mathrm{struct}}=\max(0,\rho)$, so a nonpositive association indicates no recovered population structure. We retain the untransformed $\rho$ for diagnostic analyses, particularly when interpreting values near zero.

\subsection{Population Fidelity Score and Center Alignment}
\label{sec:pfs}

\textbf{Population Fidelity Score (PFS)} is the geometric mean of accuracy,
adaptability, and structure:
$
\centering
\mathrm{PFS}
=
\left(
S_{\mathrm{acc}}
\cdot
S_{\mathrm{adapt}}
\cdot
S_{\mathrm{struct}}
\right)^{1/3}. $ We use the geometric mean because all three dimensions are required for population fidelity. A synthetic population must approximate the responses of specific groups, reproduce the observed amount of between-group variation, and assign that variation to the corresponding groups. An arithmetic mean lets strong dimensions offset failure on another. The geometric mean instead makes weak components constrain the overall score, as in multidimensional indices such as the Human Development Index \citep{klugman2011hdi}.

\par\noindent\textbf{Center alignment ($S_{center}$).}\ 
Center alignment is reported alongside the PFS but is not included in it. For a set of retained cells $I$, define the survey center as $\bar{s}_I = \frac{1}{|I|}\sum_{i\in I}s_i,$ and the model center as
$
\bar{m}_I = \frac{1}{|I|}\sum_{i\in I}m_i.
$
The distance between them is
$
C_I = \mathrm{nEMD}(\bar{s}_I,\bar{m}_I),
$ and $
S_{\mathrm{center}} = 1-C_I.
$
We report center alignment over all retained cells and, for cultural fine-tuning analyses, over retained cells in the target country. Because the centers weight retained cells equally, center alignment measures agreement between the model and survey centers rather than a respondent-weighted population mean. A model can match this center while remaining nearly invariant across demographic groups. We therefore report center alignment separately from the PFS since a well-aligned center does not compensate for missing variation or structure.

\subsection{Population Fidelity within Demographic Subgroups}
\label{sec:groups}

Each quantity can be recomputed within a subset of cells. For a demographic group $g$ with cell set $\mathcal{I}_g$, we calculate accuracy, adaptability, structure, center alignment, and PFS using only cells in $\mathcal{I}_g$. We consider levels within seven demographic categories: country, survey wave, sex, age, education, employment, and marital status. For each subgroup, we recompute all metrics using only the cells belonging to that subgroup, rather than averaging population-level scores. This allows the framework to identify population segments for which a model or intervention is more or less faithful.

\section{Experimental Design}
\label{sec:experimental-design}
\label{sec:experimental-overview}

\subsection{Models and Conditions}
\label{sec:LLMs}

Appendix \ref{appendixExpPipe} provides an overview of the experimental pipeline. We evaluate six open-weight models spanning approximately 2B--31B parameters: Gemma-4-31B-it, Gemma-4-E4B-it, Qwen3-VL-8B-Thinking, Qwen3-VL-2B-Thinking, Llama-3.2-3B, and Muse-Glimmer-30B. Each is evaluated as released and after cultural fine-tuning toward Germany and Mexico.  We also evaluate OpenAI's \emph{GPT-5.6 Terra} (Terra), a proprietary model accessed through its API. Terra is evaluated as released under FA elicitation only because its endpoint does not provide usable NTP at nonzero reasoning effort. Mixtral-8x7B is included through two reproduction references: archived generations from the original Machine Bias study and responses regenerated from the public model weights.

The design contains 84 model--condition--question series: 80 from 20 conditions evaluated on four questions, plus four Terra series. The non-Terra series are evaluated under both NTP and FA, yielding 160 run--mode combinations; Terra is FA-only, for a total of 164 combinations.

\subsection{Cultural Fine-Tuning}
\label{sec:finetuning}

In this study, we follow the approach taken by CultureLLM \citep{li2024culturellm}. For each target culture, the training data combine CultureLLM's selected WVS seed questions with semantically augmented paraphrases and culture-specific target answers derived from aggregate survey responses.

We use CultureLLM's German condition and construct a Mexican condition. CultureLLM's original Spanish partition pools respondents from Mexico and Argentina, whereas our reference population is Mexico. We therefore reconstruct the training records using only Mexican WVS responses. We denote the resulting German and Mexican fine-tuned conditions by $\mathcal{F}_{\text{de}}$ and $\mathcal{F}_{\text{mx}}$, respectively.

Each variant is implemented as a low-rank update to its corresponding as-released model using LoRA with rank $r=8$ and scaling parameter $\alpha=16$ \citep{hu2021lora}. We use QLoRA for the two largest models \citep{dettmers2023qlora}. The German and Mexican variants of a model use the same training and inference configuration. None of the four evaluation items---happiness, political ideology, religious attendance, or social trust---appears in the fine-tuning data.

\subsection{Elicitation and Evaluation}
\label{sec:capacity}
\label{sec:evaluation-design}

We use the interview-style prompts (Section~\ref{sec:original-machine-bias}) under both elicitation modes, NTP and FA; FA responses are sampled at temperature 0.7 and parsed against the offered options. A 32-prompt capacity probe precedes each run and flags an NTP run as non-informative if its mean probability mass on valid answer tokens is below $0.10$; none is flagged (Appendix~\ref{app:capacity}). A cell enters the
analysis only with at least 20 valid FA responses. NTP is deterministic under our decoding configuration and is run once. Because FA samples stochastically, we run each of the 18 locally evaluated conditions three times. For consistency with conditions available for only one FA run, primary results use the first FA run; figures show three-run means and 95\% CIs where repeats are available (Appendix~\ref{app:replicates}).

We compute every metric at three scopes: population-wide, using all retained cells; target-country, using cells from the fine-tuning target country (German for $\mathcal{F}{_\text{de}}$, Mexican for $\mathcal{F}{_\text{mx}}$); and subgroup-specific, recomputing the metric within each level of country, survey wave, sex, age, education, employment, and marital status (Section~\ref{sec:groups}). We report fine-tuning effects at both population-wide and target-country scopes.

Each fine-tuned condition is compared only with its corresponding as-released model. Under FA elicitation, the set of retained cells can differ across conditions because a cell must contain at least 20 valid responses for that condition. To ensure that paired comparisons are based on the same population, each fine-tuned condition and its as-released baseline are scored on the intersection of the cells retained by both. Thus, each paired difference reflects changes in model responses on the same cells, rather than differences in cell composition. With six models, four questions, and two elicitation modes, this yields 48 paired comparisons per cultural target.

All primary analyses are conducted separately by question; averages across the four questions are reported only as descriptive summaries. Appendix~\ref{app:metrics} details metric implementation and retention; Appendix~\ref{app:capacity} provides details on experimental setup, fine-tuning, response coverage, and prompts.

\section{Results}
\label{sec:results}

\subsection{LLMs Approximate the Survey Center but Miss Group Structure}

Models often achieve high cell-level accuracy but rarely recover the structure of differences among cells. To contextualize accuracy, we use a survey-center baseline that ignores
demographic information (Appendix~\ref{app:survey-center-baseline}), predicting the equal-cell survey center $\bar{s}_I$ for every cell. This baseline attains $S_{\mathrm{acc}}\in[0.848,0.915]$ across questions and matches or exceeds model accuracy in 157 of 164 run--mode combinations; only seven combinations outperform it. Thus, much of the observed accuracy reflects proximity to the survey center rather than recovery of demographic differences.

Across the 164 combinations, median $S_{\mathrm{acc}}=0.832$ whereas median $S_{\mathrm{struct}}=0.060$ (Appendix~\ref{app:sensitivity}, Table~\ref{tab:sensitivity}, default row). $S_{\mathrm{struct}}$ is the limiting component in 159 combinations, $S_{\mathrm{adapt}}$ in five, and $S_{\mathrm{acc}}$ in none. The survey-center baseline itself has $S_{\mathrm{center}}=1$ but $A=0$, and therefore $\mathrm{PFS}=0$, illustrating why center agreement alone is insufficient. Another 30 combinations have $\mathrm{PFS}=0$ because $\rho\leq0$; their narrow range, $\rho\in[-0.026,0]$, indicates absent rather than systematically reversed population structure.

\begin{figure}[tb]
\centering
\begin{subfigure}[t]{\columnwidth}
\centering
\includegraphics[width=\linewidth]
{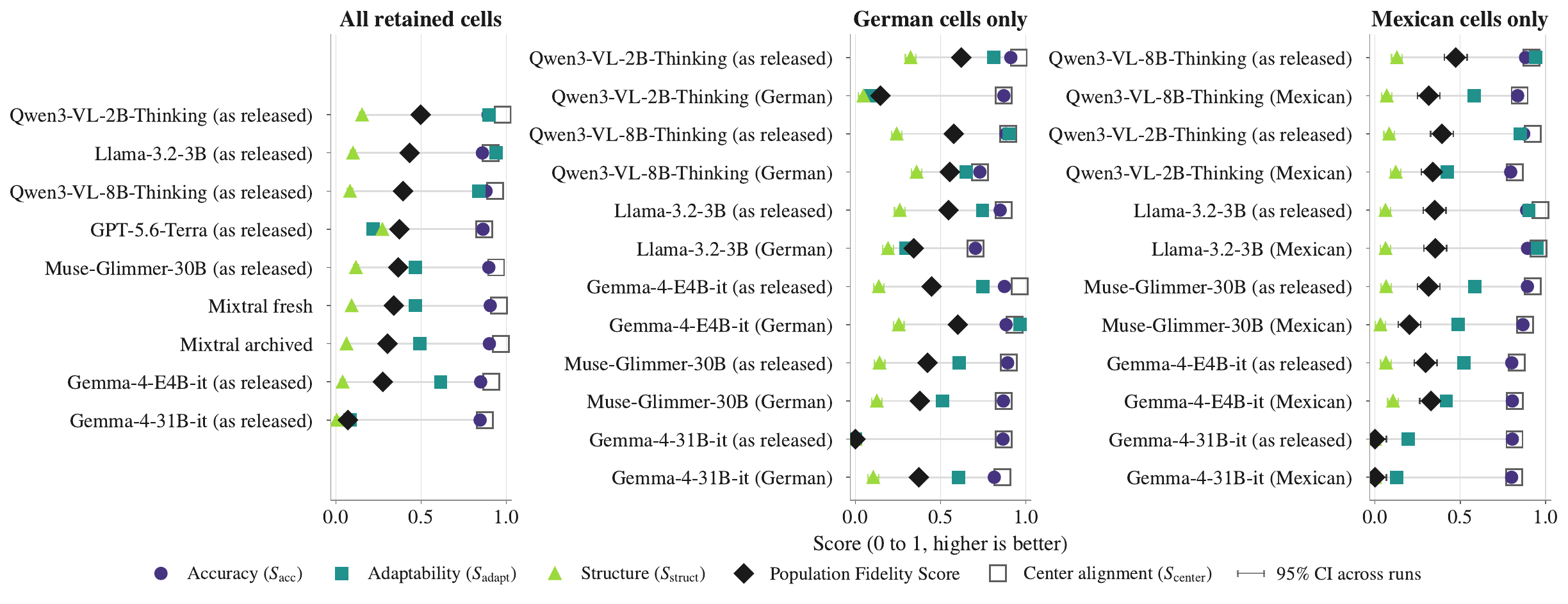}
\caption{Component profile.}
\label{fig:components}
\end{subfigure}

\medskip
\begin{subfigure}[t]{0.5\columnwidth}
\centering
\includegraphics[width=\linewidth]
{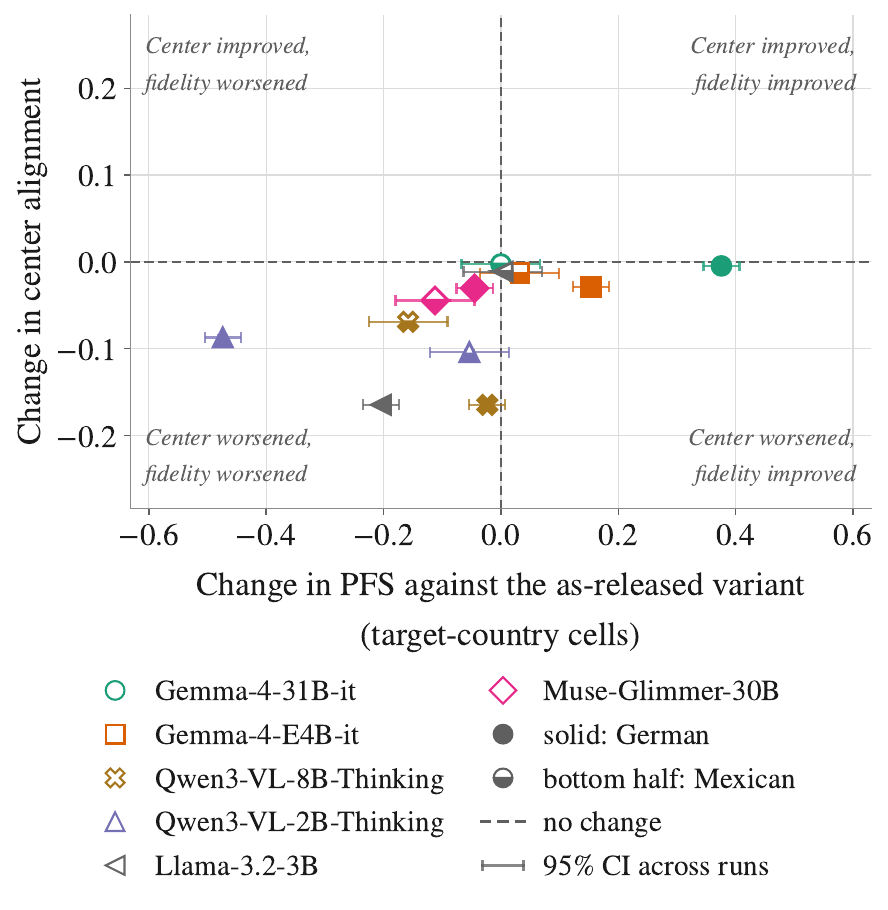}
\caption{Fine-tuning shifts.}
\label{fig:shift}
\end{subfigure}
\caption{
Population Fidelity for happiness under FA elicitation. \textbf{(a)} Components, PFS, and center alignment for the indicated cell sets. \textbf{(b)} Fine-tuned minus as-released changes on target-country cells. Markers show three-run FA means using each run's paired support; Table~\ref{tab:replicates-paired}, Appendix~\ref{app:replicates}, uses the stricter all-runs common-cell intersection. All-cell results are in Figure~\ref{fig:happiness-all-app}, Appendix \ref{app:figures}.
}
\label{fig:happiness-results}
\end{figure}

Figure~\ref{fig:components} shows this profile for happiness under FA elicitation: accuracy is high for every condition, whereas structure is the weakest component for almost all. The other questions show the same pattern (Appendix~\ref{app:figures}). 
This parallels the machine bias pattern of \citet{boelaert2025machine}, in which errors are largely explained by each cell's distance from the model's average answer, and a model that barely varies is accurate only insofar as its center sits near the survey cells. Population Fidelity complements this diagnosis by showing that such accuracy can coexist with weak recovery of the observed structure of between-group differences.

\subsection{More Demographic Variation Does Not Imply Better Group Structure}\label{appMoreDemog}

\citet{boelaert2025machine} identify low adaptability as the core of machine bias. Our ratio $A$ formalizes their comparison of median pairwise distances among model and survey cells (e.g., $0.033/0.115\approx0.29$ for Mixtral-8x7B on happiness); structure asks whether the variation a model does produce falls between the right groups. Among the 52 as-released run--mode combinations excluding the Mixtral reference series, 43 compress variation relative to the survey ($A<1$), whereas nine amplify it ($A>1$). Compression occurs for every combination on happiness and political ideology, while amplification appears only for religious attendance and social trust (Table~\ref{tab:adaptability-counts}, Appendix~\ref{app:adaptability-results}). Yet greater variation rarely corresponds to better group structure: only one of the nine amplifying combinations reaches $\rho>0.1$ (Figure~\ref{fig:structure-app}, Appendix \ref{app:figures}, plots $A$ against $\rho$ under NTP). Thus, models can respond differently across demographic profiles without reproducing which groups are relatively similar or different in the survey. Adaptability and structure therefore capture distinct requirements of population fidelity: matching the amount of variation is not sufficient if that variation is distributed across the wrong groups.

\subsection{CultureLLM-Style Fine-Tuning Shows No Consistent Fidelity Gains}
\label{sec:finetuning-results}

In the primary analysis, CultureLLM-style fine-tuning does not improve PFS on average in either scope. Here, $\Delta\mathrm{PFS}$ denotes fine-tuned minus as-released PFS, so positive values indicate improved population fidelity. Population-wide, effects are mixed: $\mathcal{F}_{\text{de}}$ raises PFS in 21 of 48 comparisons and lowers it in 23 (mean $\Delta\mathrm{PFS}=-0.024$), while $\mathcal{F}_{\text{mx}}$ raises it in 19 and lowers it in 23 (mean $-0.014$). Target-country means are also negative: $-0.021$ in Germany and $-0.010$ in Mexico. Thus, the aggregate result reflects heterogeneous effects rather than a consistent benefit from fine-tuning. Model- and question-specific results are reported in Appendix~\ref{app:model-results}.

The repeated FA runs support the same conclusion but show greater sampling sensitivity for PFS than center alignment. Across runs, mean $\Delta\mathrm{PFS}$ on all retained cells and mean $\Delta S_{\mathrm{center}}$ in both scopes vary by at most $0.005$. Within target countries, pooled 95\% CI half-widths for individual-comparison $\Delta\mathrm{PFS}$ are $0.073$ in Germany and $0.110$ in Mexico, versus $0.005$ and $0.003$ for $\Delta S_{\mathrm{center}}$, respectively (Appendix~\ref{app:replicates}).

Changes in center alignment do not necessarily track changes in PFS. Figure~\ref{fig:shift} shows this for happiness within Germany: Gemma-4-31B-it's $\mathcal{F}_{\text{de}}$ increases target-country PFS by $0.371$ while barely changing center alignment ($\Delta S_{\mathrm{center}}=-0.004$), whereas Qwen3-VL-8B's $\mathcal{F}_{\text{de}}$ decreases both
($\Delta\mathrm{PFS}=-0.015$, $\Delta S_{\mathrm{center}}=-0.164$). The same appears in the cross-question FA summary. Across models and questions, $\mathcal{F}_{\text{mx}}$ moves responses toward the Mexican survey center ($\Delta S_{\mathrm{center}}=+0.037$) while target-country PFS changes little ($\Delta\mathrm{PFS}=-0.002$); $\mathcal{F}_{\text{de}}$ moves responses away from the German center ($\Delta S_{\mathrm{center}}=-0.087$) while target-country PFS changes by $-0.017$. 

The components clarify this pattern. Within the target countries, fine-tuning reduces adaptability: median $S_{\mathrm{adapt}}$ falls from $0.754$ to $0.496$ in Germany and from $0.616$ to $0.421$ in Mexico. Structure changes little ($\Delta S_{\mathrm{struct}}=0.000$ in Germany and $-0.004$ in Mexico) and remains the limiting PFS component in 44 of 48 $\mathcal{F}_{\text{de}}$ and 47 of 48 $\mathcal{F}_{\text{mx}}$ comparisons. Thus, fine-tuning can shift average responses and alter between-group variation without recovering the observed pattern of group differences. PFS captures these distinctions across fine-tuning objectives. Appendix~\ref{app:distribution-matched} illustrates the same diagnostics for CultureLLM-style, distribution-matched, and subgroup-matched fine-tuning on a common backbone. Because these conditions differ in training data, population targets, and evaluation-item overlap, the comparison is diagnostic rather than controlled. On Qwen3-VL-2B-Thinking, the distribution-matched condition retains higher adaptability and structure than the CultureLLM-style variants, whereas the subgroup-matched condition remains structure-limited.

\subsection{Center Alignment Can Mask Poor Fidelity within Demographic Subgroups}

\begin{figure}[t]
\centering
\includegraphics[width=\linewidth]
{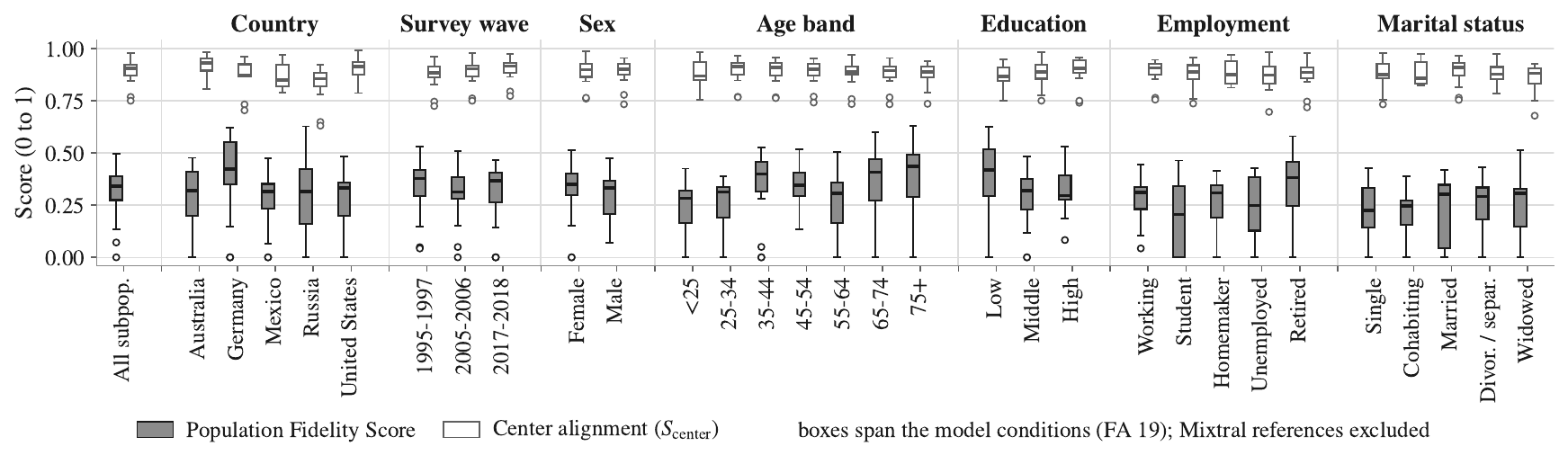}
\caption{
PFS and center alignment by demographic level for happiness under FA elicitation. Boxes summarize 19 model conditions. Results for all questions: Figure~\ref{fig:subgroups-app}.
}
\label{fig:subgroups-main}
\end{figure}

Within demographic levels, subgroup centers are generally well matched while PFS varies widely, so a matched center is weak evidence of fidelity inside the group. For happiness under FA elicitation, median $S_{\mathrm{center}}$ across demographic levels ranges from $0.848$ to $0.928$ (Figure~\ref{fig:subgroups-main}). Across model conditions, PFS and $S_{\mathrm{center}}$ are only moderately correlated within demographic levels, with mean Spearman correlations ranging from $0.263$ to $0.425$ (Appendix~\ref{app:demographic-results}).

The imbalance persists across demographic levels, and fine-tuning does not repair it. Across the 4{,}440 subgroup-level scores (148 run--mode combinations by 30 levels, Mixtral references excluded), $S_{\mathrm{struct}}$ is limiting in 4{,}097 and $S_{\mathrm{adapt}}$ in 343, while $S_{\mathrm{acc}}$ never is; PFS varies much more across age and employment than across sex: the mean within-condition range is $0.287$ for age and $0.270$ for employment, versus $0.054$ for sex, with these between-level differences driven more by structure than by adaptability (Table~\ref{tab:families-app}, Appendix \ref{app:demographic-results}). German and Mexican fine-tuning improve PFS in only $42.5\%$ and $40.8\%$ of subgroup comparisons, respectively; within Germany and Mexico, PFS rises in only 18 of 48 country-level comparisons for each target (Figure~\ref{fig:country-shift}, Appendix \ref{app:demographic-results}).

\subsection{Robustness Across Elicitation, Sampling, and Metric Specifications}

The main patterns are broadly consistent across elicitation modes. Among the 20 conditions evaluated under both NTP and FA, PFS rank correlations range from $0.695$ for political ideology to $0.880$ for religious attendance. Agreement is generally higher for $S_{\mathrm{acc}}$ and $S_{\mathrm{center}}$, while $S_{\mathrm{adapt}}$ is more sensitive to elicitation, with correlations from $0.502$ to $0.913$ (Table~\ref{tab:mode-agreement}, Appendix \ref{app:mode-agreement}). Fine-tuning effects are also directionally similar: NTP and FA agree on the sign of $\Delta\mathrm{PFS}$ in 38 of 48 comparisons, or 35 of 41 after excluding cases with zero change under either mode (Appendix~\ref{app:mode-agreement}).

The conclusions are also stable across alternative metric specifications.
Appendix~\ref{app:sensitivity} varies the dispersion statistic, adaptability transformation, structure correlation and clipping, aggregation rule, and cell weighting. Across these cases, $S_{\mathrm{struct}}$ remains the limiting component in at least 156 of 164 combinations, and neither CultureLLM-style target improves population-wide PFS on average under any specification. Replacing the geometric mean with an arithmetic mean changes PFS more because it allows stronger components to offset weak structure. Mean PFS rises from $0.243$ to $0.451$ and no longer reaches zero, illustrating why the geometric mean is better aligned with our requirement that all three dimensions be jointly satisfied.

\section{Discussion, Limitations, and Conclusion}
\label{sec:discussion-conclusion}

Overall, our work extends existing efforts to understand the extent to which LLMs are capable of simulating known patterns of human attitudes and beliefs. While early work featured forms of social or demographic bias, in which one group's attitudes might be systematically favored over others, more recent research has found that LLMs themselves exhibit their own characteristic biases. Most notably, they tend to reduce variation across demographic groups. This paper builds upon such work to formulate a more general framework for evaluating the \textit{population fidelity} of LLM-generated survey response patterns. 

In general, we find that LLM-generated responses do approximate the centers of human survey responses. However, they do a poor job of recovering differences among groups, a task which is often central to social and cultural research. In other words, weak fidelity to the typical response patterns we observe among human survey respondents typically occurs not simply from recovering too little demographic variation, but from failing to reproduce which groups respond in ways that are relatively similar or different to each other. Our Population Fidelity metric makes this distinction explicit by separating fidelity into three components: group-level accuracy, the amount of between-group variation, and its structure. In our experiments, structure is usually the limiting dimension, also under the extra adaptation objectives evaluated in Appendix~\ref{app:distribution-matched}.

Our analysis is limited in a number of ways, which future work can seek to address. First, the scope of the data sources could be expanded. We have evaluated just one  dataset, the WVS, and from that focused on a limited subset: three waves, five countries, four attitudes, and a set of relatively coarse demographic cells. Second, even though respondents are international, we have used English prompts and fine-tuning data. Likewise, our examination of fine-tuning approaches evaluated six models and two cultural targets, whereas the two additional fine-tuning methods we pursued (SimLLCultureDist and SubPOP) were tested on only one backbone. While these experiments demonstrate the applicability of PFS across adaptation objectives, testing across more objectives and sources and under additional conditions could help to understand the promise and limits of the framework. Finally, the method used to elicit responses also matters. Whether a model appears to compress or amplify differences between demographic groups depends partly on whether responses are derived from next-token probabilities (NTP) or sampled as full answers (FA). Adaptability is more sensitive to this choice than the other components of Population Fidelity, so compression and amplification should generally be compared within the same elicitation mode. Further tests that systematically vary elicitation methods would help to understand these dynamics more fully.

Despite these limitations, this paper has established that Population Fidelity provides a common framework for asking not only whether an intervention improves alignment with a target population, but which aspects of population representation it does or does not recover. Across the evaluations, models rarely recover the structure of group differences observed in human populations; evaluations based on agreement with the survey center tend to obscure such misrepresentations. Additional experiments have illustrated how the same framework can distinguish the effects of different fine-tuning objectives. Future work should extend these tests to broader populations, languages, surveys, and model families; scale distribution-based methods beyond a single backbone; and examine richer representations of cultural and individual values, including datasets designed to capture heterogeneous and potentially conflicting values within the same population, as well as established frameworks that represent systems of values rather than isolated survey attitudes. More generally, these extensions would allow researchers to ask not simply whether a model resembles a population on average, but whether it reproduces the ways in which that population is internally differentiated.

\newpage

\subsection*{AI Use Statement}

Generative AI tools were used during the preparation of this work to help
improve the clarity and organization of the manuscript, to support
brainstorming and the organization of ideas, and to assist with software
development and debugging. AI-assisted text and code were reviewed and,
where relevant, revised and tested by the authors. The authors take full
responsibility for the final manuscript, analyses, code, and reported
results.

\subsection*{Ethics Statement}

This work evaluates the extent to which LLM-generated survey responses
represent patterns observed in human survey data. The analysis uses the
World Values Survey through the demographic cells and survey items defined by \citet{boelaert2025machine}
and does not involve the recruitment of new
human participants. Because demographic conditioning and population
simulation can reproduce or amplify stereotypes, results for demographic
groups should not be interpreted as claims about individual members of
those groups. Population Fidelity is intended as an evaluation framework
for identifying discrepancies between model-generated and observed
population patterns, not as evidence that demographic categories fully
characterize the attitudes or behavior of the populations they describe.

\subsection*{Reproducibility Statement}

We provide the information needed to reproduce the experiments in the main
paper and appendices, including the survey and model evaluation procedure,
Population Fidelity metrics, model and fine-tuning configurations,
elicitation settings, retention criteria, reproduction checks, and
sensitivity analyses. The accompanying repositories contain the
code, prompts, configuration files, run manifests, and scripts used to
produce the reported results and figures. One repository contains the
evaluation of the as-released models and the Machine Bias
reproduction,\footnote{\url{https://github.com/CriticalMaking/LLM-population-fidelity}}
while the second contains the cultural fine-tuning experiments and their
evaluation.\footnote{\url{https://github.com/neemiasbsilva/CultureFinetuneSteerMLLM}}
Appendix~\ref{app:reproduction} documents the Machine Bias reproduction;
Appendix~\ref{app:framework-details} describes the Population Fidelity
framework, metric implementation, and toy example;
Appendix~\ref{app:experimental-details} provides the experimental
configuration and prompts; Appendix~\ref{app:supplementary} reports
supplementary results; Appendix~\ref{app:robustness} presents robustness
and sensitivity analyses; and Appendix~\ref{app:distribution-matched}
evaluates additional adaptation objectives.

\subsubsection*{Acknowledgments}
This study was supported in part by the National Council for Scientific and Technological Development - CNPq (processes 441444/2023-7, 403646/2026-0, and 444724/2024-9) and INCT TILD-IAR (proc. 408490/2024-1). This research was also supported in part by a CIFAR AI Catalyst Grant, ``Towards Socially Grounded AI Safety: Integrating Causal and Institutional Reasoning in Language Models'' (PI: Matt Ratto; Co-PI: Zhijing Jin, University of Toronto).

\bibliographystyle{iclr2027_conference}
\bibliography{references}

\clearpage
\appendix

\section*{Appendix}

\section{Machine Bias Design and Reproduction Details}
\label{app:reproduction}

This appendix describes the original Machine Bias design, our two-stage
reproduction, and the equivalence tests underlying Section~\ref{sec:results}. Files obtained from the original replication
package are recorded in a SHA-256 manifest that is checked before each
run. The manifest, code, configurations, and complete reproduction
outputs are provided in the accompanying repository.

\subsection{Original Design}

\citet{boelaert2025machine} compare LLM responses with three WVS
waves---1995--1997, 2005--2006, and 2017--2018---using four
attitudinal items: happiness (4 ordered options), political ideology
(10), religious attendance (7), and social trust (2). Within each
country--wave, respondents are stratified by sex, age, education,
employment, and marital status. Rare demographic profiles are merged
until each retained group contains at least 20 respondents. This
procedure yields 687 cells for happiness, religious attendance, and
social trust and 639 for political ideology, which was not administered
in one Russian wave.

Each cell is represented by an interview-style prompt containing its
demographic profile. Under next-token probability (NTP) elicitation,
the probability mass assigned to valid answer tokens is extracted and
renormalized, producing a deterministic response distribution. Under
full-answer (FA) elicitation, answers are sampled at temperature 0.7
and parsed using a strict response format. Complete questions and
prompt templates appear in Appendix~\ref{app:prompts}.

Cell-level correspondence is measured using the normalized Earth
Mover's Distance in Equation~\ref{eq:nemd}. The measure lies in
$[0,1]$ and penalizes disagreement between adjacent response options
less than disagreement between distant options.
\citet{boelaert2025machine} use thresholds of 0.05, 0.10, 0.15, and
0.30 to define quality bands from ``very good'' to ``very bad.'' They
also compare LLM performance with a leave-one-out demographic linear
model and random permutations of the survey responses.

The original study identifies machine bias through three findings.
First, most model--cell predictions fall into the
mediocre-or-worse nEMD bands, performing worse than the demographic
baseline and only slightly better than the random baseline. Second,
multidimensional scaling shows that model cells occupy a narrower
response space than survey cells. For Mixtral-8x7B on happiness, the
median pairwise nEMD is 0.033 among model cells and 0.115 among WVS
cells. Third, a cell's distance from the model's average response
explains much of its prediction error, with $R^2$ values reaching
0.93--0.95. This factor contributes more than the demographic
characteristics in 11 of the 12 model--question combinations.

\subsection{Archived Reproduction}\label{appendixArchivedRepro}

No model inference is performed during the archived reproduction. We
pass the released generations through our independently implemented
processing and analysis pipeline and compare the resulting counts,
distances, regressions, tables, and figures with the published outputs.
The full-precision checkpoints available for Mixtral-8x7B on happiness
are reproduced exactly in Table~\ref{tab:checkpoints}.

\begin{table}[htb]
\centering
\tiny
\begin{adjustbox}{max width=0.65\textwidth}
\begin{tabular}{lr}
\toprule
Checkpoint & Value \\
\midrule
WVS observations & 26{,}981 \\
Unique NTP profiles & 13{,}904 \\
Happiness cells & 687 \\
Mean valid NTP token mass & 0.9845 \\
Overall NTP nEMD & 0.0308 \\
Overall FA nEMD & 0.0354 \\
Median pairwise WVS nEMD & 0.1146 \\
Median pairwise NTP nEMD & 0.0327 \\
\bottomrule
\end{tabular}
\end{adjustbox}
\caption{
Our reproduction matches every full-precision checkpoint reported by \citet{boelaert2025machine} for Mixtral-8x7B on happiness. Figure~\ref{fig:repro-density-app} shows a representative reproduced figure; complete reproduction outputs are provided in the repository.
}
\label{tab:checkpoints}
\end{table}

\begin{figure}[t]
\centering
\includegraphics[width=0.98\textwidth]
{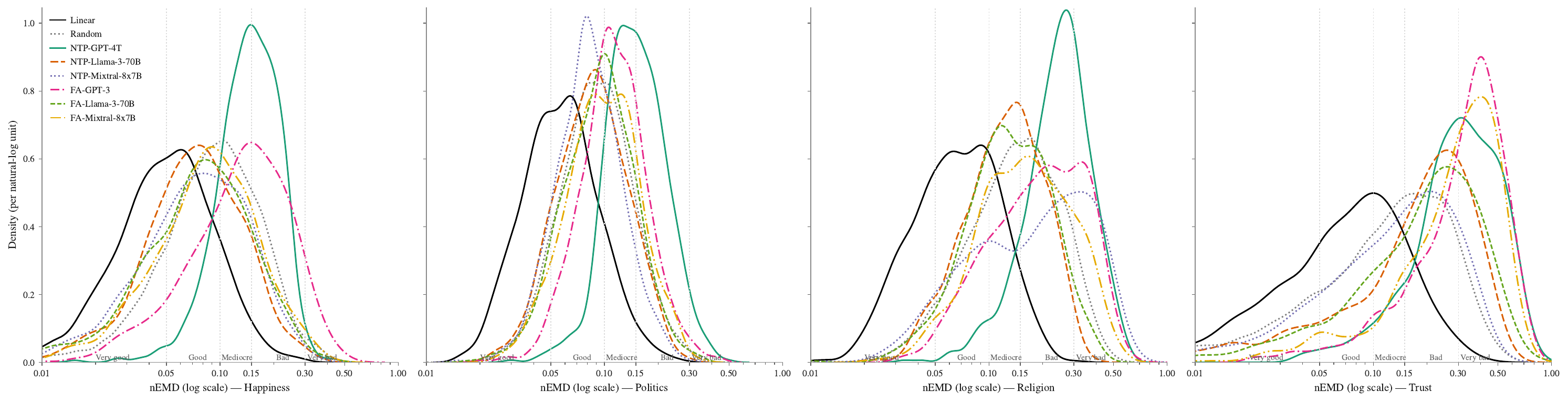}
\caption{Reproduction of Figure~2 from
\citet{boelaert2025machine}: cell-level nEMD densities for the six
published model--elicitation series, shown with the demographic and
random baselines and the original response-quality bands.}
\label{fig:repro-density-app}
\end{figure}

The pipeline also reproduces Figures~2--4 and~6 from
\citet{boelaert2025machine}, including the nEMD distributions,
social-coefficient regressions, multidimensional-scaling geometry, and
centroid regressions. We additionally recover the supplementary
analyses of prompting strategy, temperature, and backtranslation for
all six published model--elicitation series: GPT-4T, Llama-3-70B, and
Mixtral-8x7B under NTP, and GPT-3, Llama-3-70B, and Mixtral-8x7B under
FA. Figure~\ref{fig:repro-density-app} presents one representative
comparison; the complete reproduced outputs are available in the
repository.

\subsection{Fresh Reproduction}

We regenerate Mixtral-8x7B responses using the original single-token,
top-1000-logprob NTP procedure and FA sampling at temperature 0.7 with
a 12-token generation budget. Each prompt produces an atomic record
containing its hash, sampling parameters, backend version, and all FA
generation attempts.

The fresh run uses a Q4\_K\_M GGUF quantization of
Mixtral-8x7B-v0.1\footnote{
The exact quantized artifact, repository revision, and checksum are pinned
in the accompanying code and verified before each run.
}.
Inference uses \texttt{llama-cpp-python} 0.3.1 with CUDA and full GPU
offloading. The run was executed on a single NVIDIA GeForce RTX 5090
(32~GB).

Archived and fresh runs are compared using difference and equivalence
tests with complete country--wave blocks as the resampling unit. The
equivalence margin is $\pm 0.005$ nEMD, one tenth of the narrowest quality
band used in the original study. Difference-test $p$-values are obtained
from cluster sign-flip permutations and adjusted within each question
across the two elicitation modes using the Holm procedure; equivalence
tests are reported without multiplicity adjustment.

Under NTP, the archived and fresh runs are equivalent within the
prespecified margin on all four questions. For happiness, overall nEMD
changes by 2.9\% relative to the archived run. Under FA, equivalence is
inconclusive for happiness and political ideology. Difference tests detect
changes for religious attendance and social trust: mean nEMD changes by
$+0.052$ and $-0.135$, with Holm-adjusted $p$-values of $0.010$ and
$0.0002$, respectively.

We retain both Mixtral series in subsequent analyses. The archived series preserves comparability with the original study, while
the fresh series provides a reproduction reference under the current
quantized checkpoint and inference runtime. Because the fresh run differs
from the archived run in both quantization and backend as well as stochastic
sampling, archived--fresh differences should not be interpreted as estimates
of sampling variability alone.

\subsection{Additional Reproducibility Controls}

For comparability with the original study, each run reproduces its
centroid regression on the German cells. A fixed holdout of 20 cells
estimates the model center, avoiding use of the same observations to
define the center and fit the regression. For the remaining cells,
$\log(1+\operatorname{nEMD})$ error is regressed on
$\log(1+\operatorname{nEMD})$ distance from that center. Holdout
indices reproduce the original R random-number stream using
Mersenne--Twister with seed $20{,}240{,}110$, with stream positions
matched by elicitation mode.

FA generation uses a deterministic seed derived separately for each
prompt, so interrupting and resuming a run does not change previously
written responses. A compatibility flag restores the original
unseeded sampling configuration when needed. Unit tests compare the
prompt bytes for all four questions and both elicitation modes with
those produced by the original implementation.

\section{Population Fidelity Framework and Metric Details}\label{app:framework-details}

\subsection{Framework Overview}\label{appendixExpPipe}

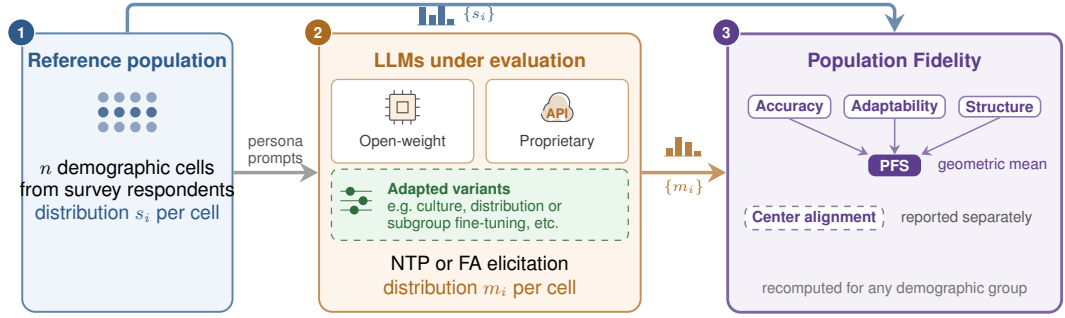
\begin{figure}[thb]
\centering
\tikzset{
  card/.style={rounded corners=6pt, line width=0.9pt, inner sep=0pt, anchor=north west},
  ctitle/.style={font=\small\sffamily\bfseries, anchor=north, align=center},
  cbody/.style={font=\footnotesize\sffamily, anchor=north, align=center, inner sep=0pt},
  badge/.style={circle, minimum size=4.4mm, inner sep=0pt, font=\scriptsize\sffamily\bfseries, text=white, draw=white, line width=0.6pt},
  chip/.style={rounded corners=3pt, line width=0.7pt, font=\scriptsize\sffamily\bfseries, inner xsep=3.2pt, inner ysep=2pt},
  sub/.style={rounded corners=3pt, line width=0.6pt, draw=ModelEdge!60, fill=white, inner sep=0pt, anchor=north},
  subtxt/.style={font=\scriptsize\sffamily, align=center, anchor=north, inner sep=0pt, text=black!75},
  flow/.style={-{Stealth[length=3mm,width=3.2mm]}, line width=1.6pt, draw=black!38},
  flab/.style={font=\scriptsize\sffamily, align=center, inner sep=1pt, text=black!65},
  icondraw/.style={line width=0.55pt, draw=black!65},
}
\newcommand{\chipicon}[1]{\begin{scope}[shift={#1}, scale=0.8]
\draw[icondraw, fill=ModelFill, rounded corners=0.6pt] (-2.6mm,-2.6mm) rectangle (2.6mm,2.6mm);
\draw[icondraw] (-1.1mm,-1.1mm) rectangle (1.1mm,1.1mm);
\foreach \p in {-1.7mm,0mm,1.7mm} {
  \draw[icondraw] (\p,2.6mm) -- (\p,3.5mm); \draw[icondraw] (\p,-2.6mm) -- (\p,-3.5mm);
  \draw[icondraw] (2.6mm,\p) -- (3.5mm,\p); \draw[icondraw] (-2.6mm,\p) -- (-3.5mm,\p);}
\end{scope}}
\newcommand{\cloudicon}[1]{\begin{scope}[shift={#1}, scale=0.8]
\fill[ModelEdge!25] (-1.6mm,-1.4mm) circle (1.6mm) (1.6mm,-1.4mm) circle (1.6mm) (0mm,0.2mm) circle (2.1mm) (-1.6mm,-3mm) rectangle (1.6mm,-1.4mm);
\draw[icondraw] (-3.2mm,-1.4mm) arc[start angle=180, end angle=270, radius=1.6mm] -- (1.6mm,-3mm) arc[start angle=270, end angle=360, radius=1.6mm] arc[start angle=0, end angle=60, radius=1.6mm] arc[start angle=-20, end angle=180, radius=2.1mm] arc[start angle=110, end angle=180, radius=1.6mm];
\node[font=\tiny\sffamily\bfseries, text=ModelEdge] at (0,-1.4mm) {API};
\end{scope}}
\newcommand{\slidersicon}[1]{\begin{scope}[shift={#1}, scale=0.8]
\foreach \y/\x in {1.9mm/-1.2mm, 0mm/1.4mm, -1.9mm/-0.4mm}{
  \draw[icondraw, draw=CultureEdge!80] (-3.2mm,\y) -- (3.2mm,\y);
  \fill[CultureEdge] (\x,\y) circle (0.9mm);}
\end{scope}}
\newcommand{\histo}[3]{\begin{scope}[shift={#1}]
\draw[line width=0.4pt, draw=#2!60] (0,0) -- (6.2mm,0);
\foreach \h [count=\k from 0] in #3 {\fill[#2!85] ({0.4mm+\k*1.5mm},0) rectangle ({1.5mm+\k*1.5mm},\h);}
\end{scope}}
\newcommand{\cellgrid}[2]{\begin{scope}[shift={#1}]
\foreach \r/\t in {0/55,1/85,2/55}{\foreach \c in {0,1,2,3}{\fill[#2!\t] ({\c*2.6mm},{-\r*2.4mm}) circle (0.85mm);}}
\end{scope}}
\resizebox{\textwidth}{!}{\begin{tikzpicture}
\node[card, draw=SurveyEdge!75, fill=SurveyFill!60, minimum width=34mm, minimum height=45mm] (c1) at (0mm,0mm) {};
\node[card, draw=ModelEdge!75,  fill=ModelFill!60,  minimum width=52mm, minimum height=45mm] (c2) at (48mm,0mm) {};
\node[card, draw=MetricEdge!80, fill=MetricFill!65, minimum width=54mm, minimum height=45mm, line width=1.3pt] (c3) at (114mm,0mm) {};
\node[ctitle, text=SurveyEdge] at ([yshift=-2mm]c1.north) {Reference population};
\node[ctitle, text=ModelEdge]  at ([yshift=-2mm]c2.north) {LLMs under evaluation};
\node[ctitle, text=MetricEdge] at ([yshift=-2mm]c3.north) {Population Fidelity};
\cellgrid{([xshift=-3.9mm,yshift=-10.5mm]c1.north)}{SurveyEdge}
\node[cbody] at ([yshift=-20.5mm]c1.north) {$n$ demographic cells\\from survey respondents\\[1pt] {\color{SurveyEdge}distribution $s_i$ per cell}};
\node[sub, minimum width=23mm, minimum height=13.5mm] (s1) at ([xshift=-12.5mm,yshift=-7.5mm]c2.north) {};
\node[sub, minimum width=23mm, minimum height=13.5mm] (s2) at ([xshift=12.5mm,yshift=-7.5mm]c2.north) {};
\node[sub, minimum width=48mm, minimum height=11.5mm, dashed, draw=CultureEdge!80, fill=CultureFill!70] (s3) at ([yshift=-22mm]c2.north) {};
\chipicon{([yshift=-4.6mm]s1.north)}
\cloudicon{([yshift=-4.5mm]s2.north)}
\node[subtxt] at ([yshift=-9mm]s1.north) {Open-weight};
\node[subtxt] at ([yshift=-9mm]s2.north) {Proprietary};
\slidersicon{([xshift=-20mm,yshift=-5.2mm]s3.north)}
\node[subtxt, text=CultureEdge, anchor=north west, align=left] at ([xshift=-15mm,yshift=-2.2mm]s3.north) {\textbf{Adapted variants}\\e.g.\ culture, distribution or\\subgroup fine-tuning, etc.};
\node[cbody] at ([yshift=-36mm]c2.north) {NTP or FA elicitation\\[1pt] {\color{ModelEdge}distribution $m_i$ per cell}};
\node[chip, draw=MetricEdge!70, fill=white, text=MetricEdge] (k1) at ([xshift=-17mm,yshift=-12mm]c3.north) {Accuracy};
\node[chip, draw=MetricEdge!70, fill=white, text=MetricEdge] (k2) at ([yshift=-12mm]c3.north) {Adaptability};
\node[chip, draw=MetricEdge!70, fill=white, text=MetricEdge] (k3) at ([xshift=17mm,yshift=-12mm]c3.north) {Structure};
\node[chip, draw=MetricEdge, fill=MetricEdge, text=white, inner xsep=5pt] (pfs) at ([yshift=-21.5mm]c3.north) {PFS};
\foreach \k/\a in {k1/north west, k2/north, k3/north east}{\draw[-{Stealth[length=1.6mm,width=1.8mm]}, line width=0.7pt, draw=MetricEdge!70] (\k.south) -- (pfs.\a);}
\node[font=\scriptsize\sffamily, text=MetricEdge, anchor=west] at ([xshift=1.5mm]pfs.east) {geometric mean};
\node[chip, dashed, draw=MetricEdge!70, fill=white, text=MetricEdge] (ctr) at ([xshift=-13mm,yshift=-30mm]c3.north) {Center alignment};
\node[font=\scriptsize\sffamily, text=black!65, anchor=west] at ([xshift=1.5mm]ctr.east) {reported separately};
\node[font=\scriptsize\sffamily, text=black!55, anchor=south] at ([yshift=1.5mm]c3.south) {recomputed for any demographic group};
\draw[flow] (c1.east |- c2.center) -- node[flab, above=1pt]{persona\\prompts} (c2.west |- c2.center);
\draw[flow, draw=ModelEdge!70] (c2.east |- c2.center) -- node[flab, below=1.2pt, text=ModelEdge]{$\{m_i\}$} (c3.west |- c2.center);
\histo{(103.9mm,-20mm)}{ModelEdge}{{1.2mm,3mm,2.2mm,0.8mm}}
\draw[flow, draw=SurveyEdge!70, rounded corners=3pt] (c1.north) -- ++(0,4.5mm) -| (c3.north);
\histo{(64mm,1.2mm)}{SurveyEdge}{{2.8mm,1.6mm,3mm,0.9mm}}
\node[flab, text=SurveyEdge, anchor=west] at (71mm,3mm) {$\{s_i\}$};
\node[badge, fill=SurveyEdge] at (c1.north west) {1};
\node[badge, fill=ModelEdge]  at (c2.north west) {2};
\node[badge, fill=MetricEdge] at (c3.north west) {3};
\end{tikzpicture}}\vspace{-0.5cm}
\caption{
The Population Fidelity framework compares survey and model distributions
cell by cell ($S_{\mathrm{acc}}$), in spread ($S_{\mathrm{adapt}}$), and in
arrangement ($S_{\mathrm{struct}}$); center alignment
($S_{\mathrm{center}}$) is reported separately. Section~\ref{sec:population-fidelity} defines each. A reference population provides $n$
demographic cells with survey distributions $s_i$. Model responses to the
corresponding demographic profiles produce distributions $m_i$. Accuracy,
adaptability, and structure compare the two sets of distributions and combine
into the PFS by geometric mean; center alignment is reported separately.
The metrics can also be recomputed within demographic subsets.
}
\label{fig:framework}
\end{figure}

Figure~\ref{fig:framework} summarizes the Population Fidelity framework.
A reference population is partitioned into demographic cells, with each
cell represented by an empirical response distribution $s_i$ for a given
survey question. The same demographic profiles are then used to elicit
responses from the model, producing corresponding distributions $m_i$.

Population Fidelity compares the survey and model distributions along three
dimensions. Accuracy measures correspondence within individual cells,
adaptability compares the amount of variation across cells, and structure
measures whether the relative differences among cells are preserved. Their
geometric mean defines the Population Fidelity Score (PFS). Center alignment compares the equal-cell survey and model centers and is
reported separately. All quantities can also be recomputed within demographic
subsets.

\subsection{Metric Implementation and Validation}
\label{app:metrics}

This appendix gives the implementation behind every number in Section~\ref{sec:results}: distances, cell retention, degenerate cases, cross-question aggregation, and the validation checks every run passes.

\paragraph{Distances.}
Equation~\ref{eq:nemd} is implemented as the city-block norm of the
difference between the cumulative survey and model distributions, divided
by $K-1$. Pairwise distances are stored as condensed vectors using the same
cell ordering for survey and model responses. Although nEMD is normalized
to $[0,1]$, differences in response scales and item meaning make averages
across questions difficult to interpret. We therefore compute and interpret
all distance-based quantities separately by question.

\paragraph{Cell-level model distributions.}
Multiple elicited profiles can map to the same demographic cell. Under
NTP, let $\mathcal{R}_i$ denote the profiles assigned to cell $i$ and
$q_{ir}$ the response distribution elicited for profile $r\in\mathcal{R}_i$, with $n_{ir}$ the number of WVS respondents in cell $i$ with that profile. Following \citet{boelaert2025machine}, we average over respondents
\[
m_i=\sum_{r\in\mathcal{R}_i} w_{ir}q_{ir},
\qquad 
w_{ir}=\frac{n_{ir}}{\sum_{r'\in\mathcal{R}_i}n_{ir'}}.
\]
Under NTP, $q_{ir}$ is the renormalized valid-token probability vector;
under FA, one answer is sampled per respondent and $m_i$ is the empirical distribution of valid sampled responses
assigned to cell $i$.

\paragraph{Retention and evaluation support.}
The survey contains 687 demographic cells for happiness, religious
attendance, and social trust, and 639 for political ideology. A cell is
scored only when its survey-response proportions are complete. Under FA,
the cell must also contain at least 20 valid model responses. Across runs,
the retained cell count ranges from 606 to 687.

Population-wide analyses use all retained cells. Target-country analyses
use German cells for German fine-tuning and Mexican cells for
Mexican fine-tuning. The complete German subset contains 144 cells
and 10{,}296 cell pairs, while the complete Mexican subset contains 131
cells and 8{,}515 pairs. The demographic-subgroup analysis recomputes all metrics within levels of
country, survey wave, sex, age, education, employment, and marital status.
Across all 164 combinations, this yields 4{,}920 subgroup-level scores; the
main subgroup analysis excludes the two Mixtral reference series, leaving
4{,}440 scores across 148 combinations.

For each paired fine-tuning comparison, the fine-tuned and as-released
conditions are recomputed on their common set of retained cells. This
prevents differences in response coverage from being interpreted as changes
in population fidelity. The common set is the intersection of the cells
retained in both conditions and is used for both population-level and
subgroup-level comparisons. Across the primary paired comparisons, these intersections contain between
617 and 687 cells. Repeated FA runs reuse the same model weights, prompts, decoding configuration, and evaluation procedure, differing only in the seed stream (Section~\ref{sec:experimental-design}). Retained cell sets are highly stable across repeated runs; Appendix~\ref{app:replicates} reports the corresponding retention and run-to-run variability analyses.

Under NTP, prompts assigning zero probability mass to all valid answer
tokens are treated as failed records rather than renormalized. No such
failure affects the retained cells in the reported analyses. For conditions
evaluated under both NTP and FA, the two elicitation modes use the same
retained cell set.

\paragraph{Undefined and degenerate cases.}
Adaptability requires at least two cells and structure at least three.
Below these thresholds, the corresponding quantities remain undefined.
If the model's pairwise-distance vector has zero variance, $\rho$ is
undefined; the model is classified as flat and assigned a structure
score of zero. No population-wide scored run--mode combination meets this condition; two country-level subgroup estimates do and are assigned $S_{\mathrm{struct}}=0$ under this rule
(Table~\ref{tab:families-app}).
The nearest case is Qwen3-VL-2B-Thinking under German fine-tuning on
social trust with FA elicitation. Its median pairwise distance is zero,
so $A=0$, but its distance vector has sufficient variation for $\rho$
to remain defined. The PFS equals zero when any component is zero and
remains undefined when a required component is undefined.

\paragraph{Cross-question aggregation.}
The cross-question PFS is the geometric mean of the four per-question
PFS values. Component summaries are arithmetic means of their
per-question values, and the limiting component is identified from
these averaged scores. The cross-question PFS is therefore not
recomputed from the averaged components. Because a zero PFS on any
question produces a zero cross-question geometric mean, per-question
values remain the primary results.

\paragraph{Implementation checks.}
Before producing tables or figures, the implementation verifies that:
\begin{enumerate}
    \item every recomputed cell-level nEMD matches the corresponding
    run-level distance table row by row;
    \item pooled quantities match independently calculated population
    summaries;
    \item $D_{\mathrm{WVS}}$ is identical across model conditions
    evaluated on the same question, elicitation mode, and retained
    cell set;
    \item all transformed component and PFS values lie within $[0,1]$;
    \item group-level cell counts do not exceed those of their pooled
    parent populations; and
    \item the survey and model pairwise-distance vectors contain the
    same cells in the same order.
\end{enumerate}
All 164 run--mode combinations pass these checks.

Center alignment in the Population Fidelity Framework gives equal
weight to retained cells. For comparability with the original study,
the reproduction output also preserves its respondent-weighted center.
Both quantities are retained and labeled according to their weighting
scheme.

\subsection{Stylized Example Calculations}
\label{app:toy-example}

Figure~\ref{fig:pfs-toy} uses three subpopulations, A--C. Their survey
distributions are normal with means $4$, $5$, and $6$ and standard deviation
$1.3$, on a 0--10 attitude scale with 101 response options
($0, 0.1, \ldots, 10$). The pooled survey distribution $\bar{s}$ is their
average.

M1 assigns $\bar{s}$ to every group. M2 re-sorts the same respondents into
three narrower groups, each still one third of respondents, with means
$3.34$, $5.00$, and $6.66$ and standard deviations $0.81$, $0.44$, and
$0.81$. M3 exchanges the survey distributions of B and C. Because each model
only reassigns the pooled respondents across groups, each averages to
$\bar{s}$, so $S_{\mathrm{center}}=1$ for all three models. Values are
rounded to two decimals, derived quantities use unrounded values, and the
exact construction is included in the released code.

The cell-level errors for M1, M2, and M3 are, respectively,
\[
[0.10,0.02,0.10],\qquad
[0.07,0.07,0.07],\qquad
[0.00,0.10,0.10].
\]
Each therefore has mean error $E\approx0.07$ and accuracy
$S_{\mathrm{acc}}\approx0.93$.

Using pair order (AB, AC, BC), the pairwise-distance vectors are
\[
d_s=[0.10,0.20,0.10],\qquad
d_{\mathrm{M1}}=[0,0,0],
\]
\[
d_{\mathrm{M2}}=[0.17,0.33,0.17],\qquad
d_{\mathrm{M3}}=[0.20,0.10,0.10].
\]
Their median pairwise distances are $0.10$ for the survey, $0$ for M1,
$0.17$ for M2, and $0.10$ for M3. The adaptability ratios are therefore
$A=0$, $1.67$, and $1.00$, and the adaptability scores are $0$, $0.60$, and
$1.00$.

M1's pairwise distances are all equal, so $\rho$ is undefined and M1 is
assigned $S_{\mathrm{struct}}=0$ (Appendix~\ref{app:metrics}). Treating
distances that are equal by symmetry as tied, the Spearman correlations
with the survey distance vector are $1.00$ for M2 and $-0.50$ for M3. After
applying $S_{\mathrm{struct}}=\max(0,\rho)$, the structure scores are $0$,
$1.00$, and $0$. Finally,
\[
\mathrm{PFS}_{\mathrm{M1}}=0,\qquad
\mathrm{PFS}_{\mathrm{M2}}\approx0.82,\qquad
\mathrm{PFS}_{\mathrm{M3}}=0.
\]
Thus, the three models reproduce the pooled survey distribution exactly and
have nearly identical accuracy, but differ in the amount and structure of
their between-group variation. M1 corresponds to the survey-center baseline
(Appendix~\ref{app:survey-center-baseline}).

\section{Experimental Configuration and Prompts}\label{app:experimental-details}
\label{app:capacity}

\subsection{Models and Fine-Tuning Configuration}

\paragraph{Model inventory.}

Table~\ref{tab:model-inventory} summarizes the six open-weight models used
in the main evaluation. Each is evaluated as released and after German and
Mexican fine-tuning. Within each model, the as-released and
fine-tuned conditions use the same tokenizer, numerical configuration, and
inference settings. Exact repository revisions are recorded in the released
run manifests.

Mixtral-8x7B is retained separately as a reproduction reference. The
archived condition uses generations released with the original Machine Bias
replication package, while the fresh condition regenerates responses from
Mixtral-8x7B-v0.1 using the original prompts and elicitation procedures.
All local open-weight training and inference experiments were run on a
single NVIDIA GeForce RTX 5090 (32~GB). The backend used for the fresh
Mixtral reproduction is described in Appendix~\ref{app:reproduction}.

\begin{table}[htb]
\centering
\tiny
\begin{adjustbox}{max width=\textwidth}
\begin{tabular}{llllr}
\toprule
LLM  & Precision & Quantization & Adapted modules & Transformers (inference) \\
\midrule
Gemma-4-31B-it~\citep{team2026gemma} & bf16 & NF4 & 230 & 5.8.1 \\
Gemma-4-E4B-it~\citep{team2026gemma} & bf16 & -- & 132 & 5.8.1 \\
Qwen3-VL-8B-Thinking~\citep{bai2025qwen3} & bf16 & -- & 144 & 5.8.1 \\
Qwen3-VL-2B-Thinking~\citep{bai2025qwen3} & bf16 & -- & 112 & 5.8.1 \\
Llama-3.2-3B~\citep{grattafiori2024llama} & bf16 & -- & 112 & 5.8.1 \\
Muse-Glimmer-30B~\citep{meta2026museglimmer} & bf16 & NF4 & 208 & 5.15.0 \\
\bottomrule
\end{tabular}
\end{adjustbox}
\caption{
Open-weight model inventory.
NF4 is 4-bit NormalFloat quantization implemented with
\texttt{bitsandbytes} 0.49.2. Adapted-modules column reports the number
of modules receiving LoRA updates. Every model runs in bfloat16 compute with
\texttt{PEFT} 0.19.1 and PyTorch 2.12.0 under Python 3.11.
}
\label{tab:model-inventory}
\end{table}

\paragraph{Fine-tuning configuration.}
For each target culture, the training data combine the WVS seed questions
selected by CultureLLM with semantically augmented paraphrases. Each record
is paired with a culture-specific target answer derived from aggregate survey
responses. The German condition uses CultureLLM's German partition. For the
Mexican condition, we reconstruct the records using only Mexican
respondents rather than CultureLLM's original Spanish partition, which pools
respondents from Mexico and Argentina.

Fine-tuning uses LoRA with rank $r=8$, scaling parameter $\alpha=16$, and
dropout $0.05$. Adapters are applied to the query, key, value, and output
projections of the text-attention blocks
(\texttt{q\_proj}, \texttt{k\_proj}, \texttt{v\_proj}, and
\texttt{o\_proj}); vision and audio components are excluded. QLoRA with NF4
quantization is used for Gemma-4-31B-it and Muse-Glimmer-30B. Within each
model family, the German and Mexican variants use the same training
configuration. None of the four evaluation questions appears in the
fine-tuning data.

Training uses supervised fine-tuning on chat-formatted records with AdamW,
no weight decay, a cosine learning-rate schedule, an effective batch size of
16, gradient clipping at 0.3, a maximum sequence length of 512 tokens, and
seed 42. Warmup lasts 50 optimizer steps, or 100 steps for the QLoRA models.
Training is allowed to continue for at most 250 epochs and is stopped early
based on validation loss after a minimum of 20 epochs, with patience of 15
evaluations. Validation is performed every 100 optimizer steps, and the
checkpoint with the lowest validation loss is retained. Table~\ref{tab:finetuning} reports the learning rate, batch size, run length and held-out accuracy of each model.

\begin{table}[htb]
\centering
\tiny
\begin{adjustbox}{max width=\textwidth}
\begin{tabular}{llllrrrr}
\toprule
 & & & & \multicolumn{2}{c}{German} & \multicolumn{2}{c}{Mexican} \\
\cmidrule(lr){5-6} \cmidrule(lr){7-8}
LLM & Adapter & LR & Batch & Epochs (steps) & Val. acc. &
Epochs (steps) & Val. acc. \\
\midrule
Gemma-4-31B-it & QLoRA, NF4 & $10^{-4}$ & $1 \times 16$ & 33 (3{,}900) & 0.97 & 45 (2{,}700) & 0.88 \\
Gemma-4-E4B-it & LoRA, bf16 & $2 \times 10^{-4}$ & $2 \times 8$ & 37 (4{,}400) & 0.97 & 45 (2{,}700) & 0.87 \\
Qwen3-VL-8B-Thinking & LoRA, bf16 & $2 \times 10^{-4}$ & $2 \times 8$ & 58 (6{,}900) & 0.97 & 45 (2{,}700) & 0.87 \\
Qwen3-VL-2B-Thinking & LoRA, bf16 & $2 \times 10^{-4}$ & $4 \times 4$ & 58 (6{,}900) & 0.97 & 45 (2{,}700) & 0.86 \\
Llama-3.2-3B & LoRA, bf16 & $2 \times 10^{-4}$ & $4 \times 4$ & 80 (9{,}500) & 0.88 & 45 (2{,}700) & 0.77 \\
Muse-Glimmer-30B & QLoRA, NF4 & $10^{-4}$ & $1 \times 16$ & 43 (5{,}100) & 0.98 & 45 (2{,}700) & 0.89 \\
\bottomrule
\end{tabular}
\end{adjustbox}
\caption{
Fine-tuning configuration and run length by model and target culture.
Batch reports the per-device batch size $\times$ gradient-accumulation
steps, yielding an effective batch size of 16. Parentheses report optimizer
steps at termination. Val.\ acc.\ is the held-out token accuracy of the
retained checkpoint.
}
\label{tab:finetuning}
\end{table}

The German dataset contains 2{,}100 question--answer pairs, split into
1{,}890 training and 210 validation records. The Mexican dataset is
constructed from the same 50 WVS seed questions and 1{,}000 augmented
paraphrases, using aggregate responses from Mexican respondents only. It
contains 1{,}050 pairs, split into 945 training and 105 validation records.

All fine-tuning records are in English: each contains a system prompt naming the target country, a WVS-derived question with numbered response options, and the corresponding target answer. The experiment therefore evaluates cultural adaptation within a common language rather than cross-lingual transfer. Training was performed on the hardware described in
Appendix~\ref{app:reproduction}.

\subsection{Response Classification, Capacity, and Coverage}

\paragraph{Response classification.}
Each FA generation is classified as \emph{valid} if it maps
unambiguously to one offered response option, \emph{invalid} if it does
not satisfy the required format, and \emph{failed} if no generation is
completed. Invalid and failed outputs are retained for audit. For NTP,
valid-token probability mass is measured before renormalization. A
response assigning no probability mass to any valid token is recorded
as degenerate rather than silently renormalized. Coverage records and
the original generated outputs are included in the repository.

\paragraph{Capacity and response coverage.}
Each run receives a 32-prompt preflight probe. Under NTP, a run is
flagged as non-informative if the mean probability mass assigned to valid
answer tokens is below 0.10. No NTP run in the reported evaluation falls
below this threshold.

Under FA, we do not apply a separate run-level threshold. Distances are
computed only for cells containing at least 20 valid responses, and dropped
cells are recorded for each run. Across the evaluated FA conditions, the
lowest overall valid-answer rate is 0.926.

\paragraph{Fine-tuning health.}
Before elicitation, every adapter used in the reported experiments is
checked using the mean norm of the weight update applied to the
attention projections and held-out token accuracy against a 50\%
usability threshold. All evaluated adapters pass this check, with
update norms between 0.2 and 5.4 and held-out accuracy between 77\%
and 98\%. No reported Population Fidelity result therefore comes from
a checkpoint classified as a training failure.

\subsection{API-Served Model}

\paragraph{API-served model.}
GPT-5.6 Terra is evaluated through a commercial chat endpoint using the
same survey prompts as the open-weight models. Requests use the model
identifier \texttt{gpt-5.6-terra} and were issued in August 2026. Reasoning
effort is fixed at \emph{medium}, FA responses are sampled at temperature
0.7, and the maximum output length is 300 tokens. A seed of 20240110 is
passed with each request as a best-effort parameter.

At nonzero reasoning effort, the endpoint does not provide usable
next-token probabilities for our NTP procedure. In a single-token probe
requesting the top 20 log probabilities, none of the valid survey-answer
tokens appeared among the returned candidates. Terra is therefore evaluated
under FA only.

Each request records the prompt, requested sampling parameters, parameters
applied by the endpoint, and returned output. Prompts are submitted as a
single user message rather than as raw completions.

\paragraph{Computational scale.}
The experiment requires approximately $7.9$ million model queries,
excluding the archived Mixtral generations. Each open-weight NTP run
evaluates 13{,}904 unique demographic profiles, while each FA run
evaluates 26{,}981 respondent-level prompts.

\subsection{Questions and Prompt Construction}
\label{app:prompts}

Every response analysed in this paper comes from the four survey items and the prompt template below, used as described in Sections~\ref{sec:machine-bias} and~\ref{sec:evaluation-design}.

Prompts for the four items in Table~\ref{tab:topics} follow the
interview-style template used by \citet{boelaert2025machine}.
As illustrated in Figure~\ref{fig:prompts}, the interviewer asks a
sequence of sociodemographic questions and the interviewee supplies
the values defining the relevant cell. The model then completes the
answer to the final attitudinal question. Categorical demographic
questions use lettered options and expect a single capital letter.
Under NTP elicitation, the probability mass assigned to the valid
first-token options is extracted and renormalized.

\begin{table}[thb]
\centering
\tiny
\begin{tabular}{llr}
\toprule
Variable & Topic & Options \\
\midrule
\texttt{d\_happy} & Happiness & 4 \\
\texttt{d\_polpos} & Political ideology & 10 \\
\texttt{d\_religiousp} & Religious attendance & 7 \\
\texttt{d\_trust} & Social trust & 2 \\
\bottomrule
\end{tabular}
\caption{Attitudinal items used in the evaluation.}
\label{tab:topics}
\end{table}

\begin{figure*}[htb]
    \centering
    \input{architecture/prompt-inteview}
    \caption{Interview-style prompt template. Placeholders are replaced
    with the demographic values defining each cell; survey year and age
    are inserted as integers, while categorical attributes use the
    corresponding lettered options. Demographic items absent from a
    merged profile are omitted. The final question is replaced by one
    of the four attitudinal items in Table~\ref{tab:topics}, and the
    prompt ends at \texttt{Answer:} for the model to complete.}
    \label{fig:prompts}
\end{figure*}

Political ideology follows the original study's mode-specific
encoding. NTP uses a 0--9 scale so that each option corresponds to a
single digit token, whereas FA uses the survey's 1--10 scale. The two
encodings are aligned by ordinal position before the response
distributions are compared. The NTP prompt for political ideology
includes a trailing space after \texttt{Answer:} to preserve the
original tokenization. Political ideology is unavailable for 48 cells
from a survey wave in which the item was not administered; these cells
are treated as having undefined targets and excluded from analyses of
that question. Unit tests verify the prompt bytes for all four
questions and both elicitation modes against the original
implementation.

\section{Supplementary Results}
\label{app:supplementary}

\subsection{Survey-Center Baseline for Accuracy}
\label{app:survey-center-baseline}

Accuracy is bounded, and on the nEMD scale a prediction that ignores
demographic information can already attain a high score. We therefore
compare model accuracy with such a baseline. For the retained cells $I$ of
a run, we take the equal-cell survey center $\bar{s}_I$ defined in
Section~\ref{sec:pfs} and predict it for every cell,
$m_i^{(0)}=\bar{s}_I$ for all $i\in I$. Its accuracy is
\[
S_{\mathrm{acc}}^{(0)}
=
1-\frac{1}{|I|}\sum_{i\in I}
\operatorname{nEMD}\left(s_i,\bar{s}_I\right),
\]
which is the accuracy attained without using demographic information. We
also express model performance relative to this baseline using the
skill-score form
\[
S_{\mathrm{acc}}^{\mathrm{above\,baseline}}
=
\frac{S_{\mathrm{acc}}-S_{\mathrm{acc}}^{(0)}}
{1-S_{\mathrm{acc}}^{(0)}}.
\]
This score is zero at the survey-center baseline, one for perfect
cell-level prediction, positive when a model exceeds the baseline, and
negative when it performs worse.

We score the baseline using the same implementation as the model
conditions, so its other components follow directly from the definitions
in Section~\ref{sec:population-fidelity}. Its center alignment is one by
construction. Because all predicted cells are identical, all pairwise
model distances are zero, giving $A=0$ and
$S_{\mathrm{adapt}}=0$; the flatness rule of
Appendix~\ref{app:metrics} sets $S_{\mathrm{struct}}=0$, and therefore
$\mathrm{PFS}=0$. Because $\bar{s}_I$ includes each cell's own survey
distribution, we also compute a leave-one-out version in which cell $i$
is excluded from the center used to predict it. This changes
$S_{\mathrm{acc}}^{(0)}$ by less than $0.0003$ at the population level
and by at most $0.005$ within demographic levels, indicating that the
baseline's performance is not driven by including each cell in its own
reference center.

Table~\ref{tab:null-baseline} compares the baseline with the models, with
each run evaluated on its own retained cells. The survey-center baseline
attains $S_{\mathrm{acc}}^{(0)}$ between $0.848$ for religious attendance
and $0.915$ for political ideology and exceeds the median model accuracy
for every question and elicitation mode. Only 7 of the 164 combinations
exceed their corresponding baseline, by at most $0.036$ in accuracy:
on happiness, both Mixtral runs under NTP, the regenerated Mixtral run
under FA, and the as-released and Mexican Muse-Glimmer-30B
conditions under NTP; on religious attendance, the Mexican
Muse-Glimmer-30B condition under NTP; and on social trust, as-released
Qwen3-VL-8B under NTP. The median above-baseline score is $-0.56$, and
52 combinations fall below $-1$, meaning that their mean cell-level
error is more than twice that of the corresponding survey-center
baseline.

\begin{table}[htb]
\centering
\tiny
\setlength{\tabcolsep}{4pt}
\begin{adjustbox}{max width=\textwidth}
\begin{tabular}{llrrccr}
\toprule
Question & Mode & Cells & $S_{\mathrm{acc}}^{(0)}$ & Model $S_{\mathrm{acc}}$ &
$S_{\mathrm{acc}}^{\mathrm{above\,baseline}}$ & Above baseline \\
\midrule
Happiness & FA & 678--687 & 0.901--0.902 & 0.860 [0.749, 0.906] & $-0.41$ [$-1.53$, 0.05] & 1 / 21 \\
 & NTP & 678--687 & 0.901--0.902 & 0.871 [0.769, 0.911] & $-0.30$ [$-1.33$, 0.11] & 4 / 20 \\
Political ideology & FA & 606--639 & 0.913--0.915 & 0.875 [0.817, 0.905] & $-0.45$ [$-1.11$, $-0.09$] & 0 / 21 \\
 & NTP & 626--639 & 0.913--0.914 & 0.892 [0.848, 0.910] & $-0.25$ [$-0.77$, $-0.03$] & 0 / 20 \\
Religious attendance & FA & 677--687 & 0.848 & 0.707 [0.441, 0.833] & $-0.92$ [$-2.67$, $-0.10$] & 0 / 21 \\
 & NTP & 677--687 & 0.848 & 0.754 [0.480, 0.884] & $-0.61$ [$-2.41$, 0.24] & 1 / 20 \\
Social trust & FA & 630--687 & 0.867--0.868 & 0.683 [0.350, 0.855] & $-1.39$ [$-3.90$, $-0.09$] & 0 / 21 \\
 & NTP & 630--687 & 0.867--0.868 & 0.703 [0.359, 0.883] & $-1.24$ [$-3.84$, 0.11] & 1 / 20 \\
\midrule
All questions & both & 606--687 & 0.848--0.915 & 0.832 [0.350, 0.911] & $-0.56$ [$-3.90$, 0.24] & 7 / 164 \\
\midrule
Demographic levels & both & 24--377 & 0.817--0.943 & 0.824 [0.185, 0.952] & $-0.72$ [$-11.58$, 0.40] & 157 / 4{,}440 \\
Target countries & both & 130--144 & 0.875--0.943 & 0.816 [0.384, 0.942] & $-1.24$ [$-4.89$, 0.01] & 1 / 96 \\
\bottomrule
\end{tabular}
\end{adjustbox}
\caption{
Model accuracy relative to the survey-center baseline.
$S_{\mathrm{acc}}^{(0)}$ is obtained by predicting the equal-cell survey
center for every retained cell, so its range reflects differences in
retained cells across runs. Model $S_{\mathrm{acc}}$ and the above-baseline
score report the median across combinations, with the minimum and maximum
in brackets; the final column counts combinations whose accuracy exceeds
their corresponding baseline. For demographic-level results, the baseline
is recomputed within each level. Target-country results compare German
fine-tuned conditions within Germany and Mexican fine-tuned
conditions within Mexico. The baseline has $\mathrm{PFS}=0$ in every case.
}
\label{tab:null-baseline}
\end{table}

The same pattern holds within demographic levels, where each level is
compared with the center of its own retained cells. Across the 4{,}440
subgroup-level scores of Section~\ref{sec:results}, models exceed their
level-specific survey-center baseline in 157 cases. Within the target
countries, German fine-tuned conditions never exceed the German
survey-center baseline (0 of 48 comparisons), while Mexican
fine-tuned conditions exceed the Mexican baseline once (1 of 48).

This baseline clarifies why high accuracy does not imply population
fidelity. Although models exceed the baseline's structure score of zero
in 134 of the 164 combinations, they exceed its accuracy in only seven.
Much of their absolute accuracy therefore reflects proximity to the survey center, consistent with the machine-bias pattern of
\citet{boelaert2025machine}. The baseline also illustrates the behavior that PFS is designed to reject: a prediction that outperforms nearly all evaluated combinations on accuracy and matches the survey center exactly receives $\mathrm{PFS}=0$ because it contains no variation between groups.

We report the above-baseline score as a diagnostic and do not include it in
PFS. Within each question, the baseline varies by at most $0.0015$ across
runs, so the above-baseline score produces almost the same ordering as
$S_{\mathrm{acc}}$, with Spearman correlations of at least $0.998$.
It therefore changes the reference point while barely affecting model
rankings. Unlike the PFS components, which lie in $[0,1]$, the above-baseline
score can take large negative values when model error substantially
exceeds baseline error; in our population-level results it reaches
$-3.90$.

\subsection{Model- and Question-Specific Results}
\label{app:model-results}

Under FA elicitation, fine-tuned conditions attain the highest PFS on three of the four questions, all with Muse-Glimmer-30B: its German condition leads on political ideology and social trust, while its Mexican condition ties GPT-5.6 Terra on religious attendance after rounding. Happiness is led by as-released
Qwen3-VL-2B (Table~\ref{tab:main-results}).

\begin{table}[htb]
\centering
\tiny
\setlength{\tabcolsep}{4.2pt}
\begin{adjustbox}{max width=\textwidth}
\begin{tabular}{llcccc}
\toprule
LLM & Condition & Happiness & Politics & Religious & Trust \\
\midrule
Mixtral (archived) & as released & 0.303 & 0.335 & 0.142 & 0.182 \\
Mixtral (fresh) & as released & 0.340 & 0.360 & 0.073 & 0.247 \\
GPT-3 (archived) & as released & 0.245 & 0.317 & 0.000 & 0.221 \\
GPT-5.6 Terra & as released & 0.373 & 0.270 & \textbf{0.726} & 0.416 \\
\midrule
Gemma-4-31B-it & as released & 0.072 & 0.381 & 0.128 & 0.300 \\
& $\mathcal{F}_{\text{de}}$ & 0.329 & 0.303 & 0.000 & 0.000 \\
& $\mathcal{F}_{\text{mx}}$ & 0.000 & 0.410 & 0.000 & 0.301 \\
Gemma-4-E4B-it & as released & 0.284 & 0.162 & 0.000 & 0.000 \\
& $\mathcal{F}_{\text{de}}$ & 0.409 & 0.253 & 0.119 & 0.000 \\
& $\mathcal{F}_{\text{mx}}$ & 0.248 & 0.315 & 0.117 & 0.000 \\
Qwen3-VL-8B-Thinking & as released & 0.413 & 0.274 & 0.000 & 0.382 \\
& $\mathcal{F}_{\text{de}}$ & 0.420 & 0.227 & 0.000 & 0.117 \\
& $\mathcal{F}_{\text{mx}}$ & 0.294 & 0.277 & 0.000 & 0.178 \\
Qwen3-VL-2B-Thinking & as released & \textbf{0.488} & 0.285 & 0.000 & 0.233 \\
& $\mathcal{F}_{\text{de}}$ & 0.141 & 0.194 & 0.225 & 0.000 \\
& $\mathcal{F}_{\text{mx}}$ & 0.338 & 0.226 & 0.000 & 0.000 \\
Llama-3.2-3B & as released & 0.457 & 0.330 & 0.000 & 0.284 \\
& $\mathcal{F}_{\text{de}}$ & 0.273 & 0.322 & 0.112 & 0.119 \\
& $\mathcal{F}_{\text{mx}}$ & 0.385 & 0.289 & 0.187 & 0.300 \\
Muse-Glimmer-30B & as released & 0.352 & 0.380 & 0.593 & 0.363 \\
& $\mathcal{F}_{\text{de}}$ & 0.356 & \textbf{0.452} & 0.540 & \textbf{0.457} \\
& $\mathcal{F}_{\text{mx}}$ & 0.329 & 0.424 & \textbf{0.726} & 0.277 \\
\bottomrule
\end{tabular}
\end{adjustbox}
\caption{
Population Fidelity Score by question under FA elicitation.
The highest score in each column is shown in bold. GPT-5.6 Terra and
Muse-Glimmer-30B ($\mathcal{F}_{\text{mx}}$) both score 0.726 on religious
attendance after rounding. GPT-3 (archived) denotes the proprietary
FA series from the original Machine Bias study, evaluated with the
same Population Fidelity metrics on its archived outputs.
}
\label{tab:main-results}
\end{table}

Averaged across questions and elicitation modes, only Gemma-4-E4B and
Muse-Glimmer-30B improve under both cultural conditions. Gemma-4-E4B has
mean $\Delta\mathrm{PFS}$ of $+0.063$ under $\mathcal{F}_{\text{de}}$ and
$+0.050$ under $\mathcal{F}_{\text{mx}}$, while Muse-Glimmer-30B improves
by $+0.046$ and $+0.025$, respectively. Qwen3-VL-2B declines by $-0.114$
and $-0.109$ and improves in none of its eight
$\mathcal{F}_{\text{mx}}$ comparisons. Fine-tuning effects also vary
substantially by question: $\mathcal{F}_{\text{de}}$ improves eight of 12
comparisons for happiness but three for social trust, while
$\mathcal{F}_{\text{mx}}$ improves two for happiness and seven for political
ideology. No model improves on all four questions under both modes for either
target.

Those PFS changes hide components that move in different directions. Mean changes over the 48 comparisons, on all demographic cells and for $\mathcal{F}_{\text{de}}$ then $\mathcal{F}_{\text{mx}}$, are $-0.077$ and $+0.014$ for $S_{\mathrm{acc}}$, $-0.103$ and $-0.064$ for $S_{\mathrm{adapt}}$, and $+0.008$ and $+0.004$ for $S_{\mathrm{struct}}$; Section~\ref{sec:finetuning-results} gives them within the target countries.

\paragraph{Comparison with archived proprietary-model outputs.}
The archived outputs from the original Machine Bias study provide a descriptive reference for comparing GPT-5.6 Terra with an earlier proprietary model under FA elicitation. \citet{boelaert2025machine} use \texttt{davinci-002}, referred to as GPT-3 in the original study, for their FA robustness analysis. We evaluate those archived outputs with the same Population Fidelity metrics used for Terra. This comparison is observational rather than controlled: the models differ in model vintage, post-training, interface, and generation procedure, so observed
differences cannot be attributed to any single factor; the corresponding
scores appear in Table~\ref{tab:main-results}.

Terra attains a higher PFS on three of the four questions: happiness
($0.373$ vs.\ $0.245$), religious attendance ($0.726$ vs.\ $0.000$),
and social trust ($0.416$ vs.\ $0.221$). Its PFS is lower for political
ideology ($0.270$ vs.\ $0.317$). The largest difference is in structure,
where Terra scores higher on three questions and is nearly unchanged on
political ideology. 

\begin{figure}[!htb]
\centering
\includegraphics[width=0.95\columnwidth]{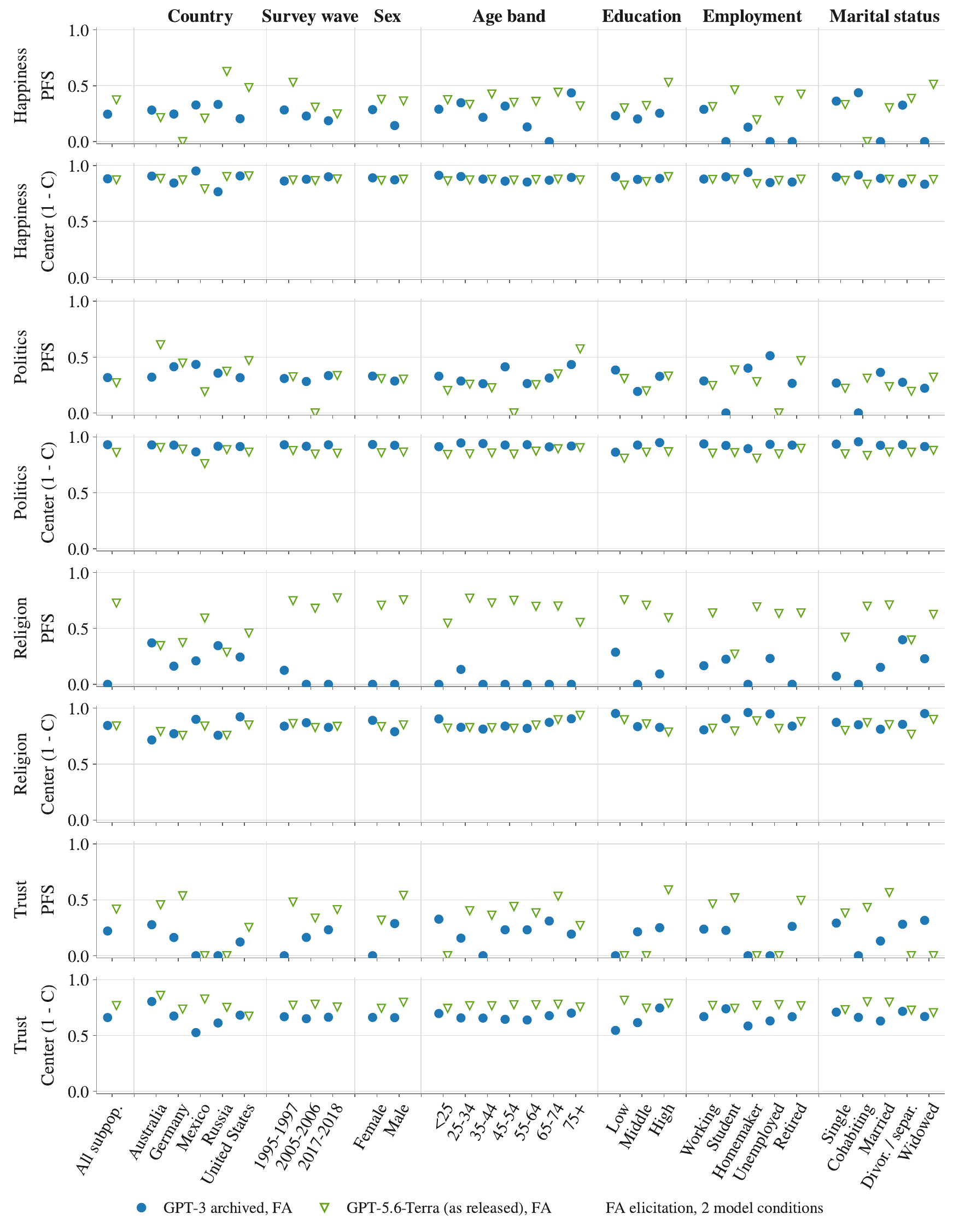}
\caption{
PFS and center alignment by demographic level for the two proprietary
models under FA elicitation, shown separately by question. Within each
question, the upper row shows PFS and the lower row center alignment;
demographic columns follow Figure~\ref{fig:subgroups-main}. Blue circles
denote archived GPT-3 outputs from the original study, and green triangles
denote GPT-5.6 Terra as released.
}
\label{fig:proprietary}
\end{figure}

Figure~\ref{fig:proprietary} shows the same comparison within demographic
levels. Across 120 level-specific comparisons, Terra has higher structure
in 109 and higher PFS in 86, but higher center alignment in only 54 and
higher adaptability in 25. Thus, in this descriptive comparison, Terra
more often recovers the observed organization of differences among groups,
without consistently improving either the amount of between-group variation
or agreement with the survey center.

\subsection{Accuracy and Adaptability by Question}
\label{app:adaptability-results}

Compression dominates, with 43 of the 52 as-released run--mode combinations at $A<1$ and only 9 at $A>1$ (Table~\ref{tab:adaptability-counts}). All 26 happiness and political ideology combinations compress; amplification occurs only for religious attendance and social trust. Only one of the nine amplifying combinations reaches $\rho>0.1$,
illustrating why adaptability and structure must be examined separately.

\begin{table}[htb]
\centering
\tiny
\begin{adjustbox}{max width=\columnwidth}
\begin{tabular}{llrrrl}
\toprule
Question & Mode & $n$ & $A<1$ & $A>1$ & $A$ range \\
\midrule
Happiness & NTP & 6 & 6 & 0 & 0.11--0.72 \\
 & FA & 7 & 7 & 0 & 0.08--0.95 \\
Political ideology & NTP & 6 & 6 & 0 & 0.11--0.77 \\
 & FA & 7 & 7 & 0 & 0.07--0.89 \\
Religious attendance & NTP & 6 & 4 & 2 & 0.29--1.61 \\
 & FA & 7 & 3 & 4 & 0.32--1.68 \\
Social trust & NTP & 6 & 5 & 1 & 0.33--1.56 \\
 & FA & 7 & 5 & 2 & 0.26--1.56 \\
\midrule
All & & 52 & 43 & 9 & 0.07--1.68 \\
\bottomrule
\end{tabular}
\end{adjustbox}
\caption{Compression is universal for happiness and political ideology, while amplification occurs only for religious attendance and social trust, as the $A<1$ and $A>1$ columns show by counting conditions producing less and more between-group variation than the survey. Counts cover
the six open-weight LLMs under both elicitation modes and GPT-5.6 Terra
under FA, excluding the Mixtral reference series.}
\label{tab:adaptability-counts}
\end{table}

\begin{figure*}[!t]
\centering
\includegraphics[width=0.93\textwidth]
{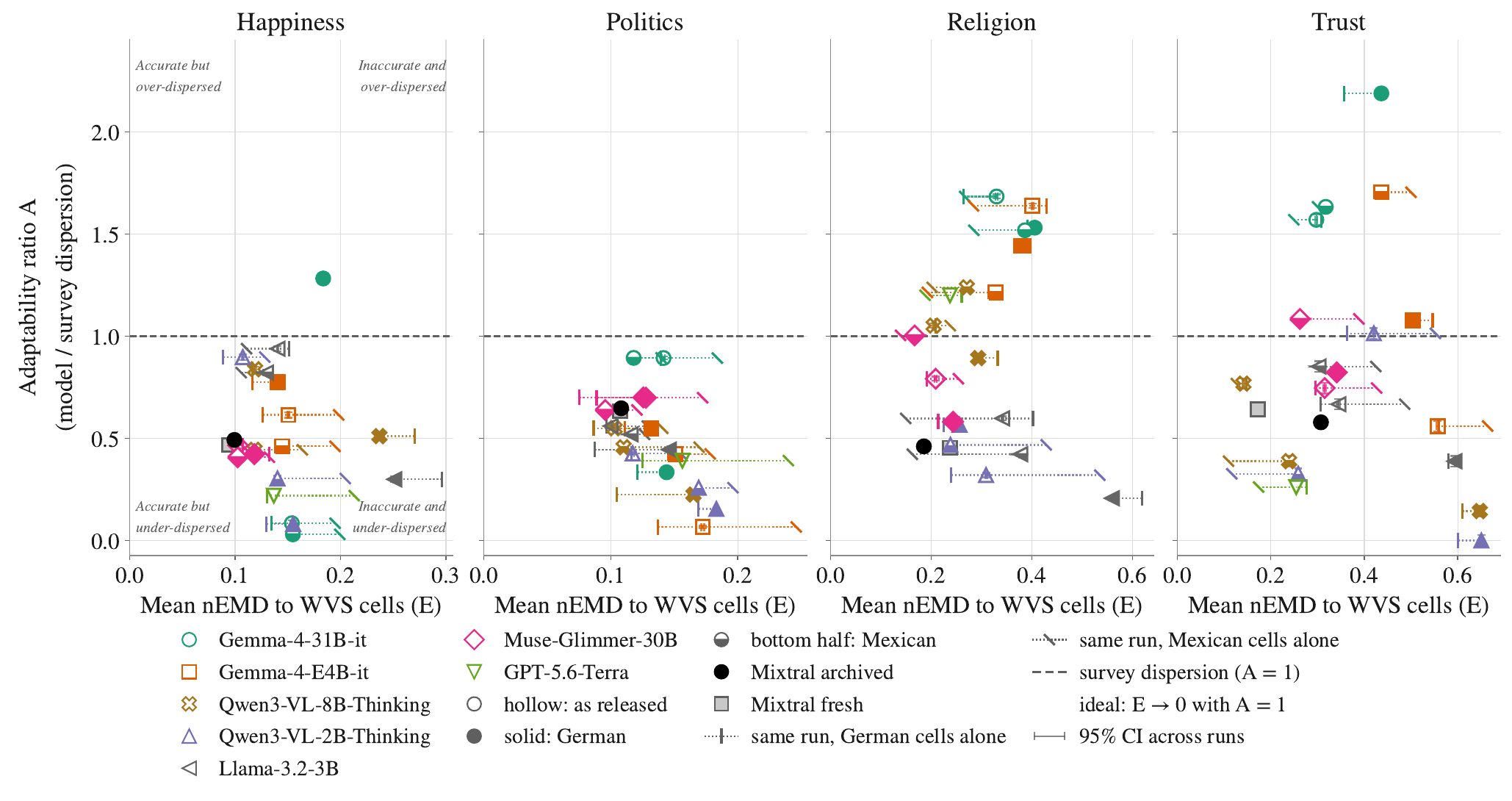}
\caption{Cell-level error and adaptability under FA elicitation. The
horizontal axis reports mean nEMD to the WVS cells, with lower values
indicating greater accuracy. The vertical axis reports the adaptability
ratio $A$. Values below $A=1$ indicate compression of between-group
variation, while values above one indicate amplification. Adaptability
must be interpreted alongside structure because matching the amount of
variation does not establish that it occurs between the corresponding
social groups.}
\label{fig:plane}
\end{figure*}

Figure~\ref{fig:plane} complements these counts by showing how
$S_{\mathrm{acc}}$ relates to $A$ under FA elicitation. Conditions
near $A=1$ reproduce the survey's overall amount of between-group
variation, but this does not establish that the variation is assigned
to the corresponding social groups.

\subsection{Demographic Breakdown}
\label{app:demographic-results}

The demographic-subgroup analysis contains 1{,}440 paired scores for each
cultural target. German fine-tuning improves PFS in 42.5\% of these
comparisons, reduces it in 49.7\%, and leaves 7.8\% unchanged. The
corresponding proportions for Mexican fine-tuning are 40.8\%,
48.5\%, and 10.7\%.

Improvements are not concentrated in the target countries. Within
Germany, German fine-tuning improves 18 of 48 comparisons, with a mean
PFS change of $-0.021$. Across the other countries, it improves 38.0\%
of comparisons, with a mean change of $-0.019$. Within Mexico,
Mexican fine-tuning also improves 18 of 48 comparisons, with a
mean change of $-0.010$. Outside Mexico, it improves 32.3\%, with a mean
change of $-0.035$.

\begin{figure}[t]
\centering
\includegraphics[width=0.7\columnwidth]
{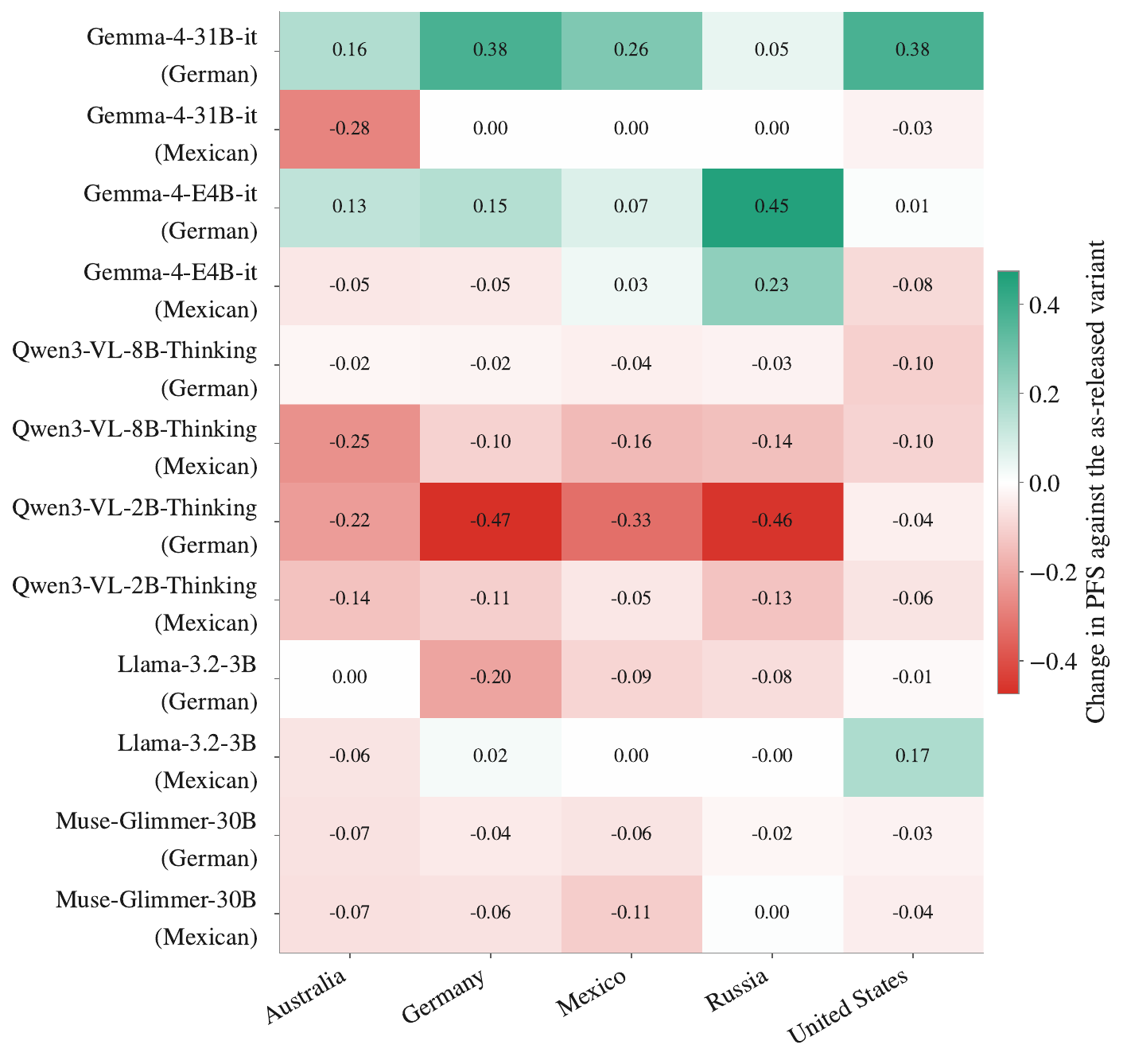}
\caption{Change in PFS by country after cultural fine-tuning for
happiness under FA elicitation. Each row represents a fine-tuned
condition and each column a surveyed country. Green indicates
improvement relative to the corresponding as-released model, and red
indicates decline. For happiness under FA, target-country changes are
not consistently more favorable than changes in the other surveyed
countries.}
\label{fig:country-shift}
\end{figure}

For each condition and demographic family, we define the within-condition spread as the maximum minus minimum PFS across that family's levels and average this quantity across conditions. The mean within-condition spread in PFS is largest across age
($0.287$) and employment ($0.270$), followed by country ($0.254$),
marital status ($0.241$), education ($0.179$), survey wave ($0.107$),
and sex ($0.054$). Although adaptability varies substantially across levels, PFS spread is
more sensitive to structure: holding each condition's structure at its family mean reduces the mean spread to between $0.018$ and $0.080$, whereas holding adaptability at its family mean leaves it between $0.048$ and $0.269$.

Mean PFS generally increases with age, although not monotonically. It
is $0.217$ and $0.207$ in the two youngest groups and $0.310$ and
$0.317$ in the two oldest. Students have the lowest mean PFS among
employment groups ($0.171$), compared with $0.274$ for retired
respondents. These estimates should be interpreted cautiously because
the oldest and student groups contain comparatively few cells and cell
pairs. Table~\ref{tab:families-app} reports the complete results and
sample sizes.

\begin{table}[htb]
\centering
\tiny
\setlength{\tabcolsep}{5pt}
\begin{adjustbox}{max width=\textwidth}
\begin{tabular}{llrrrrrr}
\toprule
Family & Level & PFS & $S_{\mathrm{acc}}$ & $S_{\mathrm{adapt}}$ & $S_{\mathrm{struct}}$ & Cells & Pairs \\
\midrule
Country & United States & 0.290 & 0.783 & 0.508 & 0.090 & 140 & 9{,}674 \\
 & Germany & 0.273 & 0.773 & 0.527 & 0.093 & 144 & 10{,}243 \\
 & Australia & 0.270 & 0.772 & 0.549 & 0.079 & 136 & 9{,}126 \\
 & Mexico & 0.224 & 0.749 & 0.511 & 0.063 & 131 & 8{,}461 \\
 & Russia & 0.219 & 0.755 & 0.488 & 0.060 & 124 & 7{,}817 \\
\midrule
Age & $<$25 & 0.217 & 0.777 & 0.447 & 0.079 & 77 & 2{,}941 \\
 & 25--34 & 0.207 & 0.762 & 0.467 & 0.072 & 157 & 12{,}238 \\
 & 35--44 & 0.235 & 0.762 & 0.484 & 0.101 & 111 & 6{,}149 \\
 & 45--54 & 0.264 & 0.763 & 0.507 & 0.111 & 114 & 6{,}505 \\
 & 55--64 & 0.263 & 0.761 & 0.505 & 0.093 & 97 & 4{,}655 \\
 & 65--74 & 0.310 & 0.780 & 0.510 & 0.117 & 87 & 3{,}713 \\
 & 75+ & 0.317 & 0.786 & 0.522 & 0.145 & 28 & 373 \\
\midrule
Sex & Female & 0.248 & 0.771 & 0.497 & 0.093 & 367 & 67{,}199 \\
 & Male & 0.237 & 0.762 & 0.498 & 0.083 & 306 & 46{,}805 \\
\midrule
Education & Low & 0.271 & 0.760 & 0.497 & 0.109 & 51 & 1{,}268 \\
 & High & 0.215 & 0.769 & 0.496 & 0.076 & 261 & 33{,}965 \\
 & Middle & 0.205 & 0.766 & 0.498 & 0.066 & 361 & 65{,}196 \\
\midrule
Employ. & Homemaker & 0.275 & 0.768 & 0.453 & 0.120 & 74 & 2{,}718 \\
 & Retired & 0.274 & 0.779 & 0.510 & 0.093 & 154 & 11{,}819 \\
 & Working & 0.241 & 0.758 & 0.470 & 0.088 & 369 & 68{,}014 \\
 & Unemployed & 0.232 & 0.778 & 0.476 & 0.091 & 33 & 527 \\
 & Student & 0.171 & 0.794 & 0.436 & 0.065 & 38 & 699 \\
\midrule
Marital st. & Cohabiting & 0.273 & 0.766 & 0.375 & 0.152 & 38 & 721 \\
 & Married & 0.265 & 0.754 & 0.448 & 0.100 & 360 & 64{,}753 \\
 & Widowed & 0.251 & 0.787 & 0.414 & 0.100 & 65 & 2{,}108 \\
 & Div. / sep. & 0.233 & 0.795 & 0.444 & 0.079 & 73 & 2{,}631 \\
 & Single & 0.202 & 0.776 & 0.425 & 0.071 & 136 & 9{,}220 \\
\midrule
Wave & 1995--1997 & 0.246 & 0.762 & 0.507 & 0.086 & 221 & 24{,}218 \\
 & 2005--2006 & 0.250 & 0.767 & 0.495 & 0.100 & 208 & 21{,}839 \\
 & 2017--2018 & 0.249 & 0.771 & 0.519 & 0.088 & 244 & 29{,}700 \\
\bottomrule
\end{tabular}
\end{adjustbox}
\caption{Demographic-subgroup Population Fidelity. Metrics are recomputed within each level and then averaged across the 148 run--mode combinations, excluding the archived and fresh Mixtral reference series. Cells and pairs report the mean numbers contributing
to each estimate. Estimates based on fewer cells, especially structure,
should be interpreted cautiously. Two country-level estimates have
zero-variance model distance vectors and receive a structure score of
zero under the flatness rule in Appendix~\ref{app:metrics}.}
\label{tab:families-app}
\end{table}

\begin{figure}[htb]
\centering
\includegraphics[height=8cm,width=0.85\columnwidth]
{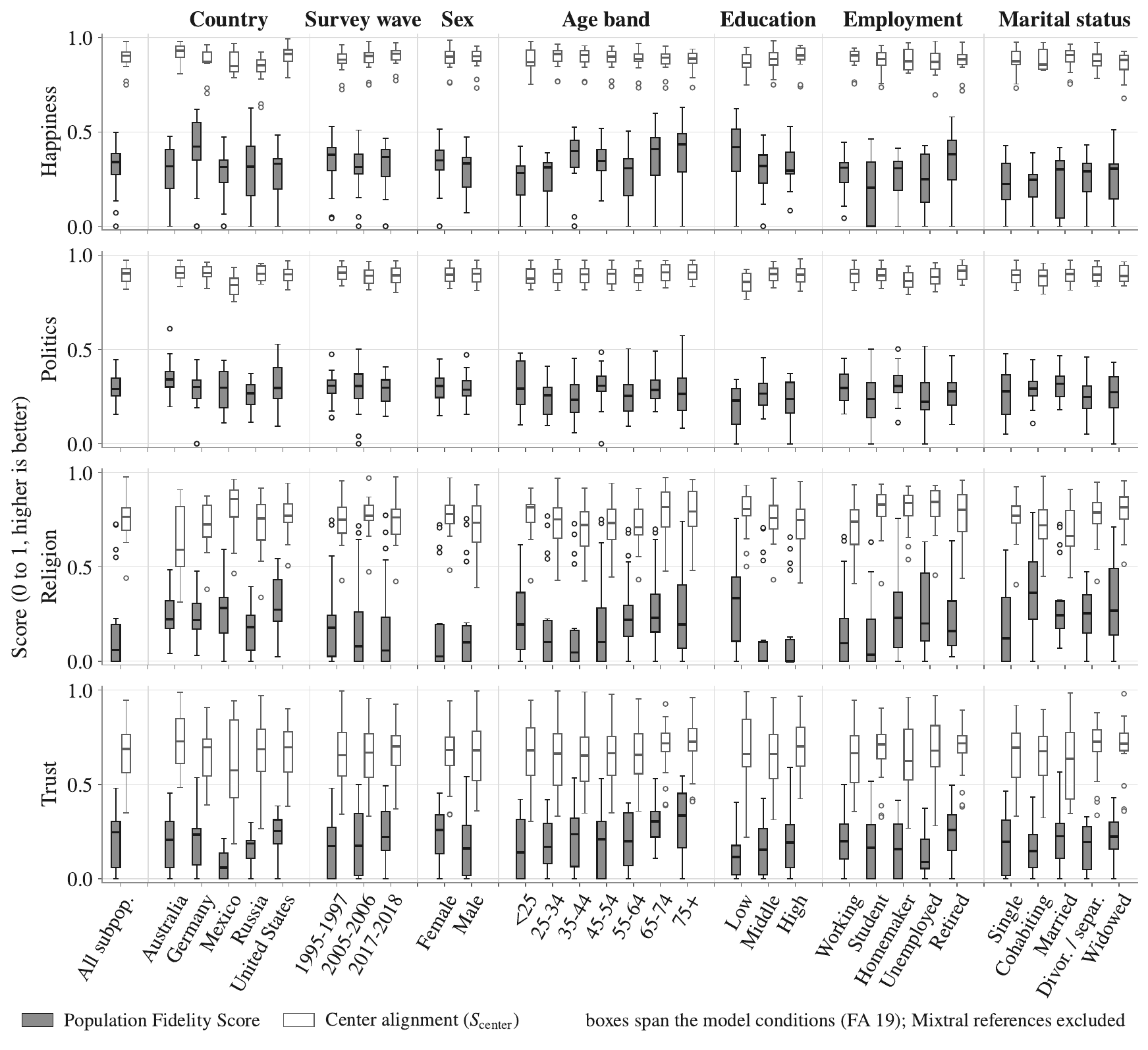}
\caption{
PFS and center alignment by demographic level under FA elicitation,
shown separately by question. Filled boxes show the distribution of PFS
and hollow boxes the distribution of center alignment across 19 model
conditions; Mixtral reference series are excluded. Demographic columns
follow Figure~\ref{fig:subgroups-main}.
}
\label{fig:subgroups-app}
\end{figure}

\paragraph{Center alignment and PFS by demographic level.}
Figure~\ref{fig:subgroups-main} compares PFS and center alignment across
demographic levels for happiness, while Figure~\ref{fig:subgroups-app}
extends the comparison to all four questions. Center alignment varies much
less across demographic levels than PFS. Excluding the Mixtral reference
series, the median center alignment across the 30 levels ranges from
$0.835$ for cohabiting respondents to $0.878$ for Australia, whereas
median PFS ranges from $0.143$ for students to $0.343$ for respondents
aged 75 and older. For happiness alone, the corresponding ranges are
$0.848$--$0.928$ for center alignment and $0.178$--$0.436$ for PFS.

This difference is also visible within particular demographic levels.
Students, unemployed respondents, respondents aged 25--34, and Russian
cells have median center alignment within $0.01$ of the pooled value
($0.860$), but substantially lower median PFS ($0.143$--$0.238$, compared
with $0.254$ for the pooled population). Among the 4{,}440 subgroup-level
scores excluding Mixtral references, 295 combine center alignment of at
least $0.85$ with PFS of at most $0.10$. Structure is the limiting
component in 278 of these cases. A model can therefore approximate a
subgroup center while recovering little of the pattern of
differences among the cells that compose it.

Across model conditions, PFS and center alignment are positively but only
moderately associated. Excluding Mixtral references, the mean Spearman
correlation across demographic levels ranges from $0.263$ for political
ideology under NTP to $0.425$ for social trust under NTP. At the pooled
population level, correlations range from $0.302$ for religious attendance
under FA to $0.674$ for political ideology under FA. Bootstrap intervals
are wide, particularly at the population level. Pearson correlations give
the same qualitative pattern. These results support treating center
alignment and Population Fidelity as related but distinct quantities.

\begin{figure*}[htb]
\centering
\includegraphics[height=10cm,width=0.70\textwidth]
{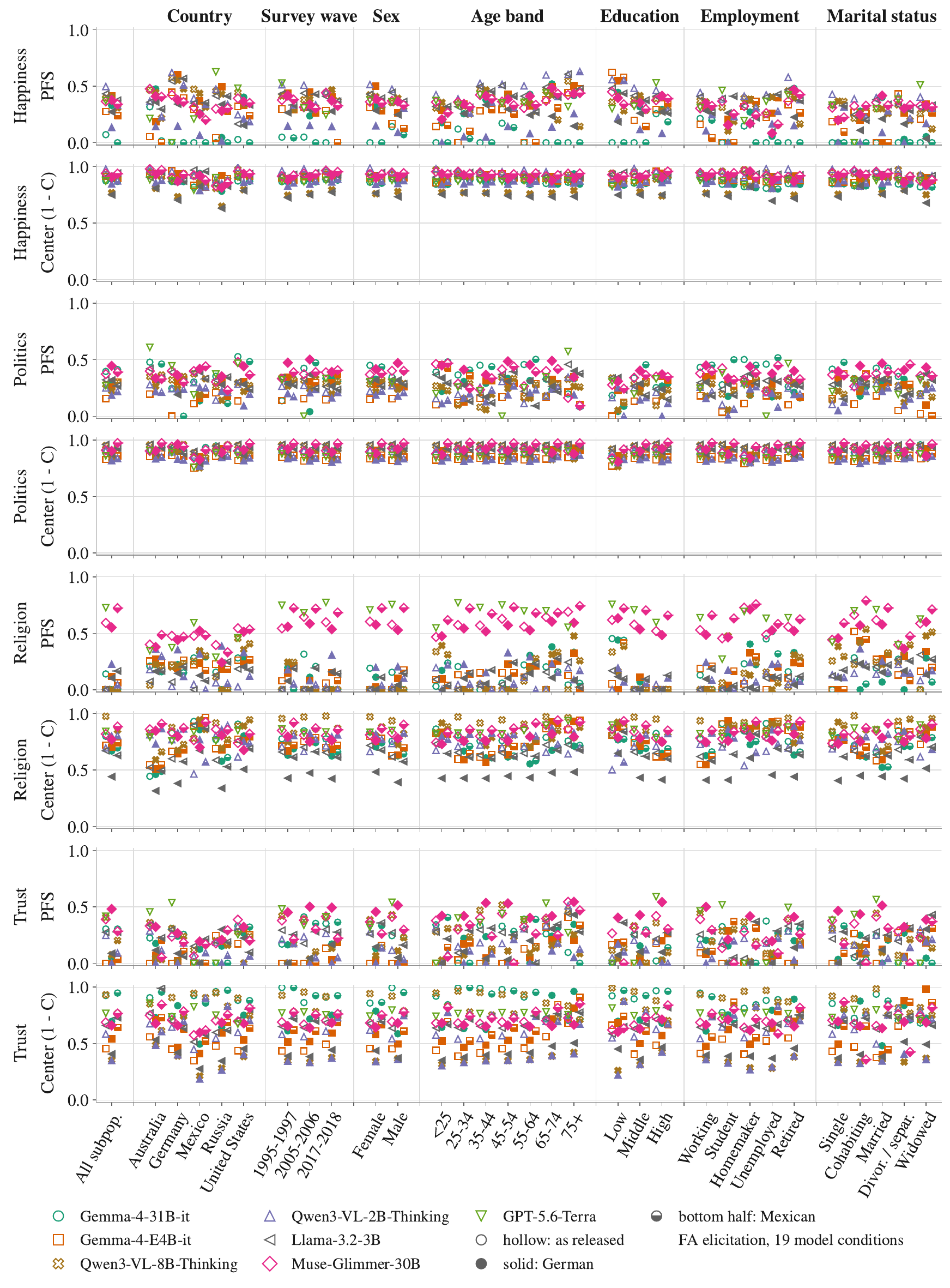}
\caption{
PFS and center alignment by demographic level for individual model
conditions under FA elicitation, shown separately by question. Within
each question, the upper row shows PFS and the lower row center alignment;
demographic columns follow Figure~\ref{fig:subgroups-main}. Color and
marker shape identify the model. Hollow markers denote as-released
conditions, filled markers German fine-tuning, and half-filled markers
Mexican fine-tuning.
}
\label{fig:subgroups-models}
\end{figure*}

The converse pattern also occurs, although less often. A PFS of at least
$0.40$ together with center alignment of at most $0.75$ occurs 84 times,
60 of them for social trust. Fine-tuned conditions account for 60 of these
cases. Thus, relatively high subgroup fidelity can also coexist with weaker center
alignment. Figure~\ref{fig:subgroups-models}
shows that these differences are distributed across model conditions rather
than being driven by a single model.

\subsection{Additional Population Fidelity Figures}\label{app:figures}

\begin{figure*}[htb]
\centering
\includegraphics[height=5cm,width=0.90\textwidth]
{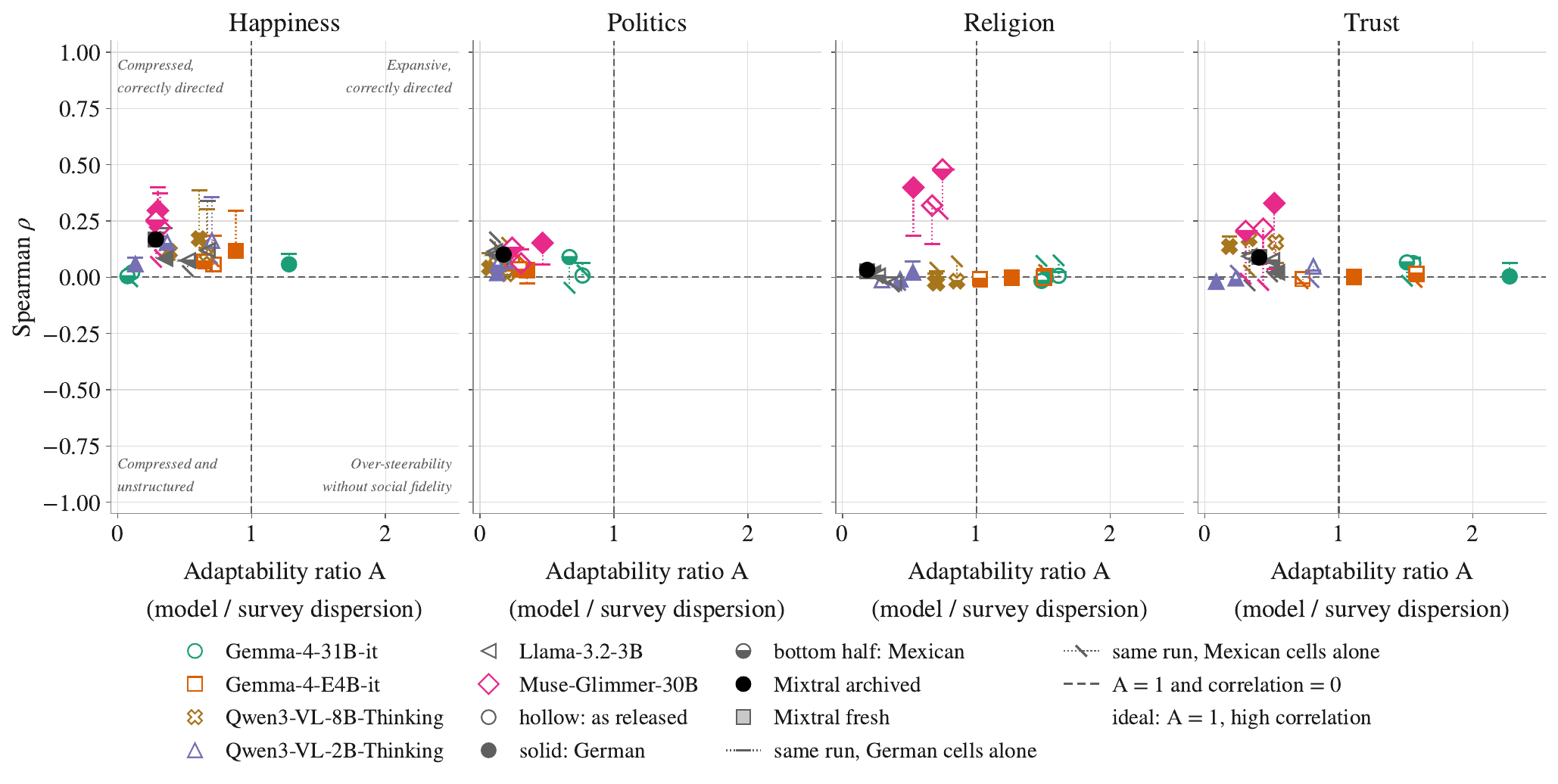}
\caption{
Adaptability and structure under NTP elicitation, shown separately by
question. The horizontal axis reports the adaptability ratio $A$, and the
vertical axis the untransformed structure correlation $\rho$. Dashed lines
mark $A=1$ and $\rho=0$. Dotted connectors link each population-wide
estimate to the corresponding estimate recomputed within the target-country
subset: Germany for German fine-tuning and Mexico for Mexican
fine-tuning. As-released conditions show both target-country estimates,
whereas fine-tuned conditions show only the estimate for their target
country. Conditions with $A>1$ and $\rho\approx0$ exhibit amplified
between-group variation without recovering the observed structure of group
differences.
}
\label{fig:structure-app}
\end{figure*}

Figure~\ref{fig:structure-app} relates the amount of model variation to
its correspondence with observed group differences.
Figures~\ref{fig:shift-app} and~\ref{fig:components-app} extend the
main-text fine-tuning and component figures to the remaining FA
questions and the cross-question summary. Complete NTP and FA figures
are available in the accompanying repository.

\begin{figure}[tbh]
\centering
\begin{subfigure}[t]{0.45\columnwidth}
\centering
\includegraphics[height = 7.0cm,width=\linewidth]
{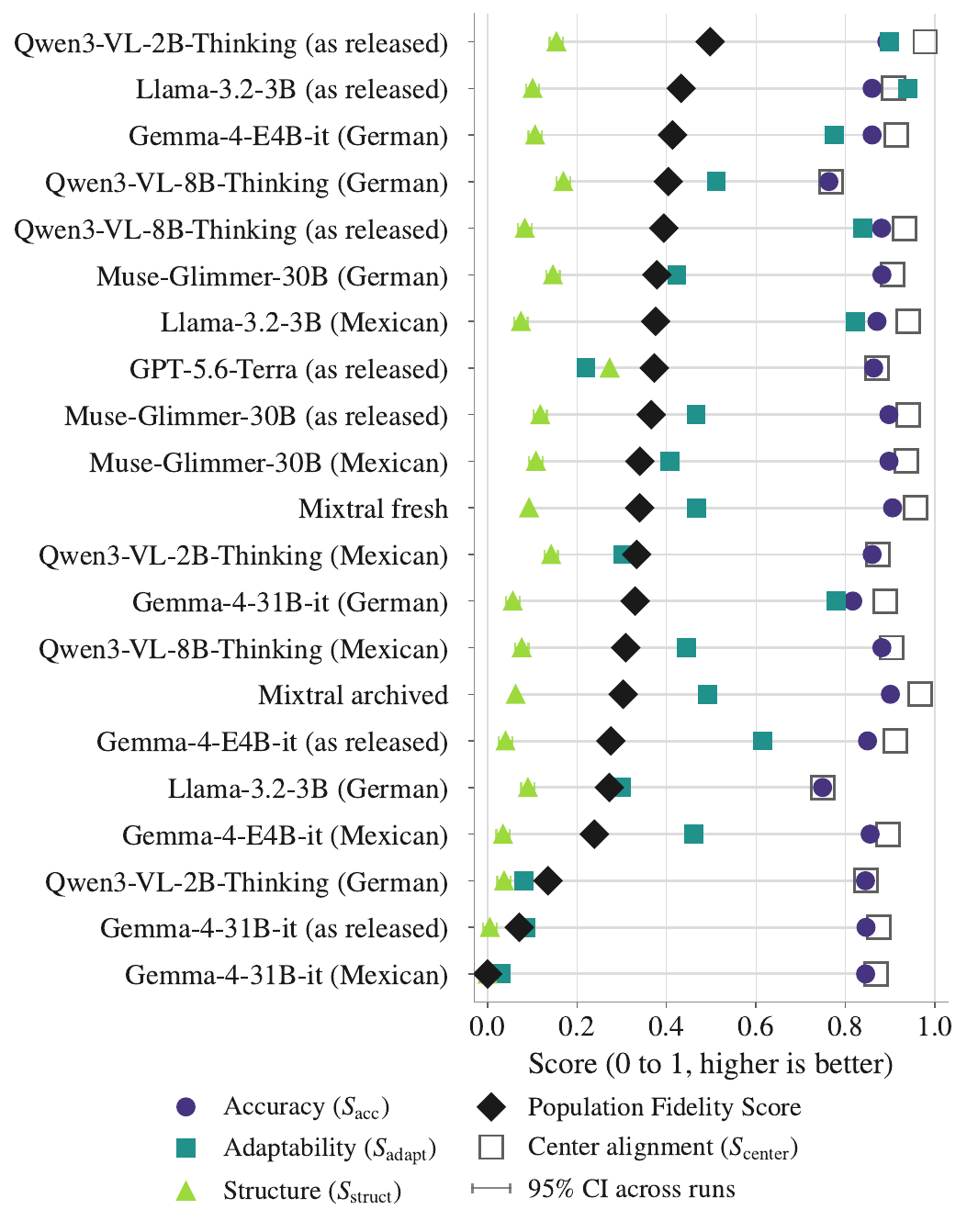}
\caption{Component profile, all retained cells.}
\end{subfigure}
\hfill
\begin{subfigure}[t]{0.54\columnwidth}
\centering
\includegraphics[height = 7.0cm,width=\linewidth]
{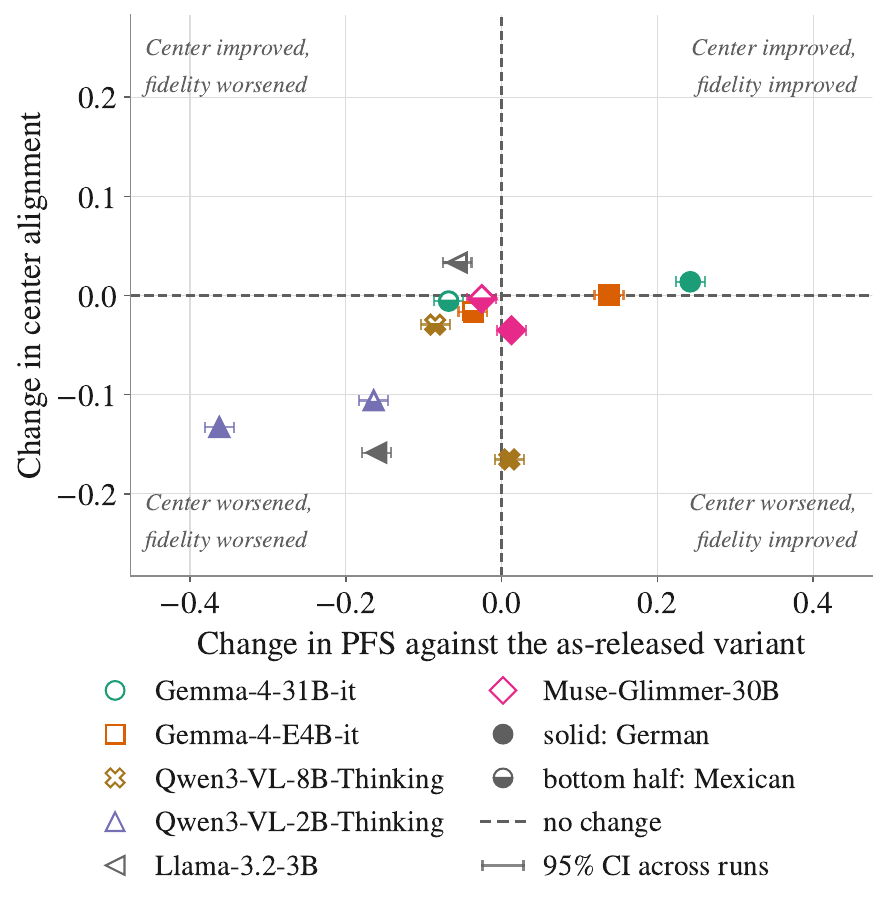}
\caption{Fine-tuning shifts, all retained cells.}
\end{subfigure}\vspace{-0.3cm}
\caption{
Population Fidelity for happiness under FA elicitation on all retained cells.
\textbf{(a)} Component scores for all 21 conditions, ordered by PFS; diamonds
show PFS and hollow squares center alignment.
\textbf{(b)} Fine-tuned minus as-released changes computed on the cells retained
by both conditions. Figure~\ref{fig:happiness-results} shows the corresponding
target-country results. Plotting conventions are as in
Figure~\ref{fig:happiness-results}.
}
\label{fig:happiness-all-app}
\end{figure}

\begin{figure*}[htb]
\centering
\includegraphics[width=0.40\textwidth]
{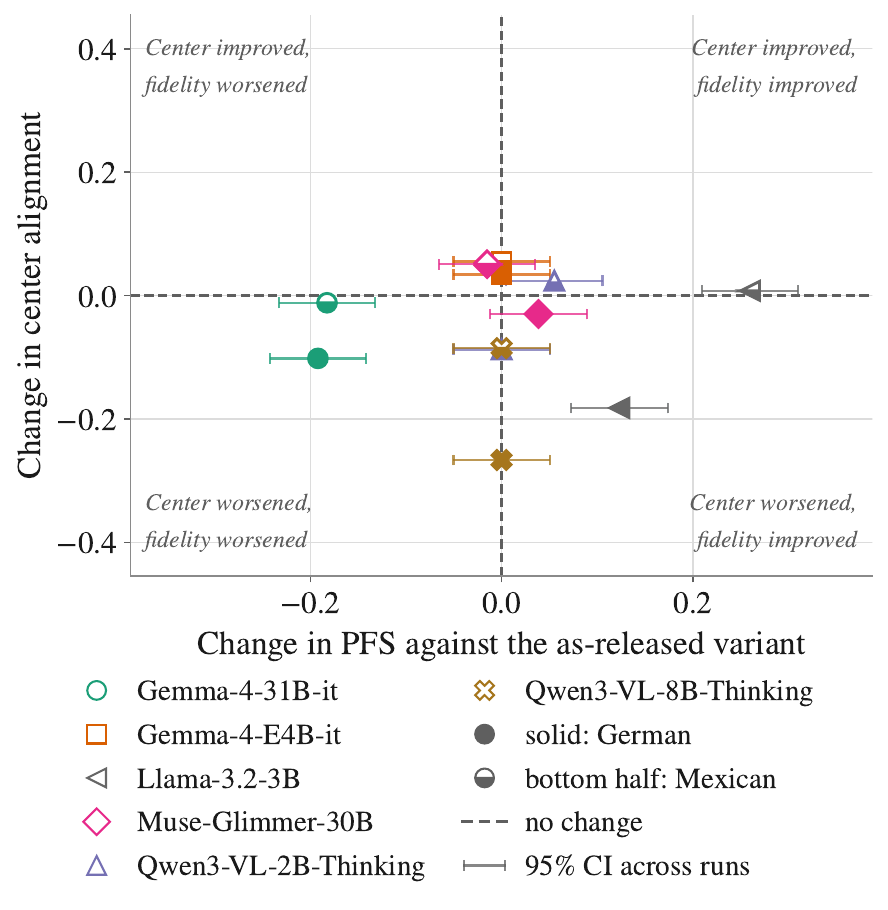}\hfill
\includegraphics[width=0.40\textwidth]
{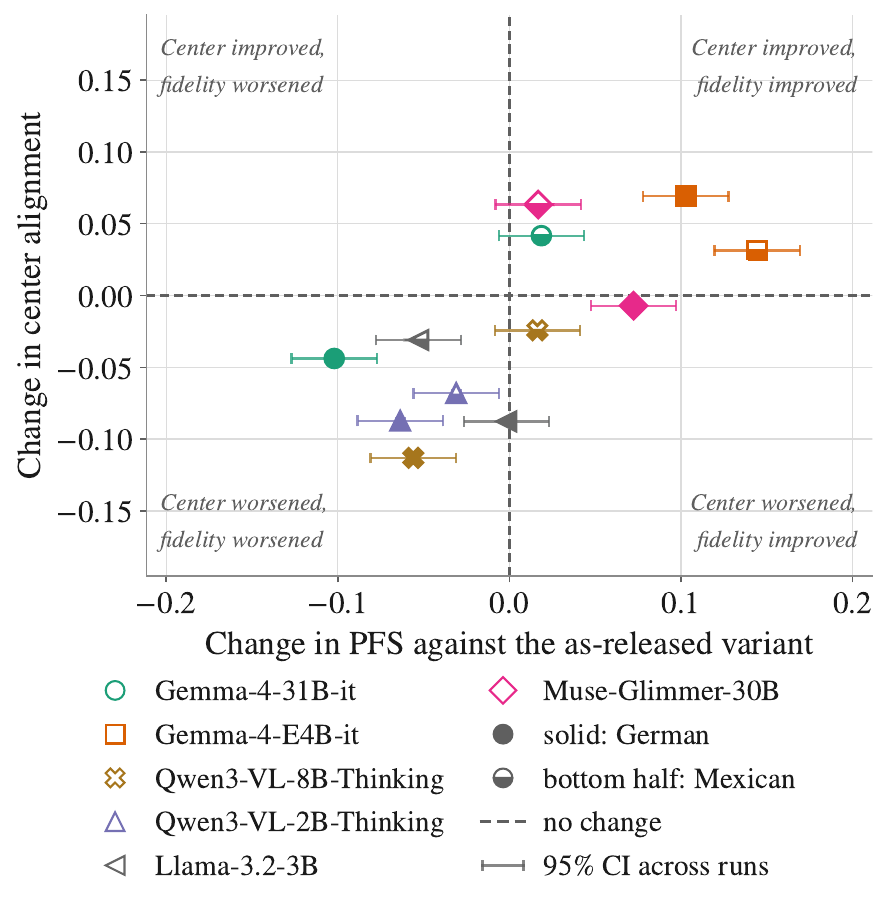}\\[2pt]
\includegraphics[width=0.40\textwidth]
{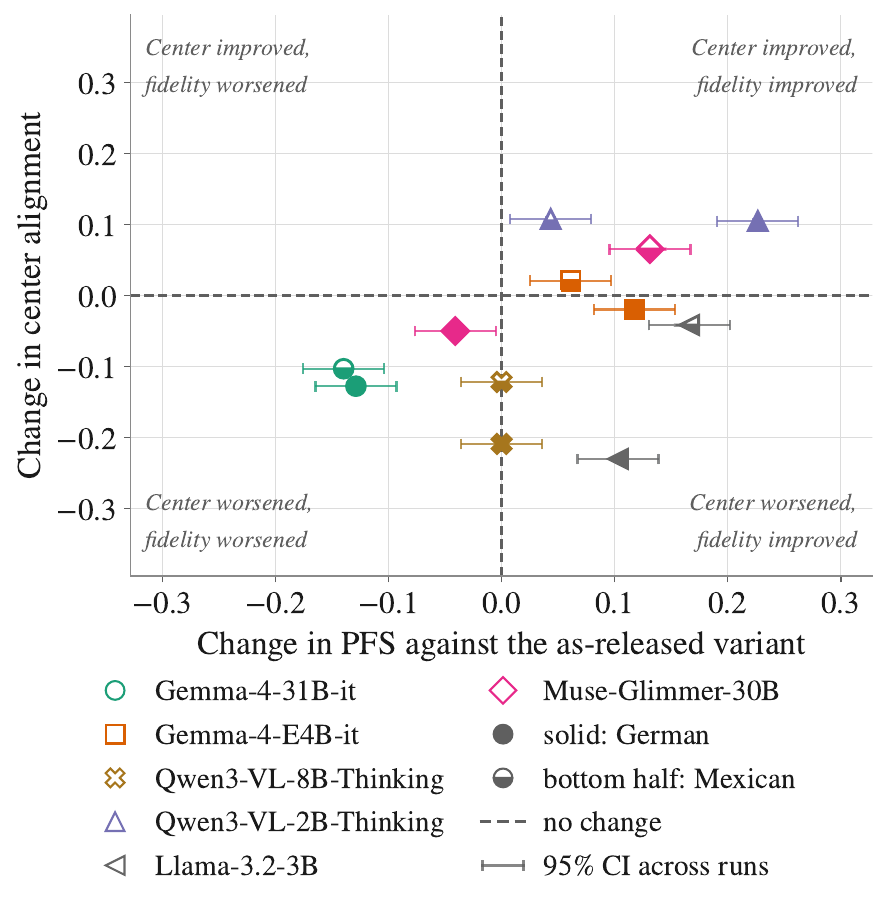}\hfill
\includegraphics[width=0.40\textwidth]
{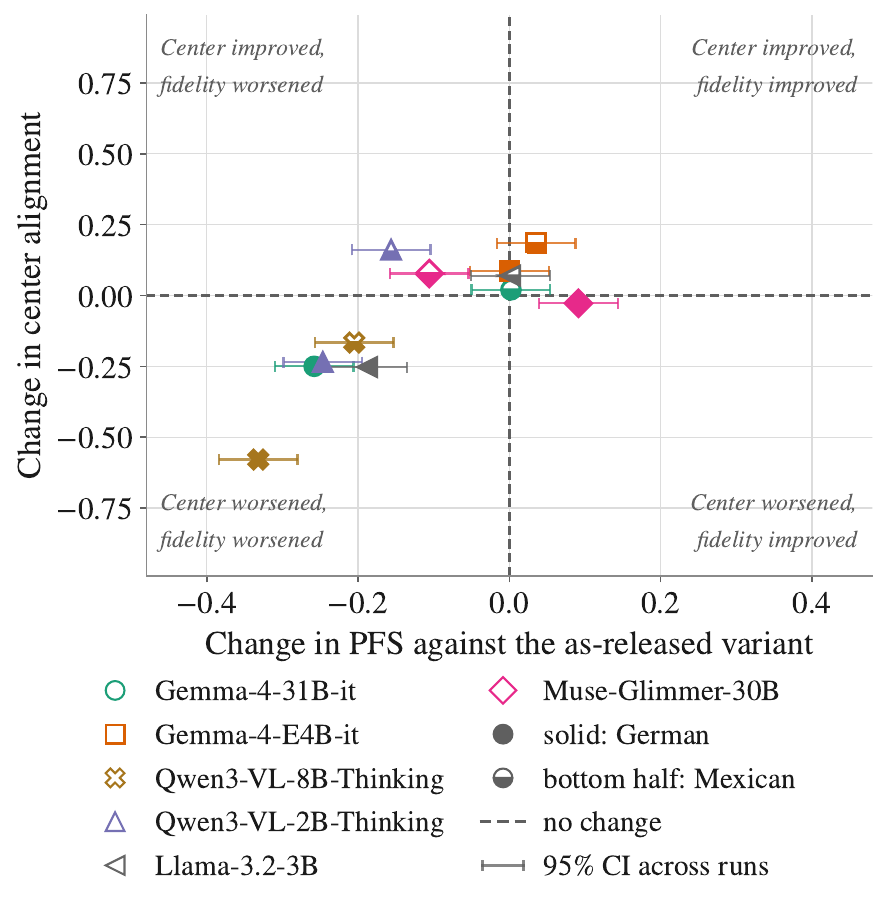}

\caption{Changes in PFS and center alignment after cultural fine-tuning
under FA elicitation, for the cross-question summary, political
ideology, religious attendance, and social trust. Each point compares a
fine-tuned condition with its corresponding as-released model. Movement
to the right indicates improved PFS, and upward movement indicates
improved center alignment. The happiness result appears in
Figure~\ref{fig:shift}.}
\label{fig:shift-app}
\end{figure*}

\begin{figure*}[htb]
\centering
\begin{subfigure}[t]{0.33\textwidth}
\centering
\includegraphics[width=\linewidth]
{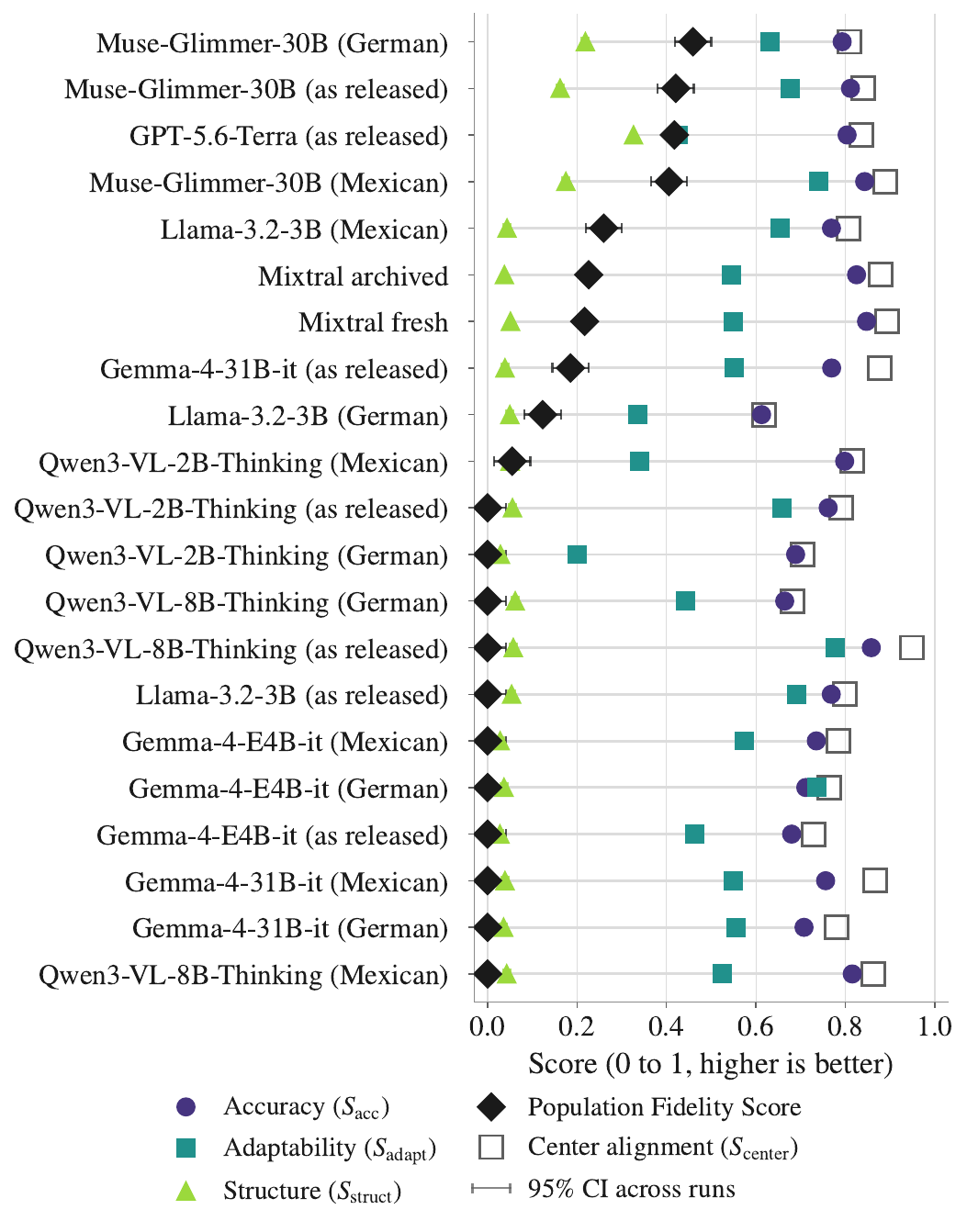}
\caption{Cross-question summary.}
\end{subfigure}
\hspace{0.02\textwidth}
\begin{subfigure}[t]{0.33\textwidth}
\centering
\includegraphics[width=\linewidth]
{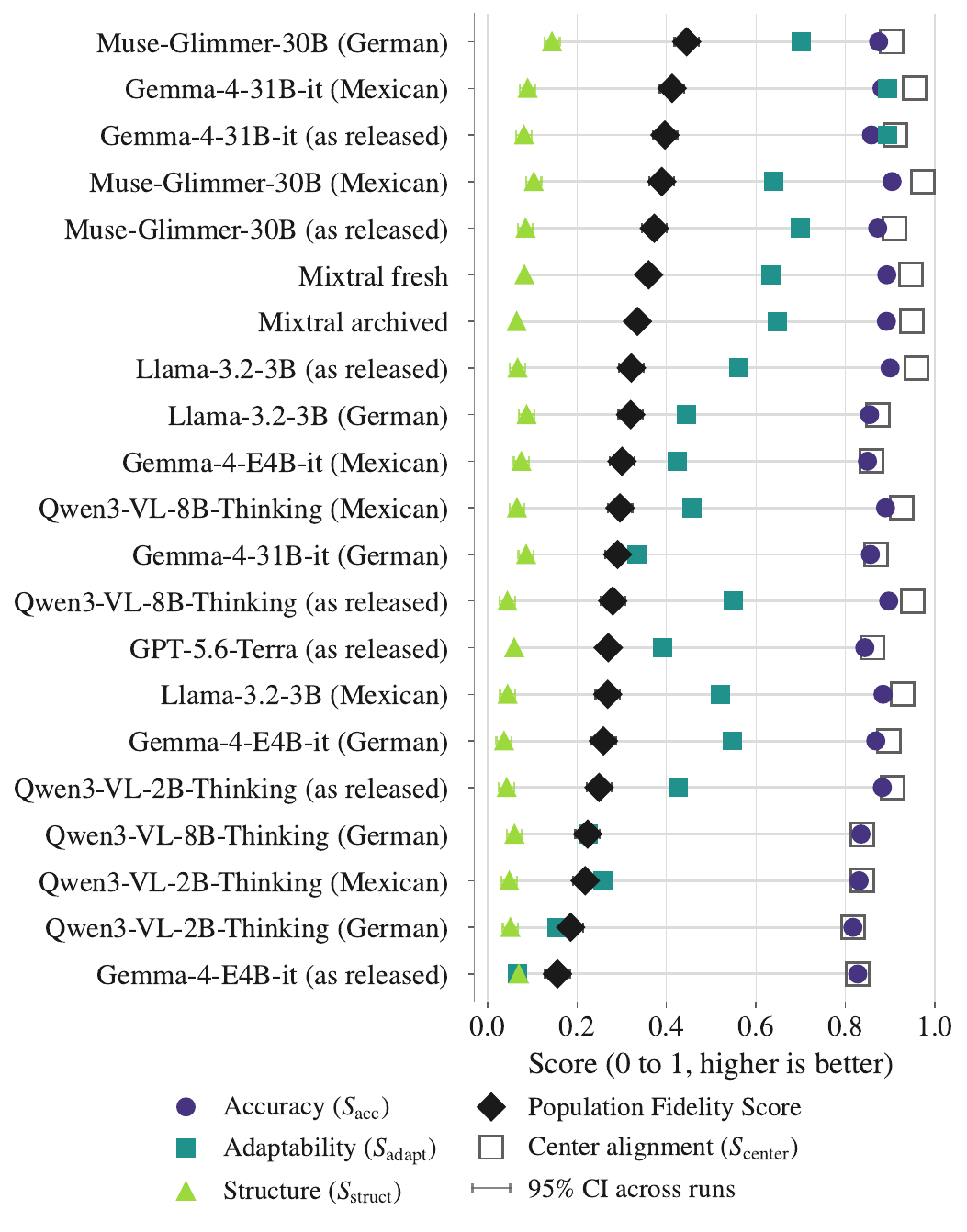}
\caption{Political ideology.}
\end{subfigure}

\vspace{2pt}
\begin{subfigure}[t]{0.33\textwidth}
\centering
\includegraphics[width=\linewidth]
{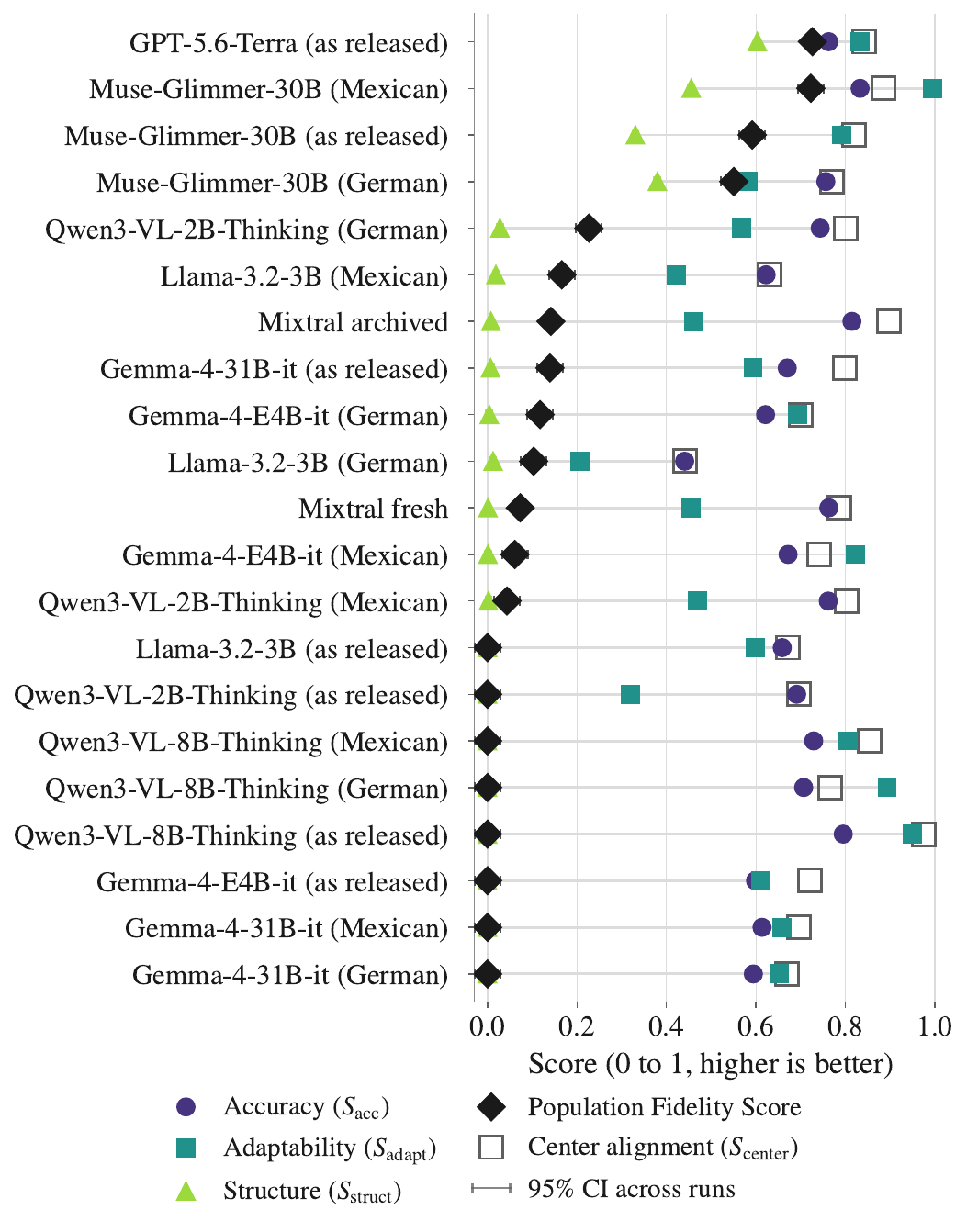}
\caption{Religious attendance.}
\end{subfigure}
\hspace{0.02\textwidth}
\begin{subfigure}[t]{0.33\textwidth}
\centering
\includegraphics[width=\linewidth]
{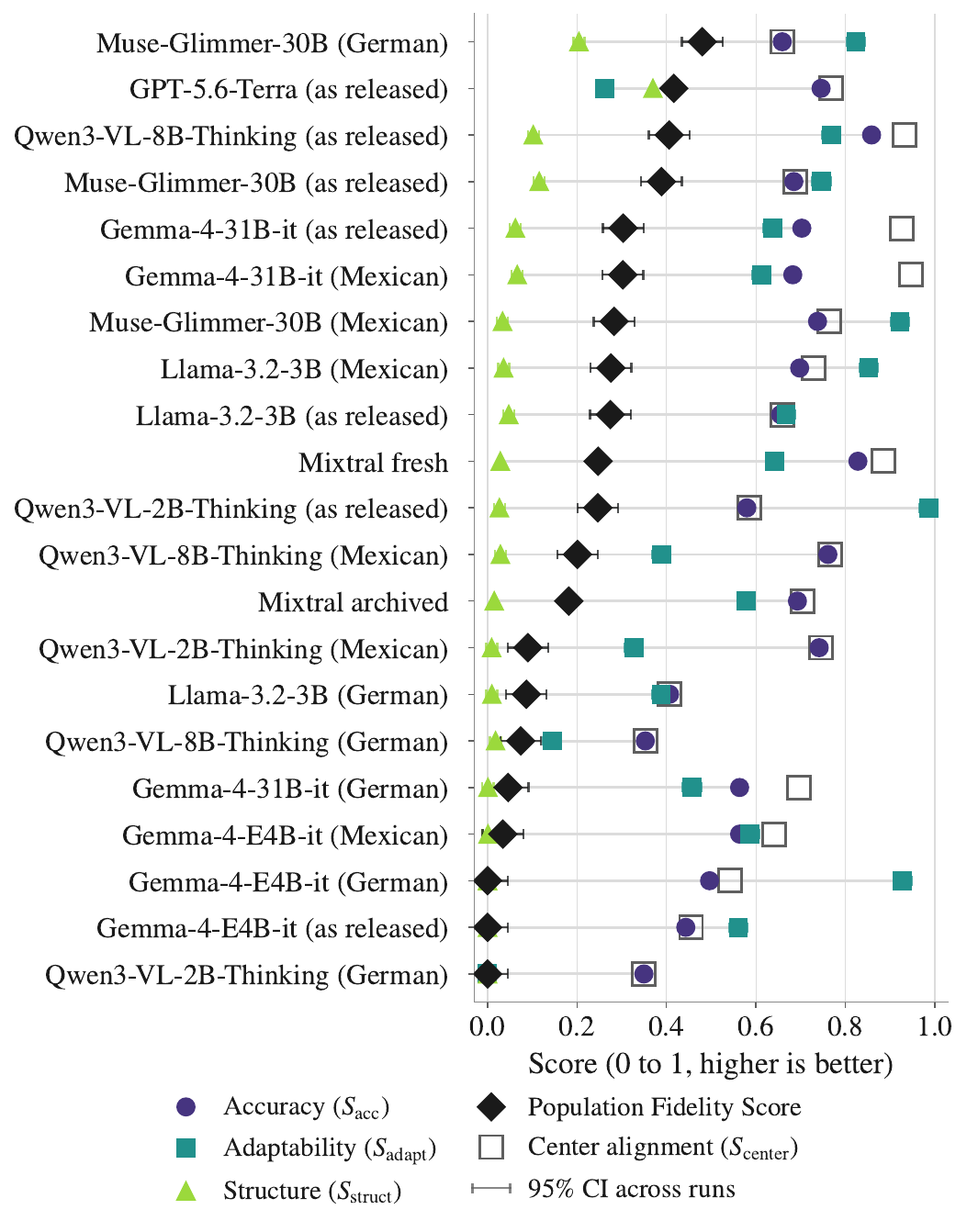}
\caption{Social trust.}
\end{subfigure}
\caption{Population Fidelity components under FA elicitation, for the
cross-question summary, political ideology, religious attendance, and
social trust. Accuracy is generally the strongest component, whereas structure is the
weakest and most often limiting; happiness is shown in
Figure~\ref{fig:components} and movement intervals in Table~\ref{tab:replicates}. Fine-tuned conditions are labelled German ($\mathcal{F}_{\text{de}}$) and Mexican ($\mathcal{F}_{\text{mx}}$). 
}
\label{fig:components-app}
\end{figure*}

\section{Robustness and Sensitivity}\label{app:robustness}

\subsection{Stability Across Repeated Full-Answer Runs}
\label{app:replicates}

Because FA elicitation is stochastic, repeated runs can produce different sampled responses and therefore slightly different scores. We quantify this run-to-run variation for the 18 locally evaluated conditions---the as-released, $\mathcal{F}_{\text{de}}$, and $\mathcal{F}_{\text{mx}}$ variants of six models---on all four questions, yielding 72 condition--question pairs evaluated three times each. Across repeats, model weights, fine-tuned variants, prompts, decoding settings, and library versions are held fixed; only the random seed stream changes. NTP is deterministic under our decoding configuration and is therefore not repeated.

For each condition--question pair $i$, we compute the mean score across its $c=3$ FA runs and a 95\% $t$-interval,
\[
\bar{x}_i \pm h_i,
\qquad
h_i =
t_{0.975,\,c-1}\frac{\sigma_i}{\sqrt{c}},
\]
where $\sigma_i$ is the standard deviation across the three runs and $h_i$ is the corresponding CI half-width. Because $c=3$, $t_{0.975,2}=4.30$, so individual intervals can be wide even when the observed run-to-run variation is modest.

To summarize run-to-run uncertainty across a set of $n$ condition--question pairs, we pool their variance estimates and report the corresponding CI half-width,
\[
h_{\mathrm{pool}}
=
\frac{t_{0.975,\nu}}{t_{0.975,2}}
\sqrt{
\frac{1}{n}\sum_{i=1}^{n} h_i^2
},
\qquad
\nu=n(c-1).
\]
Table~\ref{tab:replicates} reports pooled half-widths across the 18 condition--question pairs for each question and across all 72 pairs, while Table~\ref{tab:replicates-models} reports the individual CI half-widths.

\begin{table}[htb]
\centering
\small
\begin{tabular}{lrrrrrr}
\toprule
Question & $S_{\mathrm{acc}}$ & $S_{\mathrm{adapt}}$ & $S_{\mathrm{struct}}$ & PFS & $S_{\mathrm{center}}$ & $\rho$ (PFS) \\
\midrule
Happiness & 0.001 & 0.013 & 0.015 & 0.019 & 0.001 & 0.960 \\
Political ideology & 0.001 & 0.008 & 0.017 & 0.027 & 0.001 & 0.920 \\
Religious attendance & 0.002 & 0.008 & 0.007 & 0.029 & 0.002 & 0.922 \\
Social trust & 0.002 & 0.021 & 0.012 & 0.045 & 0.002 & 0.926 \\
\midrule
All questions & 0.002 & 0.013 & 0.013 & 0.031 & 0.002 & 0.972 \\
\bottomrule
\end{tabular}
\caption{
Run-to-run uncertainty across repeated FA runs. Entries are pooled 95\% CI half-widths, $h_{\mathrm{pool}}$, across the condition--question pairs in each row. The final column reports the mean of the three pairwise Spearman rank correlations in PFS between repeated runs; for ``All questions,'' correlations are computed after pooling the condition--question observations across the four questions. Smaller CI half-widths and larger rank correlations indicate greater stability across repeated elicitation.
}
\label{tab:replicates}
\end{table}

Run-to-run variation is small relative to differences between model conditions. In the variance decomposition, the run-to-run variance component is $2.9\%$ of the between-condition component for PFS, corresponding to an intraclass correlation of $0.971$. The corresponding ratios are $1.8\%$ for $S_{\mathrm{struct}}$ and below $0.3\%$ for $S_{\mathrm{acc}}$, $S_{\mathrm{adapt}}$, and $S_{\mathrm{center}}$. Thus, accuracy and center alignment are highly stable across repeats, while structure has the largest run-to-run variance relative to between-condition differences among the PFS components.

The qualitative conclusions are similarly stable. In 70 of the 72 condition--question pairs, the same component is limiting in all three runs: $S_{\mathrm{struct}}$ in 69 pairs and $S_{\mathrm{adapt}}$ in one. This reinforces the main result that structure, rather than accuracy, is typically the limiting dimension of population fidelity. Among the as-released conditions, the adaptability ratio $A$ never crosses one across runs, so no condition switches between compression and amplification. Model-condition rankings are also stable: pairwise Spearman correlations in PFS between runs range from $0.889$ to $0.965$ across questions. Finally, 63 of the 72 pairs retain identical cell sets in all three runs, so for most pairs the observed movement reflects sampled responses rather than changes in evaluation support.

Much of the remaining PFS sensitivity occurs near the clipping boundary for $S_{\mathrm{struct}}$. Because $S_{\mathrm{struct}}=\max(0,\rho)$, a small run-to-run change that moves $\rho$ across zero can move PFS between zero and a positive value. Nineteen of the 72 condition--question pairs touch this boundary across runs. Near-zero PFS values should therefore not be interpreted as fine-grained rankings: they primarily indicate little or no recovered population structure.

The paired fine-tuning effects are also stable across repeated FA runs (Table~\ref{tab:replicates-paired}). Across the 24 model--question comparisons for each cultural target, the run-specific mean $\Delta\mathrm{PFS}$ on all retained cells and the run-specific mean $\Delta S_{\mathrm{center}}$ in both scopes vary by at most $0.005$. For population-wide $\Delta\mathrm{PFS}$, the sign is unchanged across all three runs in 23 of 24 $\mathcal{F}_{\text{de}}$ comparisons and 18 of 24 $\mathcal{F}_{\text{mx}}$ comparisons. The sign of $\Delta S_{\mathrm{center}}$ is unchanged in all 24 comparisons for each target and scope.

Within the target countries, we summarize individual-comparison run-to-run uncertainty by pooling the 24 model--question-specific variance estimates using the same procedure above ($n=24$, $\nu=48$). The resulting pooled 95\% CI half-width is $0.073$ for $\Delta\mathrm{PFS}$ in Germany and $0.110$ in Mexico. The corresponding pooled half-widths for $\Delta S_{\mathrm{center}}$ are only $0.005$ and $0.003$, respectively. Thus, center-alignment shifts are substantially more stable to FA sampling than target-country PFS changes. These intervals quantify uncertainty attributable to repeated FA sampling with the survey, prompts, model parameters, and evaluation definitions held fixed; they are not confidence intervals for the mean effect across model--question comparisons. Appendix~\ref{app:sensitivity} separately examines sensitivity to the metric specification.

\begin{table}[htb]
\centering
\scriptsize
\setlength{\tabcolsep}{3pt}
\begin{adjustbox}{max width=\textwidth}
\begin{tabular}{llrrrrrrrr}
\toprule
& & \multicolumn{4}{c}{$\mathcal{F}_{\text{de}}$} & \multicolumn{4}{c}{$\mathcal{F}_{\text{mx}}$} \\
\cmidrule(lr){3-6}\cmidrule(lr){7-10}
& & \multicolumn{2}{c}{All retained cells} & \multicolumn{2}{c}{Germany} & \multicolumn{2}{c}{All retained cells} & \multicolumn{2}{c}{Mexico} \\
\cmidrule(lr){3-4}\cmidrule(lr){5-6}\cmidrule(lr){7-8}\cmidrule(lr){9-10}
Model & Question & $\Delta\mathrm{PFS}$ & $\Delta S_{\mathrm{center}}$ & $\Delta\mathrm{PFS}$ & $\Delta S_{\mathrm{center}}$ & $\Delta\mathrm{PFS}$ & $\Delta S_{\mathrm{center}}$ & $\Delta\mathrm{PFS}$ & $\Delta S_{\mathrm{center}}$ \\
\midrule
Gemma-4-31B-it & Happiness & $+0.242 \pm 0.007$ & $+0.014 \pm 0.001$ & $+0.376 \pm 0.011$ & $-0.005 \pm 0.002$ & $-0.068 \pm 0.025$ & $-0.006 \pm 0.001$ & $0.000 \pm 0.000$ & $-0.003 \pm 0.004$ \\
 & Political ideology & $-0.102 \pm 0.057$ & $-0.044 \pm 0.001$ & $+0.278 \pm 0.226$ & $-0.014 \pm 0.001$ & $+0.019 \pm 0.069$ & $+0.042 \pm 0.002$ & $-0.042 \pm 0.385$ & $+0.087 \pm 0.007$ \\
 & Religious attendance & $-0.129 \pm 0.027$ & $-0.128 \pm 0.004$ & $+0.014 \pm 0.042$ & $-0.167 \pm 0.005$ & $-0.140 \pm 0.024$ & $-0.103 \pm 0.003$ & $+0.010 \pm 0.012$ & $-0.045 \pm 0.006$ \\
 & Social trust & $-0.258 \pm 0.098$ & $-0.250 \pm 0.001$ & $-0.070 \pm 0.044$ & $+0.090 \pm 0.006$ & $+0.002 \pm 0.003$ & $+0.020 \pm 0.003$ & $0.000 \pm 0.000$ & $-0.065 \pm 0.003$ \\
 & All questions & $-0.062 \pm 0.010$ & $-0.102 \pm 0.001$ & $+0.150 \pm 0.068$ & $-0.024 \pm 0.003$ & $-0.047 \pm 0.014$ & $-0.012 \pm 0.001$ & $-0.008 \pm 0.095$ & $-0.006 \pm 0.003$ \\
\addlinespace
Gemma-4-E4B-it & Happiness & $+0.138 \pm 0.029$ & $+0.001 \pm 0.000$ & $+0.154 \pm 0.020$ & $-0.029 \pm 0.003$ & $-0.037 \pm 0.014$ & $-0.016 \pm 0.000$ & $+0.032 \pm 0.037$ & $-0.013 \pm 0.003$ \\
 & Political ideology & $+0.103 \pm 0.066$ & $+0.069 \pm 0.003$ & $+0.307 \pm 0.138$ & $+0.041 \pm 0.004$ & $+0.144 \pm 0.023$ & $+0.031 \pm 0.001$ & $+0.086 \pm 0.437$ & $+0.082 \pm 0.001$ \\
 & Religious attendance & $+0.118 \pm 0.030$ & $-0.020 \pm 0.003$ & $-0.026 \pm 0.008$ & $-0.017 \pm 0.005$ & $+0.061 \pm 0.146$ & $+0.021 \pm 0.005$ & $-0.051 \pm 0.121$ & $+0.079 \pm 0.003$ \\
 & Social trust & $0.000 \pm 0.000$ & $+0.087 \pm 0.004$ & $+0.021 \pm 0.090$ & $+0.014 \pm 0.002$ & $+0.035 \pm 0.152$ & $+0.185 \pm 0.006$ & $0.000 \pm 0.000$ & $+0.172 \pm 0.010$ \\
 & All questions & $+0.090 \pm 0.014$ & $+0.034 \pm 0.001$ & $+0.114 \pm 0.062$ & $+0.002 \pm 0.002$ & $+0.051 \pm 0.026$ & $+0.055 \pm 0.002$ & $+0.017 \pm 0.095$ & $+0.080 \pm 0.003$ \\
\addlinespace
Qwen3-VL-8B-Thinking & Happiness & $+0.010 \pm 0.006$ & $-0.165 \pm 0.002$ & $-0.023 \pm 0.059$ & $-0.165 \pm 0.007$ & $-0.085 \pm 0.079$ & $-0.029 \pm 0.004$ & $-0.158 \pm 0.106$ & $-0.069 \pm 0.003$ \\
 & Political ideology & $-0.056 \pm 0.025$ & $-0.113 \pm 0.003$ & $-0.080 \pm 0.106$ & $-0.038 \pm 0.004$ & $+0.016 \pm 0.072$ & $-0.024 \pm 0.002$ & $-0.044 \pm 0.109$ & $-0.042 \pm 0.002$ \\
 & Religious attendance & $0.000 \pm 0.000$ & $-0.209 \pm 0.007$ & $+0.010 \pm 0.089$ & $-0.203 \pm 0.013$ & $0.000 \pm 0.000$ & $-0.122 \pm 0.008$ & $+0.048 \pm 0.125$ & $+0.079 \pm 0.014$ \\
 & Social trust & $-0.332 \pm 0.193$ & $-0.579 \pm 0.011$ & $-0.288 \pm 0.344$ & $-0.517 \pm 0.022$ & $-0.205 \pm 0.055$ & $-0.166 \pm 0.005$ & $+0.049 \pm 0.311$ & $-0.039 \pm 0.004$ \\
 & All questions & $-0.094 \pm 0.053$ & $-0.267 \pm 0.002$ & $-0.095 \pm 0.089$ & $-0.231 \pm 0.005$ & $-0.069 \pm 0.027$ & $-0.085 \pm 0.001$ & $-0.026 \pm 0.127$ & $-0.018 \pm 0.004$ \\
\addlinespace
Qwen3-VL-2B-Thinking & Happiness & $-0.362 \pm 0.035$ & $-0.133 \pm 0.002$ & $-0.474 \pm 0.077$ & $-0.087 \pm 0.011$ & $-0.164 \pm 0.032$ & $-0.106 \pm 0.001$ & $-0.054 \pm 0.217$ & $-0.104 \pm 0.010$ \\
 & Political ideology & $-0.064 \pm 0.060$ & $-0.087 \pm 0.005$ & $-0.027 \pm 0.166$ & $-0.063 \pm 0.004$ & $-0.031 \pm 0.062$ & $-0.068 \pm 0.005$ & $-0.047 \pm 0.215$ & $-0.052 \pm 0.005$ \\
 & Religious attendance & $+0.227 \pm 0.003$ & $+0.105 \pm 0.005$ & $+0.324 \pm 0.106$ & $+0.061 \pm 0.015$ & $+0.043 \pm 0.187$ & $+0.108 \pm 0.002$ & $-0.046 \pm 0.200$ & $+0.107 \pm 0.007$ \\
 & Social trust & $-0.247 \pm 0.048$ & $-0.235 \pm 0.009$ & $-0.266 \pm 0.171$ & $-0.246 \pm 0.027$ & $-0.156 \pm 0.185$ & $+0.161 \pm 0.014$ & $+0.034 \pm 0.575$ & $+0.453 \pm 0.007$ \\
 & All questions & $-0.111 \pm 0.011$ & $-0.088 \pm 0.002$ & $-0.111 \pm 0.032$ & $-0.084 \pm 0.009$ & $-0.077 \pm 0.074$ & $+0.024 \pm 0.003$ & $-0.028 \pm 0.130$ & $+0.101 \pm 0.001$ \\
\addlinespace
Llama-3.2-3B & Happiness & $-0.160 \pm 0.052$ & $-0.159 \pm 0.005$ & $-0.204 \pm 0.044$ & $-0.165 \pm 0.009$ & $-0.057 \pm 0.064$ & $+0.033 \pm 0.002$ & $+0.003 \pm 0.053$ & $-0.011 \pm 0.006$ \\
 & Political ideology & $-0.002 \pm 0.014$ & $-0.088 \pm 0.003$ & $-0.198 \pm 0.346$ & $+0.006 \pm 0.021$ & $-0.053 \pm 0.048$ & $-0.031 \pm 0.003$ & $-0.056 \pm 0.146$ & $+0.012 \pm 0.014$ \\
 & Religious attendance & $+0.103 \pm 0.030$ & $-0.230 \pm 0.006$ & $+0.077 \pm 0.323$ & $-0.218 \pm 0.009$ & $+0.166 \pm 0.081$ & $-0.042 \pm 0.004$ & $+0.080 \pm 0.343$ & $-0.028 \pm 0.005$ \\
 & Social trust & $-0.188 \pm 0.164$ & $-0.253 \pm 0.001$ & $-0.211 \pm 0.133$ & $-0.280 \pm 0.014$ & $+0.001 \pm 0.044$ & $+0.070 \pm 0.009$ & $+0.062 \pm 0.393$ & $+0.063 \pm 0.011$ \\
 & All questions & $-0.062 \pm 0.042$ & $-0.182 \pm 0.001$ & $-0.134 \pm 0.069$ & $-0.164 \pm 0.005$ & $+0.014 \pm 0.027$ & $+0.008 \pm 0.002$ & $+0.022 \pm 0.053$ & $+0.009 \pm 0.004$ \\
\addlinespace
Muse-Glimmer-30B & Happiness & $+0.013 \pm 0.030$ & $-0.035 \pm 0.001$ & $-0.045 \pm 0.096$ & $-0.030 \pm 0.004$ & $-0.025 \pm 0.010$ & $-0.003 \pm 0.003$ & $-0.112 \pm 0.136$ & $-0.044 \pm 0.002$ \\
 & Political ideology & $+0.072 \pm 0.023$ & $-0.007 \pm 0.005$ & $-0.040 \pm 0.034$ & $+0.022 \pm 0.010$ & $+0.017 \pm 0.057$ & $+0.063 \pm 0.007$ & $+0.042 \pm 0.191$ & $+0.072 \pm 0.006$ \\
 & Religious attendance & $-0.041 \pm 0.031$ & $-0.050 \pm 0.005$ & $-0.037 \pm 0.074$ & $-0.038 \pm 0.001$ & $+0.131 \pm 0.007$ & $+0.065 \pm 0.002$ & $+0.004 \pm 0.037$ & $+0.160 \pm 0.003$ \\
 & Social trust & $+0.091 \pm 0.030$ & $-0.028 \pm 0.000$ & $+0.031 \pm 0.183$ & $-0.048 \pm 0.004$ & $-0.106 \pm 0.043$ & $+0.077 \pm 0.004$ & $+0.104 \pm 0.125$ & $+0.043 \pm 0.007$ \\
 & All questions & $+0.034 \pm 0.019$ & $-0.030 \pm 0.002$ & $-0.023 \pm 0.036$ & $-0.024 \pm 0.003$ & $+0.004 \pm 0.027$ & $+0.051 \pm 0.001$ & $+0.010 \pm 0.005$ & $+0.058 \pm 0.001$ \\
\addlinespace
All models & Happiness & $-0.020 \pm 0.013$ & $-0.080 \pm 0.001$ & $-0.036 \pm 0.003$ & $-0.080 \pm 0.001$ & $-0.073 \pm 0.011$ & $-0.021 \pm 0.001$ & $-0.048 \pm 0.023$ & $-0.041 \pm 0.003$ \\
 & Political ideology & $-0.008 \pm 0.005$ & $-0.045 \pm 0.002$ & $+0.040 \pm 0.039$ & $-0.007 \pm 0.003$ & $+0.019 \pm 0.007$ & $+0.002 \pm 0.000$ & $-0.010 \pm 0.091$ & $+0.026 \pm 0.004$ \\
 & Religious attendance & $+0.046 \pm 0.007$ & $-0.089 \pm 0.002$ & $+0.060 \pm 0.065$ & $-0.097 \pm 0.001$ & $+0.044 \pm 0.035$ & $-0.012 \pm 0.002$ & $+0.008 \pm 0.082$ & $+0.059 \pm 0.003$ \\
 & Social trust & $-0.156 \pm 0.027$ & $-0.210 \pm 0.001$ & $-0.130 \pm 0.061$ & $-0.165 \pm 0.002$ & $-0.071 \pm 0.042$ & $+0.058 \pm 0.005$ & $+0.042 \pm 0.073$ & $+0.104 \pm 0.006$ \\
 & All questions & $-0.034 \pm 0.006$ & $-0.106 \pm 0.001$ & $-0.017 \pm 0.020$ & $-0.087 \pm 0.000$ & $-0.020 \pm 0.005$ & $+0.007 \pm 0.001$ & $-0.002 \pm 0.020$ & $+0.037 \pm 0.002$ \\
\bottomrule
\end{tabular}
\end{adjustbox}
\caption{
Paired fine-tuning effects across the three repeated FA runs. Each model--question entry reports the mean fine-tuned-minus-as-released change across the three runs $\pm$ its 95\% confidence interval, computed on the cells retained in all runs of both conditions. ``All questions'' averages the four question-specific changes within each run before constructing the interval, while ``All models'' averages the six model-specific changes within each run. Positive values indicate increases relative to the corresponding as-released condition.
}
\label{tab:replicates-paired}
\end{table}

\begin{table}[htb]
\centering
\tiny
\setlength{\tabcolsep}{3pt}
\begin{adjustbox}{max width=\textwidth}
\begin{tabular}{llrrrrrrrrrrrrrrr}
\toprule
 & & \multicolumn{5}{c}{As released} & \multicolumn{5}{c}{$\mathcal{F}_{\text{de}}$} & \multicolumn{5}{c}{$\mathcal{F}_{\text{mx}}$} \\
\cmidrule(lr){3-7}\cmidrule(lr){8-12}\cmidrule(lr){13-17}
Model & Question & Acc. & Adapt. & Struct. & PFS & Center & Acc. & Adapt. & Struct. & PFS & Center & Acc. & Adapt. & Struct. & PFS & Center \\
\midrule
Gemma-4-31B-it & Happiness & 0.000 & 0.018 & 0.005 & 0.023 & 0.001 & 0.000 & 0.025 & 0.003 & 0.002 & 0.002 & 0.001 & 0.065 & 0.000 & 0.000 & 0.001 \\
 & Political ideology & 0.001 & 0.006 & 0.020 & 0.034 & 0.001 & 0.001 & 0.023 & 0.034 & 0.044 & 0.002 & 0.002 & 0.020 & 0.034 & 0.053 & 0.003 \\
 & Religious attendance & 0.003 & 0.002 & 0.003 & 0.025 & 0.003 & 0.001 & 0.013 & 0.000 & 0.000 & 0.002 & 0.002 & 0.007 & 0.000 & 0.000 & 0.001 \\
 & Social trust & 0.003 & 0.030 & 0.005 & 0.009 & 0.001 & 0.005 & 0.016 & 0.002 & 0.102 & 0.007 & 0.001 & 0.031 & 0.005 & 0.005 & 0.004 \\
 & All questions & 0.001 & 0.009 & 0.006 & 0.013 & 0.001 & 0.001 & 0.011 & 0.009 & 0.030 & 0.002 & 0.001 & 0.020 & 0.009 & 0.014 & 0.001 \\
\addlinespace
Gemma-4-E4B-it & Happiness & 0.001 & 0.014 & 0.008 & 0.020 & 0.001 & 0.002 & 0.019 & 0.007 & 0.010 & 0.001 & 0.000 & 0.016 & 0.013 & 0.031 & 0.001 \\
 & Political ideology & 0.000 & 0.002 & 0.028 & 0.021 & 0.000 & 0.003 & 0.022 & 0.019 & 0.046 & 0.003 & 0.001 & 0.013 & 0.022 & 0.031 & 0.001 \\
 & Religious attendance & 0.001 & 0.019 & 0.000 & 0.000 & 0.002 & 0.002 & 0.007 & 0.003 & 0.030 & 0.004 & 0.003 & 0.012 & 0.004 & 0.146 & 0.005 \\
 & Social trust & 0.007 & 0.089 & 0.000 & 0.000 & 0.006 & 0.003 & 0.032 & 0.000 & 0.000 & 0.003 & 0.004 & 0.022 & 0.005 & 0.147 & 0.003 \\
 & All questions & 0.002 & 0.025 & 0.008 & 0.008 & 0.002 & 0.001 & 0.012 & 0.005 & 0.015 & 0.002 & 0.001 & 0.009 & 0.007 & 0.057 & 0.002 \\
\addlinespace
Qwen3-VL-8B-Thinking & Happiness & 0.001 & 0.019 & 0.030 & 0.051 & 0.002 & 0.003 & 0.023 & 0.049 & 0.046 & 0.004 & 0.001 & 0.021 & 0.054 & 0.074 & 0.002 \\
 & Political ideology & 0.002 & 0.029 & 0.010 & 0.022 & 0.003 & 0.001 & 0.005 & 0.028 & 0.036 & 0.001 & 0.002 & 0.016 & 0.063 & 0.093 & 0.001 \\
 & Religious attendance & 0.006 & 0.027 & 0.000 & 0.000 & 0.004 & 0.003 & 0.034 & 0.000 & 0.000 & 0.003 & 0.004 & 0.018 & 0.000 & 0.000 & 0.008 \\
 & Social trust & 0.008 & 0.078 & 0.040 & 0.052 & 0.008 & 0.003 & 0.007 & 0.040 & 0.160 & 0.003 & 0.005 & 0.025 & 0.033 & 0.073 & 0.004 \\
 & All questions & 0.003 & 0.024 & 0.014 & 0.020 & 0.003 & 0.001 & 0.011 & 0.018 & 0.046 & 0.001 & 0.002 & 0.011 & 0.024 & 0.037 & 0.003 \\
\addlinespace
Qwen3-VL-2B-Thinking & Happiness & 0.003 & 0.043 & 0.031 & 0.033 & 0.003 & 0.001 & 0.016 & 0.024 & 0.036 & 0.001 & 0.001 & 0.027 & 0.029 & 0.033 & 0.002 \\
 & Political ideology & 0.001 & 0.021 & 0.044 & 0.081 & 0.004 & 0.000 & 0.011 & 0.022 & 0.022 & 0.000 & 0.001 & 0.007 & 0.012 & 0.020 & 0.000 \\
 & Religious attendance & 0.005 & 0.001 & 0.000 & 0.000 & 0.004 & 0.002 & 0.009 & 0.001 & 0.003 & 0.002 & 0.004 & 0.020 & 0.009 & 0.187 & 0.002 \\
 & Social trust & 0.012 & 0.050 & 0.015 & 0.048 & 0.011 & 0.002 & 0.000 & 0.000 & 0.000 & 0.002 & 0.005 & 0.031 & 0.034 & 0.230 & 0.004 \\
 & All questions & 0.004 & 0.018 & 0.015 & 0.027 & 0.003 & 0.001 & 0.006 & 0.009 & 0.011 & 0.001 & 0.002 & 0.012 & 0.013 & 0.080 & 0.001 \\
\addlinespace
Llama-3.2-3B & Happiness & 0.003 & 0.041 & 0.034 & 0.052 & 0.005 & 0.001 & 0.011 & 0.026 & 0.023 & 0.000 & 0.002 & 0.018 & 0.031 & 0.052 & 0.004 \\
 & Political ideology & 0.002 & 0.009 & 0.047 & 0.080 & 0.000 & 0.004 & 0.023 & 0.053 & 0.072 & 0.003 & 0.001 & 0.018 & 0.054 & 0.126 & 0.003 \\
 & Religious attendance & 0.009 & 0.026 & 0.000 & 0.000 & 0.007 & 0.002 & 0.015 & 0.010 & 0.030 & 0.002 & 0.004 & 0.012 & 0.023 & 0.081 & 0.004 \\
 & Social trust & 0.003 & 0.049 & 0.014 & 0.028 & 0.004 & 0.002 & 0.027 & 0.023 & 0.189 & 0.002 & 0.005 & 0.054 & 0.024 & 0.054 & 0.006 \\
 & All questions & 0.003 & 0.019 & 0.016 & 0.027 & 0.002 & 0.001 & 0.011 & 0.017 & 0.055 & 0.001 & 0.002 & 0.016 & 0.019 & 0.045 & 0.002 \\
\addlinespace
Muse-Glimmer-30B & Happiness & 0.002 & 0.036 & 0.047 & 0.047 & 0.002 & 0.001 & 0.007 & 0.056 & 0.052 & 0.001 & 0.002 & 0.010 & 0.034 & 0.038 & 0.001 \\
 & Political ideology & 0.007 & 0.009 & 0.010 & 0.016 & 0.006 & 0.002 & 0.016 & 0.023 & 0.027 & 0.004 & 0.002 & 0.014 & 0.058 & 0.074 & 0.004 \\
 & Religious attendance & 0.002 & 0.010 & 0.015 & 0.006 & 0.003 & 0.007 & 0.018 & 0.049 & 0.030 & 0.007 & 0.001 & 0.005 & 0.014 & 0.008 & 0.001 \\
 & Social trust & 0.002 & 0.056 & 0.047 & 0.061 & 0.002 & 0.003 & 0.036 & 0.055 & 0.050 & 0.002 & 0.006 & 0.046 & 0.013 & 0.030 & 0.003 \\
 & All questions & 0.002 & 0.018 & 0.018 & 0.021 & 0.002 & 0.002 & 0.012 & 0.026 & 0.022 & 0.002 & 0.002 & 0.013 & 0.019 & 0.024 & 0.001 \\
\midrule
All models & Happiness & 0.001 & 0.016 & 0.015 & 0.020 & 0.001 & 0.001 & 0.009 & 0.017 & 0.017 & 0.001 & 0.001 & 0.016 & 0.016 & 0.022 & 0.001 \\
 & Political ideology & 0.002 & 0.008 & 0.015 & 0.025 & 0.002 & 0.001 & 0.009 & 0.016 & 0.022 & 0.001 & 0.001 & 0.008 & 0.023 & 0.038 & 0.001 \\
 & Religious attendance & 0.003 & 0.009 & 0.003 & 0.005 & 0.002 & 0.002 & 0.009 & 0.010 & 0.011 & 0.002 & 0.002 & 0.007 & 0.006 & 0.052 & 0.002 \\
 & Social trust & 0.003 & 0.031 & 0.013 & 0.020 & 0.003 & 0.002 & 0.012 & 0.015 & 0.056 & 0.002 & 0.002 & 0.019 & 0.011 & 0.060 & 0.002 \\
 & All questions & 0.002 & 0.017 & 0.012 & 0.018 & 0.002 & 0.001 & 0.009 & 0.014 & 0.029 & 0.001 & 0.001 & 0.012 & 0.014 & 0.042 & 0.002 \\
\bottomrule
\end{tabular}
\end{adjustbox}
\caption{
Run-to-run uncertainty by model, question, and condition. Question-specific entries are 95\% CI half-widths for the mean score across the three FA runs, $t_{0.975,\,c-1}\sigma_i/\sqrt{c}$. ``All questions'' and ``All models'' entries pool the corresponding run-level variance estimates using their combined degrees of freedom.
}
\label{tab:replicates-models}
\end{table}

\subsection{Agreement Across Elicitation Modes}
\label{app:mode-agreement}

To assess whether the results depend on how responses are elicited, we
compare the rankings of the 20 conditions evaluated under both NTP and
FA. Table~\ref{tab:mode-agreement} reports the Spearman correlation
between modes for the PFS, its components, and center alignment,
calculated separately for each question. Higher correlations indicate
that the relative ordering of model conditions is preserved across
elicitation methods.

\begin{table}[htb]
\centering
\tiny
\begin{adjustbox}{max width=\columnwidth}
\begin{tabular}{lrrrrr}
\toprule
Question & PFS & $S_{\mathrm{struct}}$ & $S_{\mathrm{acc}}$ & $S_{\mathrm{adapt}}$ & $S_{\mathrm{center}}$ \\
\midrule
Happiness & 0.838 & 0.791 & 0.968 & 0.788 & 0.925 \\
Political ideology & 0.695 & 0.549 & 0.744 & 0.567 & 0.792 \\
Religious attendance & 0.880 & 0.926 & 0.968 & 0.913 & 0.925 \\
Social trust & 0.864 & 0.823 & 0.974 & 0.502 & 0.986 \\
\bottomrule
\end{tabular}
\end{adjustbox}
\caption{Agreement between NTP and FA elicitation. Each entry is the
Spearman rank correlation between the two modes across the 20
conditions evaluated under both. The structure column uses the
untransformed correlation $\rho$.}
\label{tab:mode-agreement}
\end{table}

Fine-tuning effects are also directionally similar across elicitation modes.
NTP and FA agree on the sign of $\Delta\mathrm{PFS}$ in 38 of 48 paired
comparisons, or 35 of 41 after excluding comparisons with zero change
under either mode.

\subsection{Metric Sensitivity}
\label{app:sensitivity}

The Population Fidelity Score requires several measurement choices,
including how between-group dispersion is summarized, how structure is
measured, how the three components are combined, and how cells are
weighted. We test whether the main conclusions depend on these choices by
recomputing the analysis under alternative specifications while holding
the underlying model responses, retained cells, and nEMD distances fixed.

\paragraph{Alternative specifications.}
We consider seven changes individually and one combined specification.
For adaptability, we replace the median pairwise distance with the mean,
or use the one-sided score $\min(A,1)$ so that amplification is not
penalized. For structure, we replace Spearman's $\rho$ with Pearson's $r$
or Kendall's $\tau_b$. Separately, we test an unclipped version of the
default Spearman measure that preserves negative correlations rather than
mapping them to zero. We define the corresponding signed score as
\[
\mathrm{PFS}_{\mathrm{signed}}
=
\operatorname{sign}(\rho)
\left(
S_{\mathrm{acc}}
\cdot
S_{\mathrm{adapt}}
\cdot
|\rho|
\right)^{1/3},
\]
where $\operatorname{sign}(\rho)$ is $-1$, $0$, or $1$ according to whether
$\rho$ is negative, zero, or positive. Thus, the magnitude of the score is
computed from $|\rho|$, while its sign preserves the direction of the
structure correlation. Unlike the default PFS, this sensitivity variant can
take negative values.

We also replace the geometric mean with the arithmetic mean and,
separately, weight accuracy and center alignment by the number of survey
respondents in each cell. The combined specification uses mean pairwise
distance, Pearson structure without clipping, and respondent weighting.
Table~\ref{tab:sensitivity} compares these alternatives with the default
definition.

\begin{table}[htb]
\centering
\tiny
\setlength{\tabcolsep}{3.6pt}
\begin{adjustbox}{max width=\textwidth}
\begin{tabular}{lrrrrrrrrrrrr}
\toprule
& \multicolumn{4}{c}{Mean / median} &
\multicolumn{4}{c}{Stability} &
\multicolumn{2}{c}{Fine-tuning $\Delta$PFS} &
\multicolumn{2}{c}{Target-country $\Delta$PFS / $\Delta S_{\mathrm{center}}$} \\
\cmidrule(lr){2-5}\cmidrule(lr){6-9}\cmidrule(lr){10-11}\cmidrule(lr){12-13}
Variant & Acc. & Adapt. & Struct. & PFS &
Struct.\ limit & Struct.\ limit (4{,}920) & PFS $\leq 0$ &
$\rho$ vs.\ default &
German & Mexican &
Germany & Mexico \\
\midrule
Default (Section~\ref{sec:population-fidelity})
& 0.775 / 0.832 & 0.496 / 0.497 & 0.082 / 0.060 & 0.243 / 0.254
& 159 & 4{,}554 & 30 & 1.000
& $-0.024$ (21/23) & $-0.014$ (19/23)
& $-0.021$ / $-0.084$ & $-0.010$ / $+0.038$ \\

Mean pairwise distance
& 0.775 / 0.832 & 0.521 / 0.508 & 0.082 / 0.060 & 0.248 / 0.265
& 164 & 4{,}679 & 30 & 0.995
& $-0.024$ (22/22) & $-0.017$ (18/24)
& $-0.017$ / $-0.084$ & $-0.009$ / $+0.038$ \\

One-sided $\min(A,1)$
& 0.775 / 0.832 & 0.545 / 0.519 & 0.082 / 0.060 & 0.246 / 0.254
& 159 & 4{,}558 & 30 & 0.997
& $-0.025$ (21/23) & $-0.014$ (19/23)
& $-0.020$ / $-0.084$ & $-0.011$ / $+0.038$ \\

Pearson $r$
& 0.775 / 0.832 & 0.496 / 0.497 & 0.098 / 0.067 & 0.257 / 0.262
& 156 & 4{,}486 & 27 & 0.988
& $-0.028$ (20/24) & $-0.025$ (16/26)
& $-0.018$ / $-0.084$ & $-0.016$ / $+0.038$ \\

Kendall $\tau_b$
& 0.775 / 0.832 & 0.496 / 0.497 & 0.055 / 0.040 & 0.213 / 0.222
& 162 & 4{,}714 & 30 & 1.000
& $-0.021$ (21/23) & $-0.012$ (19/23)
& $-0.018$ / $-0.084$ & $-0.009$ / $+0.038$ \\

Unclipped structure
& 0.775 / 0.832 & 0.496 / 0.497 & 0.080 / 0.060 & 0.216 / 0.254
& 160 & 4{,}596 & 30 & 0.997
& $-0.016$ (22/26) & $-0.016$ (22/26)
& $-0.012$ / $-0.084$ & $-0.001$ / $+0.038$ \\

Arithmetic mean
& 0.775 / 0.832 & 0.496 / 0.497 & 0.082 / 0.060 & 0.451 / 0.455
& 159 & 4{,}554 & 0 & 0.669
& $-0.058$ (18/30) & $-0.015$ (21/27)
& $-0.071$ / $-0.084$ & $-0.020$ / $+0.038$ \\

Respondent-weighted accuracy and center
& 0.773 / 0.833 & 0.496 / 0.497 & 0.082 / 0.060 & 0.243 / 0.255
& 159 & 4{,}554 & 30 & 1.000
& $-0.024$ (21/23) & $-0.014$ (19/23)
& $-0.021$ / $-0.090$ & $-0.010$ / $+0.038$ \\

Combined
& 0.773 / 0.833 & 0.521 / 0.508 & 0.094 / 0.067 & 0.232 / 0.263
& 159 & 4{,}606 & 27 & 0.979
& $-0.023$ (22/26) & $-0.033$ (16/32)
& $+0.016$ / $-0.090$ & $-0.017$ / $+0.038$ \\
\bottomrule
\end{tabular}
\end{adjustbox}
\caption{
Sensitivity of Population Fidelity to alternative metric specifications.
Acc., Adapt., Struct., and PFS report the mean and median across the 164
evaluated combinations. ``Struct.\ limit'' reports how often structure is
the lowest PFS component, first at the population level and then across
4{,}920 subgroup-level scores, including the Mixtral reference
series. $\rho$ vs.\ default is the Spearman
correlation between PFS under each specification and under the default.
Fine-tuning columns report mean paired changes in PFS; parentheses give
the numbers of increases and decreases. The final columns report mean
changes in PFS and center alignment within the corresponding target
country. For specifications without clipping, the structure value retains
its sign and PFS may be negative.
}
\label{tab:sensitivity}
\end{table}

\paragraph{Main conclusions are stable.}
The central results are largely unchanged across specifications. Accuracy
remains substantially higher than structure: mean accuracy ranges only from
$0.773$ to $0.775$, whereas mean structure ranges from $0.055$ to $0.098$.
Structure remains the limiting component in at least 156 of 164
population-level evaluations and 4{,}486 of 4{,}920 subgroup-level
evaluations; accuracy is never limiting. Among specifications retaining the
geometric mean, PFS rankings remain highly similar to the default
($\rho \geq 0.979$).

At the population-wide scope, neither CultureLLM-style condition improves
PFS on average under any specification. Among geometric-mean specifications,
mean paired changes range from $-0.028$ to $-0.016$ for German fine-tuning
and from $-0.033$ to $-0.012$ for Mexican fine-tuning. Target-country
results are somewhat more specification-sensitive. Within Mexico, center
alignment improves by approximately $0.038$ across specifications, whereas
target-country PFS is negative under all but the unclipped specification,
where the mean change is effectively zero. Within Germany, target-country
PFS is negative under the default and most alternative specifications but
becomes positive under the combined specification
($\Delta\mathrm{PFS}=+0.016$). Thus, the population-wide conclusion is
robust, while the sign of a target-country mean can depend on the metric
specification.

\paragraph{Where the specification matters.}
Some choices affect the numerical scale without changing the main
conclusions. Replacing median with mean pairwise distance raises average
adaptability from $0.496$ to $0.521$. Pearson correlation produces somewhat
higher structure scores than Spearman, while Kendall correlation produces
somewhat lower scores. Removing the zero clipping converts most
nonpositive-structure cases into negative PFS values rather than zeros.

The largest change comes from replacing the geometric mean with an
arithmetic mean. Mean PFS rises from $0.243$ to $0.451$, all exact zeros
disappear, and the ranking correlation with the default falls to $0.669$.
This occurs because the arithmetic mean allows high accuracy and
adaptability to compensate for very low structure, whereas the geometric
mean treats the three dimensions as jointly required.

Finally, paired fine-tuning comparisons are evaluated on the intersection
of cells retained by the fine-tuned and as-released conditions. Scoring
each condition on its own retained cells produces slightly different
changes because response coverage itself then affects adaptability and
structure. Across the sensitivity specifications, this alternative does
not change the sign of the pooled mean fine-tuning effects. We therefore
retain the common-cell construction so that paired differences reflect
changes in model responses rather than changes in evaluation support.

\section{Distribution-Matched and Subgroup-Matched Fine-Tuning}
\label{app:distribution-matched}

The CultureLLM-style conditions used in the main analysis train toward a
single target answer per question. The reduction in between-group variation
reported in Section~\ref{sec:results} may therefore reflect that objective
rather than survey-based adaptation in general. To examine whether Population
Fidelity distinguishes alternative adaptation objectives, we evaluate two
additional methods on a common backbone. We compare the as-released model with
four adaptation conditions: CultureLLM-style German fine-tuning
($\mathcal{F}_{\text{de}}$), CultureLLM-style Mexican fine-tuning
($\mathcal{F}_{\text{mx}}$), distribution matching (SimLLCultureDist), and
subgroup matching (SubPOP).

SimLLCultureDist \citep{cao2025specializing} trains toward aggregate
country-level response distributions. These targets contain country-level response distributions but no within-country demographic subgroup information. SubPOP \citep{suh2025language}, by contrast, trains toward response
distributions for demographic subgroups and therefore explicitly incorporates
subgroup-level information. These conditions provide complementary examples of how fine-tuning approaches with different supervision targets score on the PFS components.

We fine-tune Qwen3-VL-2B-Thinking using the released data and training
recipes of SimLLCultureDist and SubPOP. SimLLCultureDist contains 6{,}841
country-level response distributions from 46 countries and uses a
first-token KL-divergence objective. SubPOP-Train contains 71{,}026
subgroup-level response distributions covering 3{,}229 American Trends Panel
questions and 22 United States subgroups, using the same loss. We evaluate
both conditions using the same procedure as the other model conditions.
Table~\ref{tab:distribution-matched} reports first-run scores; neither
condition is included among the 164 combinations analyzed in the main text.

The training data also differ in their overlap with the evaluation items.
The distribution-matched data contain the happiness and social-trust items
for four of the five evaluation countries. SubPOP-Train contains eight
religious-attendance items and one social-trust item for United States
subgroups only. Happiness and political ideology are absent from SubPOP-Train,
and neither Germany nor Mexico is represented as a subgroup-training target.

\begin{table}[htb]
\centering
\tiny
\setlength{\tabcolsep}{4pt}
\begin{adjustbox}{max width=\textwidth}
\begin{tabular}{llrrrrrrrrrr}
\toprule
 & & \multicolumn{5}{c}{FA} & \multicolumn{5}{c}{NTP} \\
\cmidrule(lr){3-7}\cmidrule(lr){8-12}
Question & Condition & $S_{\mathrm{acc}}$ & $A$ & $S_{\mathrm{struct}}$ & PFS & $S_{\mathrm{center}}$ &
$S_{\mathrm{acc}}$ & $A$ & $S_{\mathrm{struct}}$ & PFS & $S_{\mathrm{center}}$ \\
\midrule
Happiness & As released & 0.892 & 0.913 & 0.143 & 0.488 & 0.979 & 0.894 & 0.704 & 0.162 & 0.467 & 0.962 \\
 & German ($\mathcal{F}_{\text{de}}$) & 0.845 & 0.077 & 0.044 & 0.141 & 0.846 & 0.856 & 0.133 & 0.058 & 0.187 & 0.859 \\
 & Mexican ($\mathcal{F}_{\text{mx}}$) & 0.860 & 0.308 & 0.146 & 0.338 & 0.873 & 0.879 & 0.371 & 0.154 & 0.369 & 0.905 \\
 & Distribution-matched & 0.894 & 0.723 & 0.129 & 0.437 & 0.953 & 0.899 & 0.598 & 0.140 & 0.422 & 0.965 \\
 & Subgroup-matched & 0.898 & 0.485 & 0.036 & 0.250 & 0.952 & 0.897 & 0.192 & 0.113 & 0.269 & 0.971 \\
\addlinespace
Political ideology & As released & 0.883 & 0.419 & 0.063 & 0.285 & 0.905 & 0.904 & 0.189 & 0.070 & 0.229 & 0.948 \\
 & German ($\mathcal{F}_{\text{de}}$) & 0.817 & 0.150 & 0.060 & 0.194 & 0.818 & 0.859 & 0.129 & 0.018 & 0.127 & 0.867 \\
 & Mexican ($\mathcal{F}_{\text{mx}}$) & 0.831 & 0.262 & 0.053 & 0.226 & 0.837 & 0.848 & 0.132 & 0.057 & 0.185 & 0.855 \\
 & Distribution-matched & 0.883 & 0.501 & 0.049 & 0.278 & 0.910 & 0.902 & 0.219 & 0.074 & 0.245 & 0.959 \\
 & Subgroup-matched & 0.835 & 0.525 & 0.047 & 0.274 & 0.858 & 0.872 & 0.121 & 0.028 & 0.144 & 0.906 \\
\addlinespace
Religious attendance & As released & 0.690 & 0.320 & 0.000 & 0.000 & 0.695 & 0.762 & 0.295 & 0.000 & 0.000 & 0.779 \\
 & German ($\mathcal{F}_{\text{de}}$) & 0.744 & 0.567 & 0.027 & 0.225 & 0.801 & 0.777 & 0.526 & 0.021 & 0.204 & 0.837 \\
 & Mexican ($\mathcal{F}_{\text{mx}}$) & 0.761 & 0.475 & 0.000 & 0.000 & 0.803 & 0.796 & 0.431 & 0.000 & 0.000 & 0.844 \\
 & Distribution-matched & 0.763 & 0.487 & 0.000 & 0.000 & 0.784 & 0.813 & 0.319 & 0.000 & 0.000 & 0.856 \\
 & Subgroup-matched & 0.751 & 0.280 & 0.005 & 0.102 & 0.764 & 0.796 & 0.138 & 0.005 & 0.082 & 0.828 \\
\addlinespace
Social trust & As released & 0.585 & 1.038 & 0.022 & 0.233 & 0.589 & 0.632 & 0.810 & 0.048 & 0.291 & 0.636 \\
 & German ($\mathcal{F}_{\text{de}}$) & 0.350 & 0.000 & 0.000 & 0.000 & 0.350 & 0.359 & 0.085 & 0.000 & 0.000 & 0.359 \\
 & Mexican ($\mathcal{F}_{\text{mx}}$) & 0.741 & 0.316 & 0.000 & 0.000 & 0.745 & 0.802 & 0.231 & 0.000 & 0.000 & 0.815 \\
 & Distribution-matched & 0.776 & 1.251 & 0.091 & 0.383 & 0.819 & 0.791 & 0.818 & 0.119 & 0.425 & 0.814 \\
 & Subgroup-matched & 0.537 & 0.509 & 0.005 & 0.111 & 0.537 & 0.605 & 0.246 & 0.046 & 0.190 & 0.605 \\
\midrule
Mean change from as released & German ($\mathcal{F}_{\text{de}}$) & $-0.073$ & $-0.474$ & $-0.024$ & $-0.111$ & $-0.088$ & $-0.085$ & $-0.281$ & $-0.046$ & $-0.118$ & $-0.101$ \\
 & Mexican ($\mathcal{F}_{\text{mx}}$) & $+0.036$ & $-0.332$ & $-0.007$ & $-0.110$ & $+0.023$ & $+0.033$ & $-0.208$ & $-0.018$ & $-0.108$ & $+0.023$ \\
 & Distribution-matched & $+0.066$ & $+0.068$ & $+0.010$ & $+0.023$ & $+0.074$ & $+0.053$ & $-0.011$ & $+0.013$ & $+0.026$ & $+0.067$ \\
 & Subgroup-matched & $-0.007$ & $-0.223$ & $-0.034$ & $-0.067$ & $-0.015$ & $-0.006$ & $-0.325$ & $-0.022$ & $-0.076$ & $-0.004$ \\
\midrule
\multicolumn{2}{l}{Both modes pooled} &
\multicolumn{2}{c}{$S_{\mathrm{acc}}$} &
\multicolumn{2}{c}{$A$} &
\multicolumn{2}{c}{$S_{\mathrm{struct}}$} &
\multicolumn{2}{c}{PFS} &
\multicolumn{2}{c}{$S_{\mathrm{center}}$} \\
\cmidrule(lr){3-12}
Median over the eight combinations & As released & \multicolumn{2}{c}{0.823} & \multicolumn{2}{c}{0.561} & \multicolumn{2}{c}{0.055} & \multicolumn{2}{c}{0.259} & \multicolumn{2}{c}{0.842} \\
 & German ($\mathcal{F}_{\text{de}}$) & \multicolumn{2}{c}{0.797} & \multicolumn{2}{c}{0.131} & \multicolumn{2}{c}{0.024} & \multicolumn{2}{c}{0.164} & \multicolumn{2}{c}{0.827} \\
 & Mexican ($\mathcal{F}_{\text{mx}}$) & \multicolumn{2}{c}{0.816} & \multicolumn{2}{c}{0.312} & \multicolumn{2}{c}{0.027} & \multicolumn{2}{c}{0.093} & \multicolumn{2}{c}{0.841} \\
 & Distribution-matched & \multicolumn{2}{c}{0.848} & \multicolumn{2}{c}{0.549} & \multicolumn{2}{c}{0.082} & \multicolumn{2}{c}{0.330} & \multicolumn{2}{c}{0.883} \\
 & Subgroup-matched & \multicolumn{2}{c}{0.815} & \multicolumn{2}{c}{0.263} & \multicolumn{2}{c}{0.032} & \multicolumn{2}{c}{0.167} & \multicolumn{2}{c}{0.843} \\
\bottomrule
\end{tabular}
\end{adjustbox}
\caption{
Population Fidelity diagnostics for Qwen3-VL-2B-Thinking under
CultureLLM-style German fine-tuning ($\mathcal{F}_{\text{de}}$),
CultureLLM-style Mexican fine-tuning ($\mathcal{F}_{\text{mx}}$),
distribution matching (SimLLCultureDist), and subgroup matching (SubPOP).
Scores are from the first run on all retained cells (687 cells; 639 for
political ideology). $A$ is the raw adaptability ratio; PFS uses the
corresponding symmetric adaptability score $S_{\mathrm{adapt}}$.
``Mean change from as released'' averages the within-question difference
from the as-released condition across the four questions. The final five
rows report the median of each condition across its eight question--mode
combinations.
}
\label{tab:distribution-matched}
\end{table}

The distribution-matched condition attains the highest median center
alignment of the five conditions on this backbone
($0.883$ versus $0.827$--$0.843$) and improves on the as-released model in
seven of eight question--mode combinations. PFS gives a more mixed result:
it improves in three combinations, declines in three, and remains zero in
two, with median PFS increasing from $0.259$ to $0.330$.

\begin{figure}[tbh]
\centering
\begin{subfigure}[t]{0.49\textwidth}
\centering
\includegraphics[width=\linewidth]{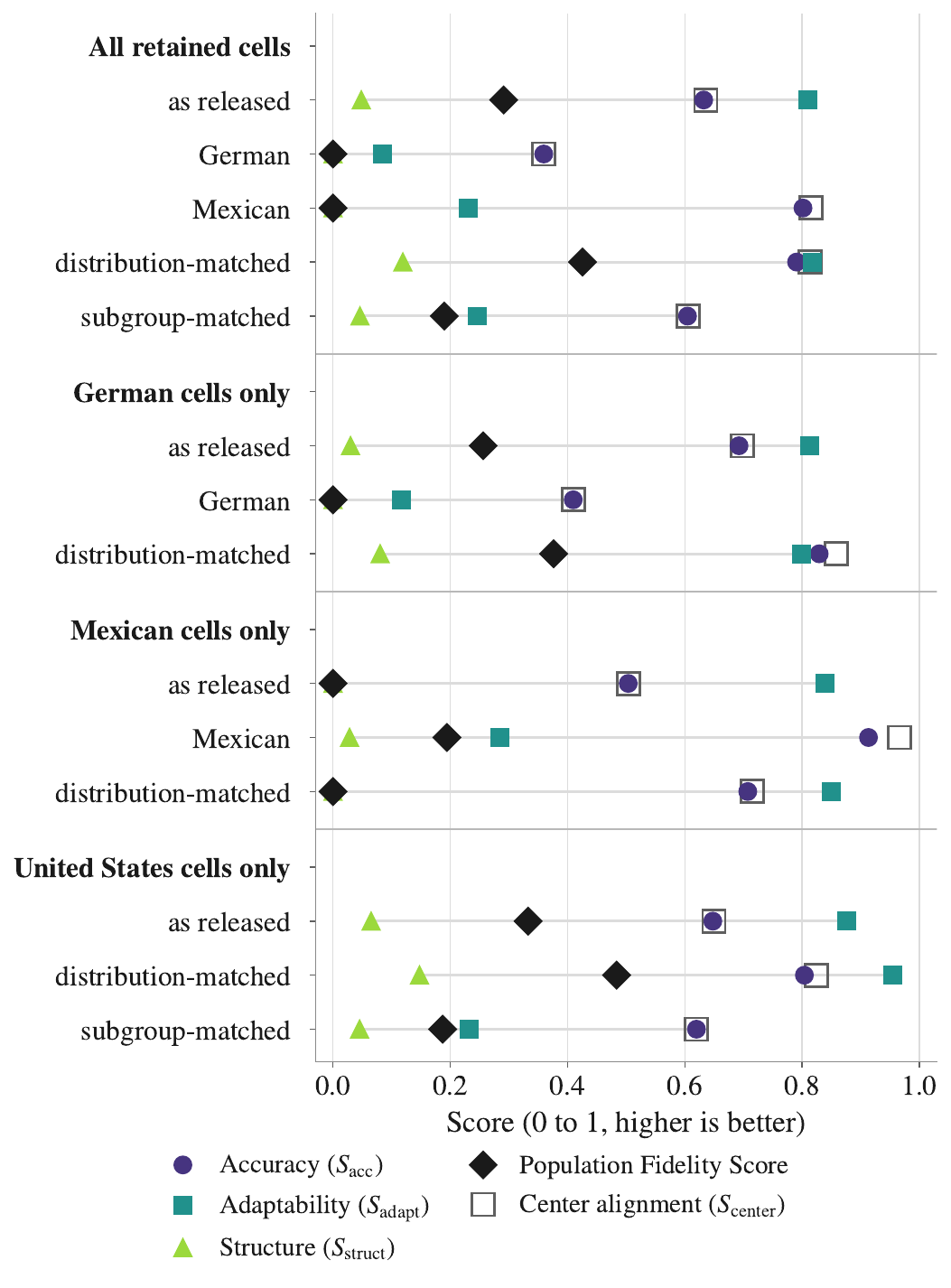}
\caption{Social trust, NTP: component profiles on all cells, German cells, and Mexican cells.}
\end{subfigure}
\hfill
\begin{subfigure}[t]{0.49\textwidth}
\centering
\includegraphics[width=\linewidth]{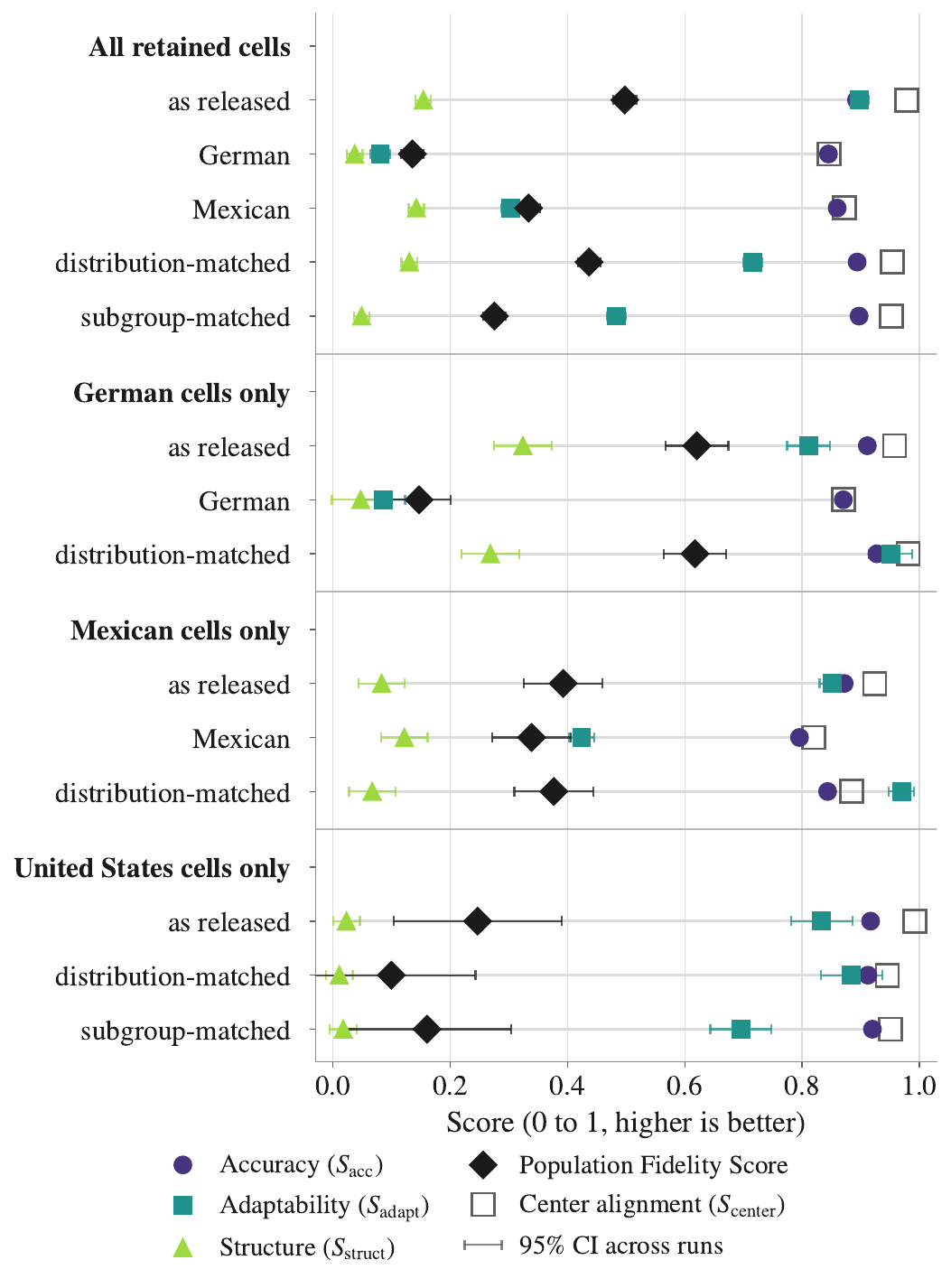}
\caption{Happiness, FA: component profiles on all cells, German cells, and Mexican cells.}
\end{subfigure}

\par\medskip

\begin{subfigure}[t]{0.40\textwidth}
\centering
\includegraphics[width=\linewidth]{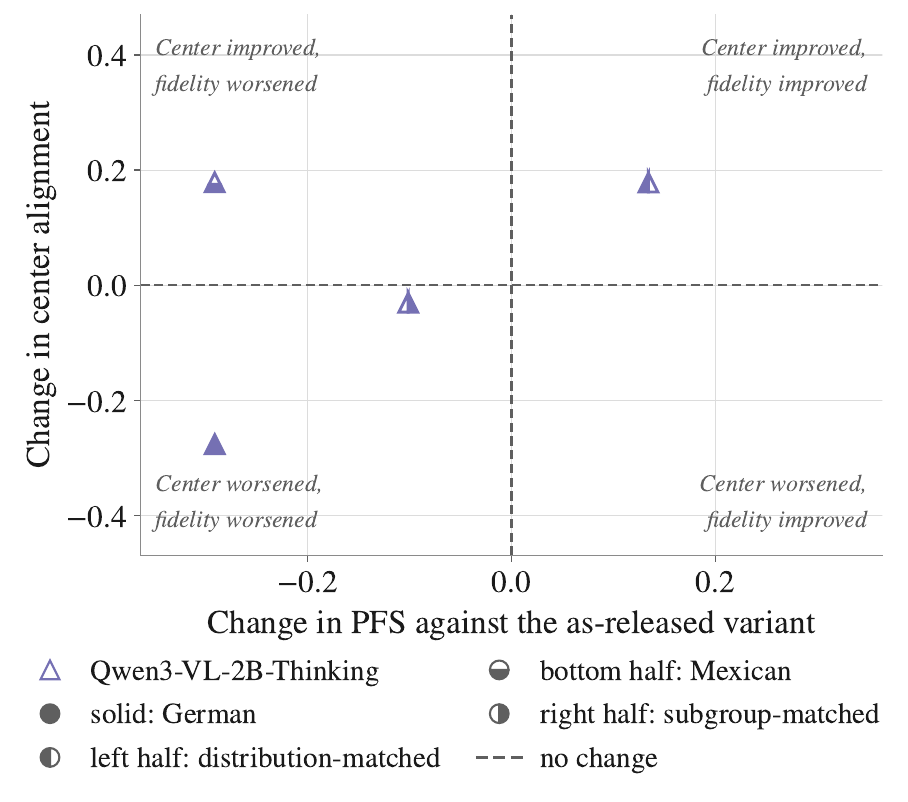}
\caption{Social trust, NTP: change from as released, all cells.}
\end{subfigure}
\hspace{0.06\textwidth}
\begin{subfigure}[t]{0.40\textwidth}
\centering
\includegraphics[width=\linewidth]{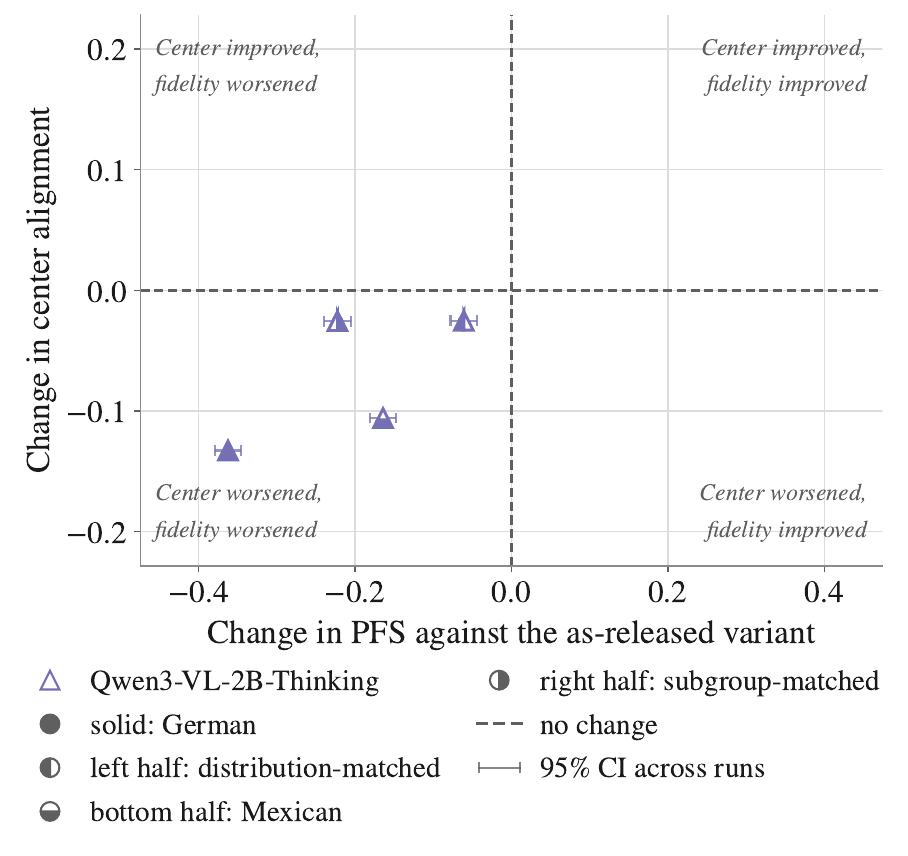}
\caption{Happiness, FA: change from as released, all cells.}
\end{subfigure}

\caption{
Population Fidelity and center alignment for five Qwen3-VL-2B-Thinking
conditions: as released, $\mathcal{F}_{\text{de}}$,
$\mathcal{F}_{\text{mx}}$, distribution-matched, and subgroup-matched.
\textbf{(a,b)} Component profiles on all retained cells, German cells, and
Mexican cells for social trust under NTP and happiness under FA,
respectively. Diamonds show PFS and hollow squares center alignment.
\textbf{(c,d)} Adapted minus as-released changes on all retained cells for
the same two settings.
}
\label{fig:distribution-matched}
\end{figure}

Center alignment can also obscure large differences in the PFS components.
For social trust under NTP, $\mathcal{F}_{\text{mx}}$ and the
distribution-matched condition attain nearly identical center alignment
($0.815$ and $0.814$), but their PFS values are $0.000$ and $0.425$,
respectively. $\mathcal{F}_{\text{mx}}$ strongly compresses between-group
variation ($A=0.231$) and recovers no positive structure
($S_{\mathrm{struct}}=0$), whereas the distribution-matched condition
preserves substantially more variation ($A=0.818$) and positive structure
($S_{\mathrm{struct}}=0.119$). Thus, similar center alignment can coexist
with very different representations of variation among groups
(Figure~\ref{fig:distribution-matched}).

The subgroup-matched condition does not consistently recover group structure
in this evaluation. Its median adaptability ratio is $A=0.263$, median
$S_{\mathrm{struct}}=0.032$, and structure is the limiting component in all
eight question--mode combinations. PFS declines relative to the as-released
model in six of eight combinations. Its two improvements occur for religious attendance, which has the greatest
item-level overlap with the SubPOP training data among our evaluation questions. For happiness under FA, for example, the distribution-matched and
subgroup-matched conditions have nearly identical center alignment
($0.953$ and $0.952$) but PFS values of $0.437$ and $0.250$, respectively,
reflecting lower adaptability and structure under subgroup matching.

These results should be interpreted as transfer across datasets and
population definitions: SubPOP is trained on United States demographic
subgroups, whereas our evaluation compares demographic cells across five WVS
countries. They therefore show that subgroup-level distributional training
does not automatically transfer to the between-group structure of these
data, rather than that subgroup matching generally fails to recover
population structure.

Because the CultureLLM-style conditions target individual countries,
Figure~\ref{fig:distribution-matched} also evaluates all conditions on German
and Mexican cells separately. Within Germany,
$\mathcal{F}_{\text{de}}$ reduces both center alignment and PFS in six of
eight combinations and compresses variation among German cells
(median $A=0.164$ versus $0.733$ for the as-released model). Within Mexico,
$\mathcal{F}_{\text{mx}}$ raises center alignment by $0.103$ on average,
but PFS improves in only two combinations and declines in four; median $A$
falls from $0.728$ to $0.453$.

The additional distribution-trained conditions likewise show that changes
in center alignment do not uniquely determine changes in PFS. Across the
target-country evaluations, structure remains the limiting component in 77
of 80 combinations. These experiments involve only one backbone and
substantially different training targets and data coverage, so they are
intended to demonstrate how PFS distinguishes adaptation profiles rather
than to establish comparative performance among fine-tuning methods.

\end{document}